\documentclass[11pt, letterpaper, twoside, openright]{book}

\usepackage[utf8]{inputenc}
\usepackage{amsfonts}
\usepackage{amsmath}
\usepackage{dialogue}
\usepackage{csquotes}

\usepackage[a4paper,width=150mm,top=23mm,bottom=25mm,bindingoffset=15mm]{geometry}
\usepackage{setspace}
\usepackage{emptypage}

\usepackage{changepage}
\def\arraystretch{1.3}
\usepackage{parskip}

\usepackage{enumitem}
\setenumerate{noitemsep,topsep=0pt,parsep=0pt,partopsep=0pt}

\usepackage{fancyhdr}
\fancypagestyle{plain}{%
    \fancyhf{}%
    \fancyfoot[C]{\thepage}%
}

\usepackage[backend=bibtex,natbib=true,
			style=authoryear,citestyle=authoryear,dashed=false,
			maxcitenames=2,firstinits=true,hyperref,uniquename=init]{biblatex}

\newcommand{\bncref}[2]{\begin{footnotesize}[BNC: #1 #2]\end{footnotesize}}

\newcounter{examplecounter}[chapter]
\newenvironment{example}{
	\begin{enumerate}[label=(\thechapter.\arabic*), ref=\arabic*, labelsep=20pt, leftmargin=55pt, ref=(\thechapter.\theenumi)]
    \setcounter{enumi}{\value{examplecounter}}
    \stepcounter{examplecounter}
    \item \setlength{\labelsep}{10pt}
}{
	\end{enumerate}
}
\newenvironment{example*}{
	\begin{enumerate}
    \setcounter{enumi}{\value{examplecounter}}
    \item \setlength{\labelsep}{10pt}
}{
	\end{enumerate}
}

\usepackage{color}

\usepackage[ruled, linesnumbered]{algorithm2e}
\usepackage[hidelinks]{hyperref}
\usepackage{arydshln}
\usepackage{graphicx}
\usepackage{wrapfig}
\usepackage{mathtools}

\usepackage{ragged2e}

\usepackage{acronym}
\usepackage[toc]{appendix}

\newcommand{\leqnomode}{\tagsleft@true}

\usepackage[multiple]{footmisc}
\usepackage{pdfpages}

\usepackage{titlesec}

\begin{document}

\frontmatter
\includepdf[pages=-]{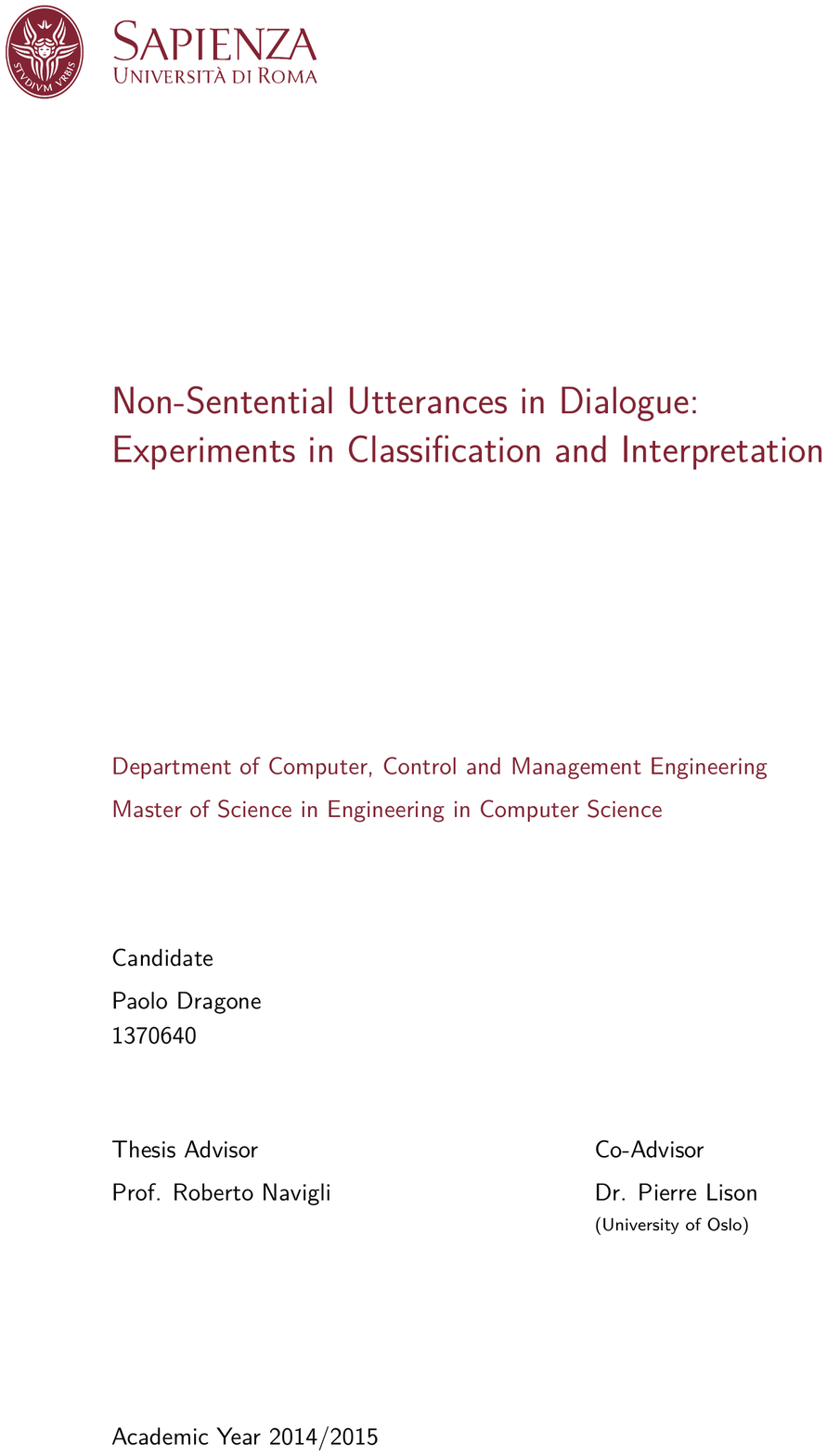}

\chapter*{\centering Abstract}

Non-sentential utterances (NSUs) are utterances that lack a complete sentential form but whose meaning can be inferred from the dialogue context, such as ``OK'', ``where?'', ``probably at his apartment''. The interpretation of non-sentential utterances is an important problem in computational linguistics since they constitute a frequent phenomena in dialogue and they are intrinsically context-dependent. The interpretation of NSUs is the task of retrieving their full semantic content from their form and the dialogue context.

NSUs also come in a wide variety of forms and functions and classifying them in the right category is a prerequisite to their interpretation. The first half of this thesis is devoted to the NSU classification task. Our work builds upon \citet{Fernandez:2007} which present a series of machine-learning experiments on the classification of NSUs. We extended their approach with a combination of new features and semi-supervised learning techniques. The empirical results presented in this thesis show a modest but significant improvement over the state-of-the-art classification performance.

The consecutive, yet independent, problem is how to infer an appropriate semantic representation of such NSUs on the basis of the dialogue context. \citet{Fernandez:thesis} formalizes this task in terms of ``resolution rules'' built on top of the Type Theory with Records (TTR), a theoretical framework for dialogue context modeling \citep{Ginzburg:interactivestance}. We argue that logic-based formalisms, such as TTR, have a number of shortcomings when dealing with conversational data, which often include partially observable knowledge and non-deterministic phenomena. An alternative to address these issues is to rely on probabilistic modeling of the dialogue context. Our work is focused on the reimplementation of the resolution rules from \citet{Fernandez:thesis} with a probabilistic account of the dialogue state. The probabilistic rules formalism \citep{LisonThesis2014} is particularly suited for this task because, similarly to the framework developed by \citet{Ginzburg:interactivestance} and \citet{Fernandez:thesis}, it involves the specification of \textit{update rules} on the variables of the dialogue state to capture the dynamics of the conversation. However, the probabilistic rules can also encode probabilistic knowledge, thereby providing a principled account of ambiguities in the NSU resolution process.
In the second part of this thesis, we present our \textit{proof-of-concept} framework for NSU resolution using probabilistic rules.

\chapter*{\centering Acknowledgments}

The past two years have been utterly intense, but beautiful nonetheless. In all my wandering around the Earth, first in Melbourne then in Oslo, I have learned an incredible amount of things and met as many interesting people. I have studied hard, explored the world, suffered homesickness and experienced many wonderful moments. It has been an exiting time and now, at the finishing line, I am walking towards the end with a smile on my face. For those opportunities I am grateful to my alma mater Sapienza which, despite all the difficulties, gave me also many positive memories.

This thesis is the product of the period I spent at University of Oslo under the supervision of Dr. Pierre Lison. Pierre has not only been a great supervisor but he has also been my mentor. He has taught me an uncountable number of things. He introduced me to the research on dialogue and proposed the topic of non-sentential utterances in the first place. My collaboration with Pierre has also led to my very first scientific publication at the SemDial 2015 in G\"oteborg. I can safely say that writing this thesis would not have been possible without his guidance. For these and many other reasons I want to thank him and wish him all the best. I also want to thank all the Language Technology Group at University of Oslo for their hospitality and for the many (or perhaps too few) happy days we spent together.

A thankful note to all my friends too, who support me everyday of my life, even (and especially) when I am not around. In particular, a big thank you to all my friends and colleagues from the course in Engineering in Computer Science with whom I walked side by side for all these years. However, a special thanks goes to my estimate colleague Alessandro Ronca who has practically been my moral supporter as well as my personal reviewer.

Last but not least, I want to express my gratitude to my parents and all my family who have always encouraged me to go beyond my own limits (even if sometimes I went perhaps too far away) and never stopped believing in me.\\\\
\begin{flushright}
Rome, 22th of October 2015

\textit{Paolo Dragone}
\end{flushright}

\tableofcontents

\newpage

\chapter*{List of acronyms}
\begin{acronym}[MaxQUD]
\acro{AL}{Active Learning}
\acro{BNC}{British National Corpus}
\acro{DGB}{Dialogue Gameboard}
\acro{FEC}{Focus-Establishing Constituent}
\acro{MaxQUD}{Maximal QUD element}
\acro{NLG}{Natural Language Generation}
\acro{NLU}{Natural Language Understanding}
\acro{NSU}{Non-Sentential Utterance}
\acro{ParPar}{Partial Parallelism}
\acro{QUD}{Question Under Discussion}
\acro{SMO}{Sequential Minimal Optimization}
\acro{SVM}{Support Vector Machine}
\acro{TSVM}{Transductive Support Vector Machine}
\acro{TTR}{Type Theory with Records}

\end{acronym}

\mainmatter

\chapter{Introduction}
\label{ch:introduction}
In dialogue, utterances do not always take the form of complete sentences. Utterances may sometimes lack some constituents -- subject, verb or complements -- because they can be understood from the previous utterances or other contextual information. These fragmentary utterances are often called \textit{non-sentential utterances} (NSUs). The following are some examples from the British National Corpus: 
\begin{example}
\begin{dialogue}
\speak{A} How do you actually feel about that?
\speak{B} \textbf{Not too happy.} \\ \bncref{JK8}{168--169}\footnote{This notation indicates the file name and the line numbers of the portion of dialogues in the British National Corpus.}
\end{dialogue}
\end{example}
\begin{example}
\label{ex:why}
\begin{dialogue}
\speak{A} They wouldn't do it, no.
\speak{B} \textbf{Why?} \\ \bncref{H5H}{202--203}
\end{dialogue}
\end{example}
\begin{example}
	\label{ex:3}
	\begin{dialogue}
		\speak{A} So will the tape last for the whole two hours?
        \speak{B} \textbf{Yes, apparently.} \\ \bncref{J9A}{76--77}
	\end{dialogue}
\end{example}

\begin{example}
	\label{ex:4}
	\begin{dialogue}
		\speak{A} Right disk number four?
        \speak{B} \textbf{Three.} \\ \bncref{HDH}{377--378}
	\end{dialogue}
\end{example}

We can understand without effort the meaning of the NSUs in the short dialogues above, even though they do not have the form of full sentences. We can easily make sense of them by extrapolating their meaning from the surrounding context, which for the above examples is given by the preceding utterance. Other possible contextual factors that affect the intended meaning of the NSUs are, for instance, the history of the dialogue, the shared environment of the conversational participants, their common knowledge and so on. From a computational linguistic perspective, making sense of this kind of utterances is a difficult problem because it involves the formalization of a robust theory of dialogue context.

Moreover, NSUs are a large variety of phenomena that need to be treated in different ways. \citet{Fernandez:2002} identify 15 different types of NSUs. One of the problems that must be addressed to make sense of NSUs is determining their type. One possible way is to classify NSUs using machine learning, as previously experimented by \citet{Fernandez:2007}.

To interpret a given NSU, one also has to \textit{resolve} its meaning i.e. construct an high-level semantic representation of the NSU by extracting the relevant information from the dialogue context. To select the right resolution procedure for the given NSU, one needs first to determine its type. That is why the two task are connected. However, they can still be formalized and employed independently.

\section{Motivation}
Non-sentential utterances are interesting in many ways. First of all, they are very frequent in dialogue. According to \citet{Fernandez:2002} and related works, the frequency of NSUs in the dialogue transcripts of the British National Corpus is about 10\% of the total number of utterances. However, this number may vary greatly if one takes into account a larger variety of phenomena or different dialogue domains e.g. \citet{Schlangen:thesis} estimates the frequency of NSUs
to be 20\% of the total number of utterances.

Despite their ubiquity, the semantic content of NSUs is often difficult to extract automatically. Non-sentential utterances are indeed intrinsically dependent on the dialogue context. It is impossible to make sense of them without accessing to the surrounding context. Their high context-dependency makes their interpretation a difficult problem from both a theoretical and computational point of view.

NSUs form a wide range of linguistic phenomena that need to be considered in the formulation of a theory of dialogue context. Only few previous works tackled this problem directly and the majority of them take place in theoretical semantics of dialogue without addressing the possible applications. This means that the interpretation of NSUs is still an understudied problem, making them possibly an even more interesting subject.

\section{Contribution}
Our work follows two parallel paths. On one hand we address the problem of the classification of NSUs by extending the work of \citet{Fernandez:2007}. On the other hand we propose a novel approach to the resolution of NSUs using \textit{probabilistic rules} \citep{Lison2015}.

The classification task is needed to select the resolution procedure but it is nonetheless an independent problem and it can arise in many different situations. Our contribution to this problem is a small but significant improvement over the accuracy of the previous works as well as the exploration of one way to tackle the scarcity of labeled data.

Our work on the resolution of NSUs takes inspiration from \citet{Fernandez:thesis} and \citet{Ginzburg:interactivestance} which provide the theoretical background for our study. Their framework is however purely logic-based therefore it can have some drawbacks in dealing with raw conversational data which often contains hidden or partially observable variables. To this end a probabilistic account of the dialogue state is preferable. In our work we implemented a new approach to NSU resolution based on the probabilistic rules formalism of \citet{Lison2015}. Probabilistic rules are similar, in some way, to the rules formalized by \citet{Ginzburg:interactivestance}, as both express updates on the dialogue state given a set of conditions. However, probabilistic rules can also take into account probabilistic knowledge, thereby making them more suited to deal with the uncertainty often associated with conversational data. Our work does not aim to provide a full theory of NSU resolution but rather be a \textit{proof-of-concept} for the resolution of NSUs via the probabilistic rules formalism. Nevertheless we detail a large set of NSU resolution rules based on the probabilistic rules formalism and provide an actual implementation of a dialogue system for NSU resolution using the OpenDial toolkit \citep{semdial2015_opendial}, which can be the baseline reference for future developments.

Our work for this thesis has produced the following publications:
\begin{itemize}
\item Paolo Dragone and Pierre Lison. Non-Sentential Utterances in Dialogue: Experiments
in classification and interpretation. In: \textit{Proceedings of the 19th workshop on the Semantics and Pragmatics of Dialogue, SEMDIAL 2015 -- goDIAL}, p. 170. G\"oteborg, 2015.
\item Paolo Dragone and Pierre Lison. An Active Learning Approach to the Classification of
Non-Sentential Utterances. In: \textit{Proceedings of the second Italian Conference on Computational Linguistics, CLiC-IT 2015}, in press. Trento, 2015.
\end{itemize}

\section{Outline}

\subsubsection{Chapter 2}
This chapter discusses the background knowledge needed for the development of the following chapters. In particular the chapter describes the concept of non-sentential utterance and the task of interpretation of NSUs with an emphasis on the previous works. Secondly the chapter contains an overview on the formal representation of the dialogue context from the theory of \citet{Ginzburg:interactivestance}. We discuss briefly the Type Theory with Records, the semantic representation of utterances and  the update rules on the dialogue context. Finally, we introduce the probabilistic approach to the definition of the dialogue context from \citet{LisonThesis2014}. We discuss the basics of Bayesian Networks (the dialogue context representation) and the probabilistic rules formalism.

\subsubsection{Chapter 3}
This chapter describes the task of the classification of non-sentential utterances. It provides details on our approach, starting from the replication of the work from \citet{Fernandez:2007} which we use as baseline. We then discuss the extended feature set we used and the semi-supervised learning techniques we employed in our experiments. Lastly we discuss the empirical results we obtained.

\subsubsection{Chapter 4}
This chapter describes the problem of resolving non-sentential utterances and our approach to address it through probabilistic rules. First we formalize the NSU resolution task and describe the theoretical notions needed to address it. We then describe our dialogue context design as a Bayesian network and our formulation for the resolution rules as probabilistic rules. In the end we describe our implementation and an extended example of its application to a real-world scenario.

\subsubsection{Chapter 5}
This is the conclusive chapter of this thesis which summarizes the work and describes possible future works.

\chapter{Background}
\label{ch:background}


\section{Non-Sentential Utterances}
\label{sec:NSU}
From a linguistic perspective, Non-Sentential Utterances – also known as \textit{fragments} – has been historically an umbrella term for many elliptical phenomena that often take place in dialogue.
In order to give a definition of Non-Sentential Utterances ourselves, we shall start by quoting the definition given by \citet{Fernandez:thesis}:
\begin{displayquote}
``In a broad sense, non-sentential utterances are utterances that do
not have the form of a full sentence according to most traditional grammars, but that
nevertheless convey a complete sentential meaning, usually a proposition or a question.''
\end{displayquote}

This is indeed a very general definition, whereas a perhaps simpler approach is taken by \citet{Ginzburg:interactivestance} which defines NSUs as \textquote{utterances without an overt predicate}. The minimal clausal structure of a sentence in English (as in many other languages) is composed of at least a noun phrase and a verb phrase. However, in dialogue the clausal structure is often truncated in favor of shorter sentences that can be understood by inferring their meaning from the surrounding context. We are interested in those utterances that, despite the lack of a complete clausal structure, convey a well-defined meaning given the dialogue context.

The context of an NSU can comprise any variable in the dialogue context but it usually suffice to consider only the \textit{antecedent} of the NSU. The ``antecedent'' of an NSU is the utterance in the dialogue history that can be used to infer its underspecified semantic content. For instance, the NSU in \ref{ex:antecedent} can be interpreted as ``Paul went to his apartment'' by extracting its semantic content from the antecedent. Generally, it is possible to understand the meaning of an NSU by looking at its antecedent.
\begin{example}
\label{ex:antecedent}
\begin{dialogue}
\speak{A} Where did Paul go?
\speak{B} To his apartment.
\end{dialogue}
\end{example}
It is often the case that an NSU and its antecedent present a certain grade of \textit{parallelism}. Usually the meaning of an NSU is associated to a certain aspect of the antecedent. 
As described in \citet{Ginzburg:interactivestance}, the parallelism between an NSU and its antecedent can be of syntactic, semantic or phonological nature. The NSU in \ref{ex:antecedent} presents syntactic parallelism -- the use of ``his'' is syntactically constrained by the fact that Paul is a male individual -- as well as semantic -- the content of an NSU is a location as constrained by the \textit{where} interrogative. This parallelism is one of the properties of NSUs that can be exploited in their interpretation (more details in Chapter \ref{ch:resolution}). Even though it is often the case, the antecedent of an NSU is not always the preceding utterance, especially in multi-party dialogues.

\subsection{A taxonomy of NSUs}
\label{sec:taxonomy}
As we briefly mentioned in Chapter \ref{ch:introduction}, non-sentential utterances come in a large variety of forms. We can categorize NSUs on the basis of their form and their intended meaning. For instance NSUs can be affirmative or negative answers to polar questions, requests for clarification or corrections.

In order to classify the NSUs, we use a taxonomy defined by \citet{Fernandez:2002}. This is a wide-coverage taxonomy resulting from a corpus study on a portion of the British National Corpus \citep{burnard2000reference}.
Table \ref{table:htaxon} contains a summary of the taxonomy with an additional categorization of the classes by their function, as defined by \citet{Fernandez:thesis} then refined by \citet{Ginzburg:interactivestance}.

Other taxonomies of NSUs are available from previous works by e.g. \citet{Schlangen:thesis}, but we opted for the one from \citet{Fernandez:2002} because it has been used in an extensive machine learning experiment by \citet{Fernandez:2007} and it is also used in the theory of \citet{Ginzburg:interactivestance}, which is our reference for the resolution part of our investigation. A detailed comparison of this taxonomy and other ones is given by \citet{Fernandez:thesis}, which also details the corpus study on the BNC that led to the definition of this taxonomy.

\begin{table}[h]
	\centering
	\begin{tabular}{|l l|}
		\hline
		\textbf{Function}                    & \textbf{NSU class}          \\ \hline
		\textit{Positive Feedback}           & Plain Acknowledgment        \\
		                                     & Repeated Acknowledgment     \\ \hline
		\textit{Metacommunicative queries}   & Clarification Ellipsis      \\
		                                     & Check Question              \\
		                                     & Sluice                      \\
		                                     & Filler                      \\ \hline
		\textit{Answers}                     & Short Answer                \\
		                                     & Affirmative Answer          \\
		                                     & Rejection                   \\
		                                     & Repeated Affirmative Answer \\
		                                     & Helpful Rejection           \\
		                                     & Propositional Modifier      \\ \hline
		\textit{Extension Moves}             & Factual Modifier            \\
		                                     & Bare Modifier Phrase        \\
		                                     & Conjunct fragment           \\
		\hline
	\end{tabular}
	\caption{Overview of the classes in the taxonomy, further categorized by their function.}
	\label{table:htaxon}
\end{table}
Follows a brief description of all the classes with some examples. \citet{Fernandez:thesis} provides more details about the rationale of each class.
\subsubsection{Plain Acknowledgment}
Acknowledgments are used to signal understanding or acceptance of the preceding utterance, usually using words or sounds like \textit{yeah}, \textit{right}, \textit{mhm}.
\begin{example}
	\begin{dialogue}
		\speak{A} I shall be getting a copy of this tape.
        \speak{B} \textbf{Right.} \\ \bncref{J42}{71--72}
	\end{dialogue}
\end{example}
\subsubsection{Repeated Acknowledgment}
This is another type of acknowledgement that make use of repetition or reformulation of some constituent of the antecedent to show understanding.
\begin{example}
	\begin{dialogue}
		\speak{A} Oh so if you press enter it'll come down one line.
        \speak{B} \textbf{Enter.} \\ \bncref{G4K}{102--103}
	\end{dialogue}
\end{example}
\subsubsection{Clarification Ellipsis}
These are NSUs that are used to request a clarification of some aspect of the antecedent that was not fully understood.
\begin{example}
	\begin{dialogue}
		\speak{A} I would try F ten.
        \speak{B} \textbf{Just press F ten?} \\ \bncref{G4K}{72--73}
	\end{dialogue}
\end{example}

\subsubsection{Check Question}
Check Questions are used to request an explicit feedback of understanding or acceptance, usually uttered by the same speaker as the antecedent.
\begin{example}
	\begin{dialogue}
		\speak{A} So (\textit{pause}) I'm allowed to record you. \\ \textbf{Okay?}
        \speak{B} Yes. \\ \bncref{KSR}{5--6}
	\end{dialogue}
\end{example}

\subsubsection{Sluice}
Sluices are used for requesting additional information related to or underspecified into the antecedent.
\begin{example}
	\begin{dialogue}
		\speak{A} They wouldn't do it, no.
        \speak{B} \textbf{Why?} \\ \bncref{H5H}{202--203}
	\end{dialogue}
\end{example}

\subsubsection{Filler}
These are fragments used to complete a previous unfinished utterance.
\begin{example}
	\begin{dialogue}
		\speak{A} {[...]} would include satellites like erm
        \speak{B} \textbf{Northallerton.} \\ \bncref{H5D}{78--79}
	\end{dialogue}
\end{example}

\subsubsection{Short Answer}
The NSUs that are typically answers to \textit{wh}-questions.
\begin{example}
	\begin{dialogue}
		\speak{A} What's plus three times plus three?
        \speak{B} \textbf{Nine.} \\ \bncref{J91}{172--173}
	\end{dialogue}
\end{example}

\subsubsection{Plain Affirmative Answer and Plain Rejection}
A type of NSUs used to answer polar questions using \textit{yes}-words and \textit{no}-words.
\begin{example}
	\begin{dialogue}
		\speak{A} Have you settled in?
        \speak{B} \textbf{Yes, thank you.} \\ \bncref{JSN}{36--37}
	\end{dialogue}
\end{example}
\begin{example}
	\begin{dialogue}
		\speak{A} (\textit{pause}) Right, are we ready?
        \speak{B} \textbf{No, not yet.} \\ \bncref{JK8}{137--138}
	\end{dialogue}
\end{example}

\subsubsection{Repeated Affirmative Answer}
NSUs used to give an affirmative answer by repeating or reformulating part of the query.
\begin{example}
	\begin{dialogue}
		\speak{A} \begin{small}You were the first blind person to be employed in the County Council?\end{small}
        \speak{B} \textbf{In the County Council, yes.} \\ \bncref{HDM}{19--20}
	\end{dialogue}
\end{example}

\subsubsection{Helpful Rejection}
Helpful Rejections are used to correct some piece of information from the antecedent.
\begin{example}
	\begin{dialogue}
		\speak{A} Right disk number four?
        \speak{B} \textbf{Three.} \\ \bncref{H61}{10--11}
	\end{dialogue}
\end{example}

\subsubsection{Propositional and Factual Modifiers}
Used to add modal or attitudinal information to the previous utterance. They are usually expressed (respectively) by modal adverbs and exclamatory factual (or factive) adjectives.
\begin{example}
	\begin{dialogue}
		\speak{A} Oh you could hear it?
        \speak{B} \textbf{Occasionally yeah.} \\ \bncref{J8D}{14--15}
	\end{dialogue}
\end{example}
\begin{example}
	\begin{dialogue}
		\speak{A} You'd be there six o'clock gone mate.
        \speak{B} \textbf{Wonderful.} \\ \bncref{J40}{164--165}
	\end{dialogue}
\end{example}

\subsubsection{Bare Modifier Phrase}
Modifiers that behave like non-sentential adjunct modifying a contextual utterance.
\begin{example}
	\begin{dialogue}
		\speak{A} {[...]} then across from there to there.
        \speak{B} \textbf{From side to side.} \\ \bncref{HDH}{377--378}
	\end{dialogue}
\end{example}

\subsubsection{Conjunct}
A Conjunct is a modifier that extends a previous utterance through a conjunction.
\begin{example}
	\begin{dialogue}
		\speak{A} I'll write a letter to Chris
        \speak{B} \textbf{And other people.} \\ \bncref{G4K}{19--20}
	\end{dialogue}
\end{example}

\subsection{The NSU corpus}
\label{sec:corpus}
The taxonomy presented in the previous section is the result of a corpus study on a portion of the dialogue transcripts in the British National Corpus, first started by \citet{Fernandez:2002}, then refined by \citet{Fernandez:thesis}.
The dialogue transcripts used in the corpus study contain both two-party and multi-party conversations. The transcripts cover a wide variety of dialogue domains including free conversation, interviews, seminars and more. \citet{Fernandez:thesis} also describes the annotation procedure and a reliability test. The reliability test was carried out on a subset of the annotated instances comparing the manual  annotation of three annotators. The test showed a good agreement between the annotators with a \textit{kappa}-score of $0.76$. From this test it is also clear that humans can reliably distinguish between the NSU classes in the taxonomy. \citet{Fernandez:thesis} provides more details about the complete analysis of the corpus.

In total about $14\,000$ sentences from $54$ files were examined by the annotators, resulting in a corpus of $1\,299$ NSUs, about $9\%$ of the total of the sentences examined. Of the extracted NSUs, $1\,283$ were successfully categorized according to the defined taxonomy making up a coverage of $98.9\%$. Table \ref{table:nsus_dist} shows the distribution of the classes in the corpus.

\begin{table}[t]
	\centering
	\begin{tabular}{|l r r |}
		\hline
		\textbf{NSU Class}                      &  \textbf{Total} & \textbf{\%} \\ \hline
		Plain Acknowledgment (Ack)              &	 		  599 &        46.1 \\
		Short Answer (ShortAns)                 & 	 		  188 &        14.5 \\
		Affirmative Answer (AffAns)             &	 		  105 &        8.0 \\
		Repeated Acknowledgment (RepAck)        & 	  		   86 &        6.6 \\
		Clarification Ellipsis (CE)             & 	  		   82 &        6.3 \\
		Rejection (Reject)                      & 	  		   49 &        3.7 \\
		Factual Modifier (FactMod)              & 	  		   27 &        2.0 \\
		Repeated Affirmative Answer (RepAffAns) & 	  		   26 &        2.0 \\
		Helpful Rejection (HelpReject)          & 	  		   24 &        1.8 \\
		Check Question (CheckQu)                & 	  		   22 &        1.7 \\
		Sluice                                  & 	  		   21 &        1.6 \\
		Filler                                  & 	  		   18 &        1.4 \\
		Bare Modifier Phrase (BareModPh)        & 	  		   15 &        1.1 \\
		Propositional Modifier (PropMod)        & 	  		   11 &        0.8 \\
		Conjunct (Conj)                         & 	   		   10 &        0.7 \\ \hline
		\textbf{Total}                          & 			 1283 &      100.0 \\
		\hline
	\end{tabular}
	\caption{The distribution of the classes in the NSU corpus.}
	\label{table:nsus_dist}
\end{table}

The annotated instances were also tagged with a reference to the antecedent of the NSU. About $87.5\%$ of annotated NSUs have their immediately preceding utterance as antecedent. \citet{Fernandez:thesis} describes a study of the distance between NSUs and their antecedents, with a comparison between two-party and multi-party dialogues.

\subsection{Interpretation of NSUs}
Due to their incomplete form, non-sentential utterances do not have an exact meaning by themselves. They need to be ``interpreted'' i.e. their intended meaning must be inferred from the dialogue context. One way to interpret NSUs is developed by \citet{Fernandez:thesis}, in turn based on \citet{Schlangen:thesis}, and it is formed by to consecutive steps, namely the \textit{classification} and the \textit{resolution} of the NSUs.
The first step for the interpretation of an NSU is its classification i.e. finding its class according to the taxonomy described in Section \ref{sec:taxonomy}. As demonstrated in \citet{Fernandez:2007}, we can infer the class of an NSU using machine learning, i.e. we can train a classifier on the corpus detailed in Section \ref{sec:corpus} and use it to classify unseen NSU instances.
The type of an NSU is used to determine the right \textit{resolution} procedure to use. The resolution of an NSU is the task of recovering the full clausal meaning from their incomplete form on the basis of contextual information. \citet{Fernandez:thesis} describes a resolution procedure in terms of \textit{rules} that, given some preconditions on the antecedent and other elements of the dialogue states, builds the semantic representation of the NSU. This approach to the resolution of NSUs has been the basis of several implementations of dialogue systems handling the resolution of NSUs such as \citet{Fernandez:shards} and \citet{Purver:2006}.

Extending the interpretation problem to raw conversational data we need also a way to ``detect'' an NSU i.e. decide whether an utterance should be considered as an NSU in the first place. Since this is not our direct concern, we employ in our experiments a simple set of heuristics to distinguish between NSU and non-NSU utterances (see Section \ref{sec:unlabeled_data_extraction}).

\section{A formal model of dialogue}
\label{sec:formal}
As theoretical base of our work we rely on the theory of dialogue context brought up by \citet{Ginzburg:interactivestance}, which presents a grammatical framework expressly developed for dialogue. The claim of \citet{Ginzburg:interactivestance} is that the rules that encode the dynamics of the dialogue have to be built into the grammar itself.
The grammatical framework is formulated using \textit{Type Theory with Records} \citep{cooper2005records}. Type Theory with Records (TTR) is a logical formalism developed to cope with semantics of natural language. TTR is used to build a semantic ontology of abstract entities and events as well as to formalize the \textit{dialogue gameboard} i.e. a formal representation of the dialogue context and its rules. The evolution of the conversation is formalized by means of \textit{update rules} on the dialogue context. \citet{Ginzburg:interactivestance} also accounts for NSUs and provides a set of dedicated rules.

\subsection{Type Theory with Records}
We will now briefly introduce the basic notions of the Type Theory with Records (TTR), with just enough detail needed by to understand the following sections, referring to \citet{Ginzburg:interactivestance} for a complete description.

In TTR, objects can be of different types. The statement $x : T$ is a \textit{typing judgment}, indicating that the object $x$ is of type $T$. If $x$ is of type $T$, $x$ is said to be a \textit{witness} of $T$. Types can either be \textit{basic} (atomic) such as IND\footnote{The type IND stands for a generic ``individual''.} or \textit{complex} i.e. dependent on other objects or types such as $\text{drive}(x,y)$. Types also include constructs such as lists, sets and so on.
Other useful constructs are records and record types. A record contains a set of assignments between labels and values whereas a record type contains a set of judgments between labels and types:
\begin{align*}
	r:\ \  \begin{bmatrix*}[l]
       l_1 &=& v_1 \\
       l_2 &=& v_2 \\
	   &\dots \\       
       l_n &=& v_n \\
    \end{bmatrix*} && \rho: \ \ 
    \begin{bmatrix*}[l]
       l_1 &:& T_1 \\
       l_2 &:& T_2 \\
	   &\dots \\       
       l_n &:& T_n \\
    \end{bmatrix*} &
\end{align*}
The record $r$ is of record type $\rho$ if and only if $v_1\!:\!T_1\ \land\ v_2\!:\!T_2\ \land\ \dots\ \land\ v_n\!:\!T_n$. Typing judgment can be used to indicate the record $r$ being of record type $\rho$ as $r:\rho$.

TTR also provides \textit{function types} of the form $T_1 \rightarrow T_2$ which maps records of type $T_1$ to records of type $T_2$. Functional application is indicated as $(x:T_1).T_2$.

\subsubsection{Utterance representation}
\label{sec:utt_repr}
At the basis of the grammatical framework of \citet{Ginzburg:interactivestance} lies the notion of \textit{proposition}. Propositions are entities used to represent facts, events and situations as well as to characterize the communicative process. In TTR propositions are records of the type:
\begin{align*}
	\text{Prop} = \begin{bmatrix*}[l]
       \text{sit} &:& \text{Record} \\
       \text{sit-type} &:& \text{RecType} \\
    \end{bmatrix*} 
\end{align*}

A simple example of proposition may be the following:
\begin{align*}
	& \textit{Paul drives a car.} \\
	&\begin{bmatrix*}[l]
       \text{sit} &=& r_1 \\
       \text{sit-type} &=& \begin{bmatrix*}[l]
	   \text{x} &:& \text{IND} \\
	   \text{p}_1 &:& \text{named}(\text{x}, \text{Paul}) \\
	   \text{y} &:& \text{IND} \\
	   \text{p}_2 &:& \text{car}(\text{y}) \\
       \text{c} &:& \text{drive}(\text{x}, \texttt{y}) \\
    \end{bmatrix*}
    \end{bmatrix*}
\end{align*} \\\\
On the other hand, questions are represented as propositional abstracts i.e. functions from the \textit{question domain} $\Delta_q$ to propositions. Following the definition of \citet{Fernandez:thesis}: $$\text{Question} = \Delta_q \rightarrow \text{Prop}$$
The question domain $\Delta_q$ is a record type containing the \textit{wh}-restrictors of the question $q$.\footnote{\citet{Ginzburg:interactivestance} extends this field to be a list of record types to take into account situations with multiple question domains.}
The \textit{wh}-restrictors are record types that characterize the necessary information needed to resolve a \textit{wh}-question e.g. for a  \textit{where} interrogative the answer must be a place instead for a \textit{when} interrogative it must be a time. Clearly, the right \textit{wh}-restrictor depends on the \textit{wh}-interrogative used. Consider the following example of a \textit{wh}-question:
\begin{align*}
	& \textit{Who drives?} \\
	&\left(\text{r} : \begin{bmatrix*}[l]
       \text{x} &:& \text{IND} \\
       \text{rest} &:& \text{person}(\text{x}) \\
    \end{bmatrix*}\right).
    \begin{bmatrix*}[l]
       \text{sit} &=& r_1 \\
       \text{sit-type} &=& \begin{bmatrix*}[l]
       		\text{c} &:& \text{drive}(\text{r.x}) \\
    	\end{bmatrix*} \\
    \end{bmatrix*}
\end{align*}
Here the question domain of the \textit{who} interrogative is an individual $\text{x}$ that is a person.

Polar questions, i.e. bare yes/no-questions, are represented as propositional abstract as the \textit{wh}-questions, with the difference that their question domain is an empty record type. An example of polar question:
\begin{align*}
	& \textit{Does Paul drive?} \\
	&\left(\text{r} : \begin{bmatrix*}[l]
		& \\
    \end{bmatrix*}\right).
    \begin{bmatrix*}[l]
       \text{sit} &=& r_1 \\
       \text{sit-type} &=& \begin{bmatrix*}[l]
      	 \text{x} &:& \text{IND} \\
	  	 \text{p}_1 &:& \text{named}(\text{x}, \text{Paul}) \\
       	 \text{c} &:& \text{drive}(\text{x}) \\
    	\end{bmatrix*} \\
    \end{bmatrix*}
\end{align*}

A special type of propositions are used to represent the content of conversational moves which need to take into a account the relation that stands between the speaker, the addressee and the content of the move. Those are called \textit{illocutionary propositions} (of type IllocProp) and the relation that they contain is called \textit{illocutionary relation}\footnote{Also called \textit{illocutionary act} or \textit{dialogue act}.}. Illocutionary relations indicates the \textit{function} of a proposition, such as ``Assert'', ``Ask'', ``Greet''. For a proposition $p$, the illocutionary proposition that holds $p$ as its content can be indicated as $\text{R}(\text{spkr}, \text{addr}, p)$, where R is the illocutionary relation, spkr and addr refer respectively to the speaker and the addressee\footnote{For brevity only the semantic content of the illocutionary proposition is shown here.}.
Examples of illocutionary propositions are:
\begin{align*}
	& \text{Assert(\text{spkr} : \text{IND}, \text{addr} : \text{IND}, \textit{p} : \text{Prop})} \\
	& \text{Ask(\text{spkr} : \text{IND}, \text{addr} : \text{IND}, \textit{q} : \text{Question})}
\end{align*}

\subsection{The dialogue context}
\label{sec:dialogue_context}
In \citet{Ginzburg:interactivestance}, the dialogue context – also known as the Dialogue Gameboard (DGB) – is a formal representation that describes the current state of the dialogue. It includes a wide range of variables needed to handle different aspects of the dialogue. However, we concentrated on the most basic ones:
\begin{itemize}
\item Facts, a set of known facts;
\item LatestMove, the latest move made in the dialogue;
\item QUD, a partially ordered set of questions under discussion.
\end{itemize}
The DGB can be represented in TTR as a record in the following way:
\begin{equation*}
	\begin{bmatrix*}[l]
       \text{Facts} &:& \text{Set(Prop)} \\
       \text{LatestMove} &:& \text{IllocProp} \\
       \text{QUD} &:& \text{poset(Question)} \\
    \end{bmatrix*}
\end{equation*}

The elements in the DGB represent the \textit{common ground} of the conversation, shared between all the participants. In this representation we abstracted away several details that would be included in the actual DGB presented by \citet{Ginzburg:interactivestance} such as the fields to track who is holding the turn, the current time and so on.
We now detail the basic variables of the DGB.

\subsubsection{Facts}
Facts is a set of known facts, shared by all the conversational participants. The elements of Facts are propositions, which are assumed to be sufficient to encode the knowledge of the participants within the context of the dialogue. The Facts encode all the records that are \textit{accepted} by all participants, i.e. facts that will not raise issues in the future development of the conversation. A complementary problem that we marginally address is the understanding -- or \textit{grounding} -- of a sentence. \citet{Ginzburg:interactivestance} develops a comprehensive theory of grounding but we do not include it in our work.

\subsubsection{LatestMove}
Dialogue utterances are made of coherent responses to the preceding utterances, that is why it is important to keep track of the history of the dialogue. In a two-party dialogue it is usually the case that the current utterance is a response to previous one, instead in a multi-party dialogue can be useful to keep track of a larger window of the dialogue history.
\citet{Ginzburg:interactivestance} keeps track of the history of the dialogue within the variable Moves while a reference to the latest (illocutionary) proposition is recorded in the field LatestMove.

\subsubsection{QUD}
QUD is a set of \textit{questions under discussion}. In a general sense, a ``question under discussion'' represents an issue being raised in the conversation which drives the future discussion. Despite the name, QUDs may arise from both questions and propositions. 

\citet{Ginzburg:interactivestance} defines QUD as a partially ordered set (poset). Its ordering determines the priority of the issues to be resolved. Of particular importance is the first element in the set according to the defined ordering which is taken as the topic of discussion of the subsequent utterances until it is \textit{resolved}. Such element is referred to as MaxQUD.

The formalization of the ordering is a rather complex matter in a generic theory of context that needs to account for the beliefs of the participants and it is especially problematic when dealing with multi-party dialogues.
The usage of QUD is of particular importance in our case because the MaxQUD is used as the antecedent in the interpretation of NSUs.

\subsection{Update rules}
\label{sec:updaterules}
The dynamics of the DGB are defined by a set of \textit{update rules} – also called \textit{conversational rules} – which are applied on the DGB throughout the course of the conversation. Update rules are formalized as a set of effects on the parameters of the DGB given that certain preconditions hold. An update rule can be represented in the following way:
\begin{equation*}
	\begin{bmatrix*}[l]
       \text{pre} &:& \begin{bmatrix*}[c] \dots \\ \end{bmatrix*} \\
       \text{effects} &:& \begin{bmatrix*}[c] \dots \\ \end{bmatrix*} \\
    \end{bmatrix*}
\end{equation*}
where both \textit{pre} and \textit{effects} are subsets of the parameters of the DGB and they respectively represent the necessary conditions for the application of the rule and the values of the involved variables right after the application of the rule.

\citet{Ginzburg:interactivestance} defines all sorts of rules needed to handle a great variety of conversational protocols.
Rules that are particularly interesting with respect to our work are those that handle queries and assertions as well as the ones that describe the dynamics of QUD and Facts.

The following rule describes how QUD is incremented when a question is posed:
\begin{equation*}
	\begin{bmatrix*}[l]
       \text{pre} &:& \begin{bmatrix*}[l]
							\text{q} &:& \text{Question} \\
							\text{LatestMove} = \text{Ask}(\text{spkr}, \text{addr}, \text{q}) &:& \text{IllocProp} \\
       					\end{bmatrix*} \\
       \text{effects} &:& \begin{bmatrix*}[l]
       							\text{qud} = \langle\, \text{q},\  \text{pre.qud}\, \rangle &:& \text{poset}(\text{Question}) \\
       						\end{bmatrix*} \\
    \end{bmatrix*}
\end{equation*}
As argued above, issues are also raised by assertions, as realized by the following rule:
\begin{equation*}
	\begin{bmatrix*}[l]
       \text{pre} &:& \begin{bmatrix*}[l] 
							\text{p} &:& Prop \\
							\text{LatestMove} = \text{Assert}(\text{spkr}, \text{addr}, \text{p}) &:& \text{IllocProp} \\
       					\end{bmatrix*} \\
       \text{effects} &:& \begin{bmatrix*}[l]
       							\text{qud} = \langle\, \text{p?},\  \text{pre.qud}\, \rangle &:& \text{poset}(\text{Question}) \\
       						\end{bmatrix*} \\
    \end{bmatrix*}
\end{equation*}

The act of answering to a question is nothing else than asserting a proposition that resolves such a question. As a consequence the other speaker can either raise another issue related to the previous one or accept the fact that the issue has been resolved.
The acceptance move is realized in the following way:
\begin{equation*}
	\begin{bmatrix*}[l]
       \text{pre} &:& \begin{bmatrix*}[l] 
							\text{p} &:& Prop \\
							\text{LatestMove} = \text{Assert}(\text{spkr}, \text{addr}, \text{p}) &:& \text{IllocProp} \\
							\text{qud} = \langle\, \text{p?},\  \text{pre.qud}\, \rangle &:& \text{poset}(\text{Question}) \\
       					\end{bmatrix*} \\
       \text{effects} &:& \begin{bmatrix*}[l]
       							\text{spkr} = \text{pre.addr} &:& \text{Ind} \\
       							\text{addr} = \text{pre.spkr} &:& \text{Ind} \\
       							\text{LatestMove} = \text{Accept}(\text{spkr},\text{addr}, \text{p}) &:& \text{IllocProp} \\
       						\end{bmatrix*} \\
    \end{bmatrix*}
\end{equation*}
The speaker can also query the addressee with a \textit{Check} move in order to ask for an explicit acknowledgment (\textit{Confirm}) to a question-resolving assertion\footnote{The rules for the \textit{Check} and \textit{Confirm} moves are omitted for brevity.}.
Acceptance and confirmation lead to an update of Facts and to a ``downdate'' of the QUD i.e. the removal of the resolved questions in QUD:
\begin{equation*}
	\label{eq:fact-update}
	\begin{bmatrix*}[l]
       \text{pre} &:& \begin{bmatrix*}[l] 
							\text{p} &:& Prop \\
							\text{LatestMove} = \text{Accept}(\text{spkr}, \text{addr}, \text{p})\ \lor \\ 
							\quad \text{Confirm}(\text{spkr}, \text{addr}, \text{p}) &:& \text{IllocProp} \\
							\text{qud} = \langle\, \text{p?},\  \text{pre.qud}\, \rangle &:& \text{poset}(\text{Question}) \\
       					\end{bmatrix*} \\
       \text{effects} &:& \begin{bmatrix*}[l]
       							\text{facts} = \text{pre.facts} \cup \{\,\text{p}\,\} &:& \text{Set}(\text{Prop}) \\
       							\text{qud} = \textit{NonResolve}(\text{pre.qud},\text{facts}) &:& \text{poset}(\text{Question}) \\
       						\end{bmatrix*} \\
    \end{bmatrix*}
\end{equation*}

While QUD represents the unresolved issues that have been introduced in the dialogue, Facts contains all the issues that have been resolved instead. That is why their update rules are closely related. The function \textit{NonResolve} in the above rule checks for any resolved issues by the just updated facts and leave the unresolved ones into QUD.

\section{Probabilistic modeling of dialogue}
\label{sec:prob_model}
In the previous section we detailed a logic-based model of dialogue from \citet{Ginzburg:interactivestance}. Another possible approach to dialogue modeling relies on probabilistic models to encode the variables and the dynamics of the dialogue context. Arguably this approach can be considered more robust to the intrinsic randomness present in dialogue. This is partially the reason why we explored this strategy as well as other advantages that will be discussed in Chapter \ref{ch:resolution}.

We based our work on the \textit{probabilistic rules} formalism developed by \citet{Lison:2012}. This formalism is particularly suited for our purpose because of their commonalities with the update rules described in Section \ref{sec:updaterules}. The probabilistic rules formalism is based on the representation of the dialogue state as a Bayesian network. 
In this section we briefly describe how Bayesian networks are structured, then we detail the probabilistic rules formalism that we employ in Chapter \ref{ch:resolution} to model the resolution of the NSUs. 


\subsection{Bayesian Networks}
Bayesian networks are probabilistic graphical models\footnote{A type of probabilistic models represented by graphs.} representing a set of random variables (nodes) and their conditional dependency relations (edges). A Bayesian network is a directed acyclic graph i.e. a direct graph that does not contain cycles (two random variables cannot be mutually dependent).
Given the random variables $X_1,\dots,X_n$ in a Bayesian network, we are interested in the \textit{joint probability distribution} $P(X_1,\dots,X_n)$ of those variables. In general, the size of the joint distribution is exponential in the number $n$ of variables therefore it is difficult to estimate when $n$ grows. In the case of Bayesian networks we can exploit the \textit{conditional independence} to reduce the complexity of the joint distribution. Given three random variables $X$, $Y$ and $Z$, $X$ and $Y$ are said to be conditionally independent given $Z$ if and only if (for all combinations of values) $P(X,Y|Z) = P(X|Z) P(Y|Z)$.
We can define for a variable $X_i$ in $X_1,\dots,X_n$ the set $parents(X_i)$ such that if there is a direct edge from $X_j$ to $X_i$ then $X_j \in parents(X_i)$.
Given a \textit{topological ordering}\footnote{A topological ordering is an ordering of the nodes such that for every two nodes $u$ and $v$ connected by a directed edge from $u$ to $v$, $u$ appears before $v$ in such ordering. A topological ordering can only be defined on directed acyclic graphs.} of the variables (nodes) of the Bayesian network, a variable $X_i$ is conditionally independent from all its predecessor that are not in $parents(X_i)$ therefore the joint probability distribution can be defined as follows: $$P(X_1,\dots,X_n) = \prod_{i=1}^n P(X_i|parents(X_i))$$
For each variable $X_i$, $P(X_i|parents(X_i))$ is the \textit{conditional probability distribution} (CPD) of $X_i$. 
The CPDs together with the directed graph fully determine the joint distribution of the Bayesian network.

The network can be used for inference by querying the distribution of a subset of variables, usually given some evidence. Given a subset of variables $\mathbf{Q} \subset \mathbf{X}$ and an assignment of values $\mathbf{e}$ of the evidence variables, the query is the posterior distribution $P(\mathbf{Q}|\mathbf{e})$. To compute the posterior distribution one needs an inference algorithm. Such algorithm can be exact – such as the \textit{variable elimination} algorithm \citep{zhang1996exploiting} – or approximate – such as the \textit{loopy belief propagation} algorithm \citep{murphy1999loopy}.

The distributions of the single variables can be learned from observed data using \textit{maximum likelihood estimation} or \textit{Bayesian learning}.

\subsection{Probabilistic rules}
\label{sec:prob_rules}
The probabilistic rules formalism is a domain-independent dialogue modeling framework. Probabilistic rules are expressed as \textit{\textbf{if} \dots\ \textbf{then} \dots\ \textbf{else} \dots} constructs mapping logical conditions on the state variables to effects encoded by either probability distributions or utility functions. The former are called \textit{probability rules} while the latter are \textit{utility rules}. While we make use of both types of rules in our work, here we concentrate only on the probability rules which are the ones used for the resolution of the NSUs.


Let $c_1, \dots, c_n$ be a sequence of logical conditions and $P(E_1), \dots, P(E_n)$ a sequence of categorical probability distributions\footnote{A categorical distribution is a probability distribution of an event having a finite set of outcomes with defined probability.} over mutually exclusive effects. A probability rule $r$ is defined as follows:
\begin{flalign*}
\ \ \ \ 
& r: \\
& \ \ \ \ \forall \mathbf{x} \\
& \ \ \ \ 
\textbf{if} \ c_1 \  \textbf{then} & \\
& \ \ \ \ \;\;\;\;\; 
\begin{cases}
	P(E_1 = e_{1,1}) = p_{1,1} \\
	\dots \\
	P(E_1 = e_{1,m_1}) = p_{1,m_1} \\
\end{cases} \\
& \ \ \ \ 
\textbf{else if} \ c_2 \  \textbf{then} & \\
& \ \ \ \ \;\;\;\;\; 
\begin{cases}
	P(E_2 = e_{2,1}) = p_{2,1} \\
	\dots \\
	P(E_2 = e_{2,m_2}) = p_{2,m_2} \\
\end{cases} \\
& \ \ \ \ 
\textbf{else} & \\
& \ \ \ \ \;\;\;\;\; 
\begin{cases}
	P(E_n = e_{n,1}) = p_{n,1} \\
	\dots \\
	P(E_n = e_{n,m_n}) = p_{n,m_n} \\
\end{cases} \\
\end{flalign*}
The random variable $E_i$ encodes a range of possible effects $e_{i,1},\dots,e_{i,m_i}$, each one with a corresponding probability $p_{i,j}$. The conditions and effects of a rule may include underspecified variables, denoted with \textbf{x}, which are universally quantified on the top of the rule. The effects are duplicated for every possible assignments (grounding) of the underspecified variables.

Each pair of condition and probability distribution over the effects $\langle c_i,P(E_i)\rangle$ is a branch $br_i$ of the rule. Overall, the rule is a sequence of branches $br_1,\dots,br_n$. The rule is ``executed'' by running sequentially through the branches. Only the first condition satisfied triggers the corresponding probabilistic effect, the subsequent branches are ignored (as in programming languages).

The dialogue state is represented as a Bayesian network containing a set of nodes (random variables). At each state update, rules are instantiated as nodes in the network. For each rule, the input edges of the node come from the condition variables whereas the output edges go towards the effect variables. The probability distribution of the rule is extracted by executing it. The probability distribution of the effect variables are then retrieved by probabilistic inference. \citet{LisonThesis2014} details the rules and update procedure.

The probabilistic rules are useful in at least three ways:
\begin{itemize}
\item They are expressly designed for dialogue modeling. They combine the expressivity of both probabilistic inference and first order logic. This is an advantage in dialogue modeling where one has to describe objects that relate to each other in the dialogue domain and, at the same time, handle uncertain knowledge of the state variables.
\item They can cope with the scarcity of training data of most dialogue domains by exploiting the \textit{internal structure} of the dialogue models. By using logical formulae to encode the conditions for a possible outcome, it is possible to group the values of the variables into \textit{partitions}, reducing the number of parameters needed to infer the outcome distribution and therefore the amount of data needed to learn the distribution.
\item The state update is handled with probabilistic inference therefore they can operate under uncertain settings which is often needed in dialogue modeling where variables are best represented as \textit{belief states}, continuously updated by observed evidence.
\end{itemize}

The probabilistic rules formalism has also been implemented into a framework called OpenDial \citep{semdial2015_opendial}. OpenDial is a Java toolkit for developing spoken dialogue systems using the probabilistic rules formalism. Using an XML-based language it is possible to define in OpenDial the probabilistic rules to handle the evolution of the dialogue state in a domain-independent way. OpenDial can either work on existent transcripts or as an interactive user interface. OpenDial can also learn parameters from small amounts of data using either supervised or reinforcement learning.

\section{Summary}

In this chapter we discussed the background knowledge needed for describing our work on non-sentential utterances.
We first described the notion of non-sentential utterances and the problem of interpreting them. We showed how those utterances can be categorized with a taxonomy from \citet{Fernandez:2002}. We described how the interpretation of non-sentential utterances can be addressed by first classifying them using the aforementioned taxonomy and then applying some kind of ``resolution'' procedure to extract their meaning from the dialogue context. In Chapter \ref{ch:classification} we will address the NSU classification problem on the basis of the experiments from \citet{Fernandez:2007}. In Chapter \ref{ch:resolution} instead we will address the NSU resolution task. \citet{Fernandez:thesis} describes a set of NSU resolution rules rooted in a TTR representation of the dialogue context. Section \ref{sec:formal} briefly described the TTR notions we employed as well as the dialogue context theory based on TTR from \citet{Ginzburg:interactivestance}.

At last we described the probabilistic modeling of dialogue from \citet{LisonThesis2014} based on the probabilistic rules formalism. As we mentioned in Chapter \ref{ch:introduction} this formalism is the framework for our formulation of the NSU resolution rules on the basis of the one developed by \citet{Fernandez:thesis}. We described in Section \ref{sec:prob_model} the basic notion of Bayesian networks which is the representation of the dialogue state employed by the probabilistic rules formalism. Finally, in Section \ref{sec:prob_rules} we explained the probabilistic rules formalism itself and its advantages.

\chapter{Classification of Non-Sentential Utterances}
\label{ch:classification}
Non-sentential utterances are pervasive dialogue phenomena. Any dialogue processing application (e.g. parsing or machine translation of dialogues, or interactive dialogue systems) has to take into account the presence of NSUs and deal with them. As described in Section \ref{sec:NSU}, the NSUs come in a great variety of forms that must be treated differently from one another. To this end, the most basic (and perhaps useful) task is classifying them. In our work we employ the taxonomy and the corpus described in Sections \ref{sec:taxonomy} and \ref{sec:corpus}. As demonstrated by \citet{Fernandez:2007}, we can use machine learning techniques to automatically classify a given NSU, using the annotated corpus as training data. \citet{Fernandez:2007} is our main theoretical reference and (to our knowledge) the state-of-the-art in performance for the task of classification of NSUs. We first replicated the approach of the aforementioned work and used it as a benchmark for our experiments. Secondly, we tried to improve the classification performances, starting from an expansion of the feature set, then employing semi-supervised learning to address the scarcity of labeled data.

\section{The data}
\label{sec:data}
The corpus from \citet{Fernandez:2007} contains $1\,283$ annotated NSU instances, each one identified by the name of the containing BNC file and their sentence number, a sequential number to uniquely identify a sentence in a dialogue transcript. The instances are also tagged with the sentence number of their antecedent which makes up the context for the classification. The raw utterances can be retrieved from the BNC using this information.

For the classification task, we make the same simplifying restriction on the corpus made by \citet{Fernandez:2007}, that is to consider only the NSUs whose antecedent is their preceding sentence. This assumption facilitates the feature extraction procedure without reducing significantly the size of the dataset (about $12\%$ of the total). The resulting distribution of the NSUs after the restriction is showed in Table \ref{table:nsus_rest_dist}.

As one can see from Table \ref{table:nsus_rest_dist}, the distribution of the instances is quite skewed, largely in favor of some classes than others. Moreover very frequent classes are usually the easiest to classify, leaving the most difficult ones with few instances as examples for the classifiers.
Although the scarcity of the training material and the imbalance of the classes are the two major problems for the classification task, we propose a set of methods to address them, as described in the following sections.

\begin{table}[t]
	\centering
	\begin{tabular}{|l r|}
		\hline
		\textbf{NSU class}                      &  \textbf{Total} \\ \hline
		Plain Acknowledgment (Ack)              &	 		  582 \\
		Short Answer (ShortAns)                 & 	 		  105 \\
		Affirmative Answer (AffAns)             &	 		  100 \\
		Repeated Acknowledgment (RepAck)        & 	  		   80 \\
		Clarification Ellipsis (CE)             & 	  		   66 \\
		Rejection (Reject)                      & 	  		   48 \\
		Repeated Affirmative Answer (RepAffAns) & 	  		   25 \\		
		Factual Modifier (FactMod)              & 	  		   23 \\
		Sluice                                  & 	  		   20 \\		
		Helpful Rejection (HelpReject)          & 	  		   18 \\		
		Filler                                  & 	  		   16 \\		
		Check Question (CheckQu)                & 	  		   15 \\
		Bare Modifier Phrase (BareModPh)        & 	  		   10 \\
		Propositional Modifier (PropMod)        & 	  		   10 \\
		Conjunct (Conj)                         & 	   		    5 \\ \hline
		\textbf{Total}                          & 			 1123 \\
		\hline
	\end{tabular}
	\caption{Distribution of the classes in the corpus after the simplifying restriction.}
	\label{table:nsus_rest_dist}
\end{table}

\subsubsection{The British National Corpus}
The British National Corpus \citep{burnard2000reference} – BNC for short – is a collection of spoken and written material, containing about 100 million words of (British) English texts from a large variety of sources. Among the others, it contains a vast selection of dialogue transcripts covering a wide range of domains.
Each dialogue transcript in the BNC is contained in an XML file along with many details about the dialogue settings. The dialogues are structured following the CLAWS tagging system \citep{Garside:1993} which segmented the utterances both at word and sentence level. The word units contains both the raw text, the corresponding lemma (headword) and the POS-tag according to the \textit{C5} tagset \citep{Leech:1994}. Each sentence is identified by an unique ID number within the file.
Sentences can also contain information about the pauses and the unclarities.
The sentences are sorted in their order of appearance and include additional information about temporal alignment in case of overlapping.

\section{Machine learning algorithms}
\label{sec:ml_alg}
We employ two different supervised learning algorithms: decision trees and support vector machines. The former are used mainly as a comparison with \citet{Fernandez:2007} which employ this algorithm as well.
For parameters tuning we implemented a coordinate ascent algorithm. As a framework for our experiments we rely on the Weka toolkit \citep{Weka}, a Java library containing the implementation of many machine learning algorithms as well as a general-purpose machine learning API.

\subsection{Classification: Decision Trees}
\label{sec:entropy}
We employ the \textit{C4.5} aglorithm \citep{Quinlan:1993} for decision tree learning. Weka contains an implementation of this algorithm called J48. The goal of decision tree learning is to create a predictive model from the training data. The construction of the decision tree is performed by splitting the training set into subsets according to the values of an attribute. This process is then repeated recursively on each subset. The construction algorithm is usually an informed search using some kind of heuristics to drive the choice of the splitting attribute. In the case of C4.5 the metric used for the attribute choice is the expected \textit{information gain}. The information gain is based on the concept of \textit{entropy}.

In information theory, the entropy \citep{shannon:1948} is the expected value of information carried by a message (or an event in general). It is also a measure of the ``unpredictability'' of an event. The more unpredictable an event is, the more information it provides when it occurs. Formally, the entropy of a random variable $X$ is
$$H(X) = - \sum_i P(x_i) \log P(x_i)$$
where $P(x_i)$ is the probability of the $i$-th value of the variable $X$.
A derived notion is the \textit{conditional entropy} of a random variable $Y$ knowing the value of another variable $X$:
\begin{align*}
H(Y|X) &= \sum_i P(x_i)\,H(Y|X\!=\!x_i) \\
&= \sum_i P(x_i) \sum_j P(y_j|x_i) \, \log P(y_j|x_i)
\end{align*}
where $x_i$ are the values of the variable $X$ and $y_j$ are the value of the variable $Y$.

For the decision tree construction, the \textit{information gain} of an attribute $A$ is the reduction of the entropy of the class $C$ gained by knowing the value of $A$:
$$IG(A,C) = H(C) - H(C|A)$$
The attribute with the highest information gain is used as splitting attribute.

\subsection{Classification: Support Vector Machines}
\label{sec:SMO}
The Support Vector Machines (SVMs) \citep{Boser:1992} is one of the most studied and reliable family of learning algorithms. An SVM is a binary classifier that uses a representation of the instances as points in a $m$-dimensional space, where $m$ is the number of attributes. Assuming that the instances of the two classes are linearly separable\footnote{An hyperplane can be drawn in the space such that the instances of one class are all in one side of the hyperplane and the instances of the other class are in the other side.}, the goal of SVMs is to find an hyperplane that separates the classes with the maximum margin. The task of finding the best hyperplane that separates the classes is defined as an optimization problem. SVMs can also be formulated to have ``soft margins'' i.e. allowing some points of a class to lay in the opposite side of the hyperplane in order to find a better solution. The SVM algorithm we use regularizes the model through a single parameter \textit{C}. 

The SVMs can also be used with non-linear (i.e. non linearly separable) data using the so called \textit{kernel method}. A kernel function maps the points from the input space into an high-dimentional space where they might be linearly separable. A popular kernel function is the (Gaussian) Radial Basis Function (RBF) which maps the input space into an infinite-dimensional space. Its popularity is partially due to the simplicity of its model which involves only one parameter $\gamma$. 

Even though SVMs are defined as binary classifiers, they can be extended to a multi-class scenario by e.g. training multiple binary classifiers using a one-vs-all or a one-vs-one classification strategy \citep{Duan:2005}.

The Weka toolkit contains an implementation of SVMs that uses the Sequential Minimal Optimization (SMO) algorithm \citep{Platt:1998}. In all our experiments we use the SMO algorithm with an RBF kernel. 

\subsection{Optimization: Coordinate ascent}
The parameter tuning of all our experiments is carried out automatically through a simple \textit{coordinate ascent}\footnote{Also known as coordinate descent which is the minimization counterpart (distinguished only by changing the sign of the function).} optimization algorithm.
Coordinate ascent is based on the idea of maximizing a multivariable function $f(\mathbf{X})$ along one direction at a time, as apposed to e.g. gradient descent which follows the direction given by the gradient of the function.
Our implementation detects the ascent direction by lookup of the function value.
The Algorithm \ref{alg:maximize} contains a procedure to maximize a function $f$ along the direction $k$ while Algorithm \ref{alg:coordinate_ascent} performs the coordinate ascent.
The step-size values decay at a rate given by the coefficient $\alpha$. The minimum step-sizes determine the stopping conditions for the \textit{maximize} function, instead the \textit{coordinateAscent} algorithm stops as soon as the found values do not change between two iterations therefore the maximum is found. The latter algorithm can be easily modified to account for a stopping condition given by a maximum number of iterations.

We implemented this algorithm ourselves because, for technical reasons, it was easier than rely on a third-party API. Its simplicity is one of its advantages but it is more prone to be stuck on local maximums than more sophisticated techniques such as gradient ascent.

For all our experiments we use the described algorithm to find the parameters that yield the maximum accuracy of the classifiers (using 10-fold cross-validation). For the SMO algorithm we optimize the parameters \textit{C} and $\gamma$, whereas for the J48 algorithm we optimize the parameters \textit{C} (confidence threshold for pruning) and \textit{M} (minimum number of instances per leaf).

\IncMargin{2em}
\begin{algorithm}[h]
 \label{alg:maximize}
 \caption{$maximize(f, k, \mathbf{X}, \delta_k, m_k)$}

 \Indm 
 \KwIn{Function $f$ to be maximized; 
 index $k$ of the parameter to maximize; 
 vector $\mathbf{X}$ of the current parameter values; 
 initial step-size value $\delta_k$ for the $k$-th parameter; 
 minimum step-size value $m_k$ for the $k$-th parameter}
 \KwOut{The value of the $k$-th parameter that maximizes the function $f$ along the corresponding direction}
 \Indp

 $y_{max} \leftarrow f(\mathbf{X})$\;
 
 \While{$|\delta_k| \geq m_k$}{
  $\mathbf{\hat{X}} \leftarrow \mathbf{X}$\;
  $\mathbf{\hat{X}}[k] \leftarrow \delta_k \times \mathbf{\hat{X}}[k]$\;
  $\hat{y} \leftarrow f(\mathbf{\hat{X}})$\;
  \eIf{$\hat{y} > y_{max}$}{
   $y_{max} \leftarrow \hat{y}$\;
   $\mathbf{X}[k] \leftarrow \mathbf{\hat{X}}[k]$\;
   }{
   $\delta_k \leftarrow -\delta_k$\;
   }
   $\delta_k \leftarrow \alpha \times \delta_k$\;
 }
 \Return $\mathbf{X}[k]$\;
\end{algorithm}
\DecMargin{2em}

\IncMargin{2em}
\begin{algorithm}[h]
 \label{alg:coordinate_ascent}
 \caption{$coordinateAscent(f, n, \mathbf{X}, \mathbf{\Delta}, \mathbf{m})$}

 \Indm 
 \KwIn{Function $f$ to be maximized;
 number $n$ of parameters of the function;
 vector $\mathbf{X}$ of initial parameter values; 
 vector $\mathbf{\Delta}$ of initial step-size values;
 vector $\mathbf{m}$ of minimum step-sizes;}
 \KwOut{The vector $\mathbf{X}$ that maximizes the function $f$}
 \Indp
 initialize $\mathbf{X}_{last}$ to random values (different from $\mathbf{X}$)\;
 \While{$\mathbf{X} \neq \mathbf{X}_{last}$}{
  $\mathbf{X}_{last} \leftarrow \mathbf{X}$\;
  \For{$k \in 1 \dots n$}{
   $\delta_k \leftarrow \mathbf{\Delta}[k]$\;
   $m_k \leftarrow \mathbf{m}[k]$\;
   $\mathbf{X}[k] \leftarrow maximize(\mathit{f}, k, \mathbf{X}, \delta_k, m_k)$\;
   }
 }
 \Return $\mathbf{X}$\;
\end{algorithm}
\DecMargin{2em}

\section{The baseline feature set}
\label{sec:baseline}
Our baseline is set to be the replicated approach of the classification experiments carried out by \citet{Fernandez:2007}, which is our reference work for our study. It contains two experiments, one with a restricted set of classes (leaving out Acknowledgments and Check Questions) and a second taking into account all classes. We are interested in the latter although the former is useful to understand the problem and to analyze the results of our classifier. The aforementioned paper also contains an analysis of the results and the feature contribution which proved useful in the replication of the experiments.
For our baseline we use only the features they describe. The feature set is composed of $9$ features exploiting a series of syntactic and lexical properties of the NSUs and their antecedents. The features can be categorized as: \textit{NSU features}, \textit{Antecedent features}, \textit{Similarity features}. Table \ref{table:baseline_feature_set} contains an overview of the feature set.

\subsubsection{NSU features}
Different NSU classes are often distinguished by their form. The following is a group of features exploiting their syntactic and lexical properties.
\begin{itemize}[label={}]
\item \texttt{nsu\_cont}\\ Denotes the ``content'' of the NSU i.e. whether it is a question or a proposition. This is useful to distinguish between question denoting classes, such as Clarification Ellipsis and Sluices, and the rest.
\item \texttt{wh\_nsu}\\ Denotes whether the NSU contains a wh-word, namely: \textit{what}, \textit{which}, \textit{who}, \textit{where}, \textit{when}, \textit{how}. This can help for instance to distinguish instances of Sluices and Clarification Ellipsis knowing that the former are \textit{wh}-questions while the latter are not.
\item \texttt{aff\_neg}\\ Denotes the presence of a yes-word, a no-word or an ack-word in the NSU. Yes-words are for instance: \textit{yes}, \textit{yep}, \textit{aye}; no-words are for instance: \textit{no}, \textit{not}, \textit{nay}; ack-words are: \textit{right}, \textit{aha}, \textit{mhm}. This is particularly needed to distinguish between Affirmative Answers, Rejections and Acknowledgments.
\item \texttt{lex}\\ Indicates the presence of lexical items at the beginning of the NSU. This feature is intended to indicate the presence of modifiers. A modal adverb (e.g. \textit{absolutely}, \textit{clearly}, \textit{probably}) at the beginning of the utterance usually denotes a Propositional Modifier. The same applies for Factual Modifiers, which are usually denoted by factual adjectives (e.g. \textit{good}, \textit{amazing}, \textit{terrible}, \textit{brilliant}) and Conjuncts which are usually denoted by conjunctions. Bare Modifier Phrases are a wider class of NSUs which do not have a precise lexical conformation but they are usually started by lexical patterns containing a Prepositional Phrase (PP) or an Adverbial Phrase (AdvP).
\end{itemize}

\begin{table}[t]
	\centering
	\def\arraystretch{1.8}
	\begin{tabular}{|l p{0.48\linewidth} l|}
		\hline
		\textbf{Feature}     &   \textbf{Description} & \textbf{Values} \\ \hline
		\texttt{nsu\_cont} & the content of the NSU (a question or a proposition) & \texttt{p,q} \\
		\texttt{wh\_nsu} & presence of a wh-word in the NSU & \texttt{yes,no} \\
		\texttt{aff\_neg} & presence of a yes/no-word in the NSU & \texttt{yes,no,e(mpty)} \\
		\texttt{lex} & presence of different lexical items at the beginning of the NSU & \texttt{p\_mod,f\_mod,mod,conj,e} \\ \hline
		\texttt{ant\_mood} & mood of the antecedent utterance & \texttt{decl,n\_decl} \\
		\texttt{wh\_ant} & presence of a wh-word in the antecedent & \texttt{yes,no} \\ 
		\texttt{finished} & whether the antecedent is (un)finished & \texttt{fin,unf} \\ \hline
		\texttt{repeat} & number of common words in the NSU and the antecedent & $0$-$3$ \\
		\texttt{parallel} & length of the common tag sequence in the NSU and the antecedent & $0$-$3$ \\
		\hline
	\end{tabular}
	\caption{An overview of the baseline feature set.}
	\label{table:baseline_feature_set}
\end{table}

\subsubsection{Antecedent features}
As for the NSUs, antecedents also show different syntactic and lexical properties that can be used as features for the classification task. This is a group of features exploiting those properties.
\begin{itemize}[label={}]
\item \texttt{ant\_mood} \\As defined by \citet{rodriguez2004form}, this feature was though to distinguish between declarative and non-declarative antecedent sentences. 
This feature is useful to indicate the presence of an answer NSU, if the antecedent is a question, or a modifier, if the antecedent is not a question.
\item \texttt{wh\_ant} \\As the corresponding NSU feature, this indicates the presence of a \textit{wh}-word in the antecedent. Usually Short Answers are answers to \textit{wh}-questions while Affirmative Answers and Rejections are are answers to polar questions i.e. \textit{yes}/\textit{no}-questions without a \textit{wh}-interrogative.
\item \texttt{finished} \\This feature encodes a truncated antecedent sentence as well as the presence of uncertainties at the end of it. Truncated sentences lack a closing full stop, question mark or exclamation mark. Uncertainties are given by the presence of pauses or unclear words or else a last word being ``non-closing'', e.g. conjunctions or articles.
\end{itemize}

\bgroup
\titlespacing\subsubsection{0pt}{6pt}{0pt}
\subsubsection{Similarity features}
\egroup
As discussed in Section \ref{sec:NSU}, some classes show some kind of parallelism between the NSU and its antecedent. The parallelism of certain classes can be partially captured by similarity measures. The following is a group of features encoding the similarity at the word and POS-level between the NSUs and their antecedents.
\begin{itemize}[label={}]
\item \texttt{repeat} \\This feature counts the content words that the NSU and the antecedent have in common (a maximum value of $3$ is taken as a simplification). A value greater than $0$ is usually a sign of Repeated Acknowledgment or Repeated Affirmative Answers.
\item \texttt{parallel} \\This feature encodes whether there is a common sequence of POS tags between the NSU and the antecedent and denotes its length.
This feature can help classify Repeated Acknowledgments, Repeated Affirmative Answers and Helpful Rejections.
\end{itemize}

\bgroup
\titlespacing\section{0pt}{12pt plus 4pt minus 2pt}{2pt plus 2pt minus 2pt}
\section{Feature engineering}
\label{sec:extended}
\egroup
The first and most straightforward method we use to address the classification problem is to find more features to describe the NSU instances. We present here the combination of features that we employ as our final approach. The extended feature set is composed of all the baseline features plus 23 new linguistic features, summing up to a total of 32 features.
Our features can be clustered into five groups: \textit{POS-level features}, \textit{Phrase-level features}, \textit{Dependency features}, \textit{Turn-taking features} and \textit{Similarity features}. Table \ref{table:final_feature_set} shows an overview of the additional features we use in the extended feature set.

\begin{table}[p]
	\centering
	\def\arraystretch{1.8}
	\begin{tabular}{|l p{0.48\linewidth} l|}
		\hline
		\textbf{Feature}     &   \textbf{Description} & \textbf{Values} \\ \hline
		\texttt{pos\_\{1,2,3,4\}} & POS tags of the first four words in the NSU & \texttt{C5 tag-set} \\
		\texttt{ending\_punct} & ending punctuation in the antecedent if any & \texttt{.,?,!,e} \\
		\texttt{has\_pause} & presence of a pause in the antecedent & \texttt{yes,no} \\
		\texttt{has\_unclear} & presence of an ``unclear'' marker in the antecedent & \texttt{yes,no} \\ \hline
		\texttt{ant\_sq} & presence of a SQ tag in the antecedent & \texttt{yes,no} \\
		\texttt{ant\_sbarq} & presence of a SBARQ tag in the antecedent & \texttt{yes,no} \\
		\texttt{ant\_sinv} & presence of a SINV tag in the antecedent & \texttt{yes,no} \\
		\texttt{nsu\_first\_clause} & first clause-level syntactic tag in the NSU & \texttt{S,SQ,\dots} \\
		\texttt{nsu\_first\_phrase} & first phrase-level syntactic tag in the NSU & \texttt{NP,ADVP,\dots} \\
		\texttt{nsu\_first\_word} & first word-level syntactic tag in the NSU & \texttt{NN,RB,\dots} \\ 
		\texttt{neg\_correct} & presence of a negation followed by a correction & {yes,no} \\
		\hline
		\texttt{ant\_neg} & presence of a \textit{neg} dependency in the antecedent & \texttt{yes,no} \\
		\texttt{wh\_inter} & presence of a \textit{wh}-interrogative fragment in the antecedent & \texttt{yes,no} \\ \hline
		\texttt{same\_who} & whether the NSU and its antecedent have been uttered by the same speaker & \texttt{same,diff,unk} \\ \hline
		\texttt{repeat\_last} & number of repeated words between the NSU and the last part of the antecedent & \texttt{numeric} \\
		\texttt{abs\_len} & number of words in the NSU & \texttt{numeric} \\
		\texttt{cont\_len} & number of content-words in the NSU & \texttt{numeric} \\
		\texttt{local\_all} & the local alignment (at character-level) of the NSU and the antecedent & \texttt{numeric} \\
		\texttt{lcs} & longest common subsequence (at word-level) between the NSU and the antecedent & \texttt{numeric} \\
		\texttt{lcs\_pos} & longest common subsequence (at pos-level) between the NSU and the antecedent & \texttt{numeric} \\
		\hline
	\end{tabular}
	\caption{An overview of the additional features comprised in the extended feature set.}
	\label{table:final_feature_set}
\end{table}
\bgroup
\titlespacing\subsubsection{0pt}{0pt}{0pt}
\subsubsection{POS-level features}
\egroup
Shallow syntactic properties of the NSUs that make use of the pieces of information already present in the BNC such as POS tags and other markers.
\begin{itemize}[label={}]
\item \texttt{pos\_\{1,2,3,4\}} \\A feature for each one of the first four POS-tags in the NSU. If an NSU is shorter than four words the value \texttt{None} is assigned to each missing POS tag. Many NSU classes share (shallow) syntactic patterns among their instances, especially at the beginning of the NSU phrase. Those features aim to capture those patterns in a shallow way through the POS tags.
\item \texttt{ending\_punct} \\A feature to encode the final punctuation mark of the antecedent if any.
\item \texttt{has\_pause} \\Marks the presence of a pause in the antecedent.
\item \texttt{has\_unclear} \\Marks the presence of an unclear passage in the antecedent.
\end{itemize}
\bgroup
\titlespacing\subsubsection{0pt}{0pt}{0pt}
\subsubsection{Phrase-level features}
\egroup
Occurrence of certain syntactic structures in the NSU and the antecedent. These features were extracted through the use of the Stanford PCFG parser \citep{klein2003accurate} on the utterances. Refer to \citet{Marcus:1993} for more information about the tag set used for the English grammar.
\begin{itemize}[label={}]
\item \texttt{ant\_\{sq,sbarq,sinv\}} \\Those features indicate the presence of the syntactic tags \texttt{SQ}, \texttt{SBARQ} and \texttt{SINV} in the antecedent. Those tags indicate a question formulated in various ways even when there is no explicit question mark at the end. Useful to recognize e.g. Short Answers.
\item \texttt{nsu\_first\_clause} \\Marks the first clause-level tag (\texttt{S}, \texttt{SQ}, \texttt{SBAR}, \dots) in the NSU. 
\item \texttt{nsu\_first\_phrase} \\Marks the first phrase-level tag (\texttt{NP}, \texttt{VP}, \texttt{ADJP}, \dots) in the NSU.
\item \texttt{nsu\_first\_word} \\Marks the first word-level tag (\texttt{NN}, \texttt{RB}, \texttt{UH}, \dots) in the NSU.
\item \texttt{neg\_correct} \\Presence of a negation word (\textit{no}, \textit{nope}, \dots), followed by a comma and a correction. 
		For instance:
		\begin{example} 
			\label{ex:neg_correct} \begin{dialogue} 
				\speak{A} Or, or were they different in your childhood? 
				\speak{B} No, always the same. \\ \bncref{HDH}{158--159}
			\end{dialogue}
		\end{example}
		This pattern is useful to describe some of the Helpful Rejections such as \ref{ex:neg_correct}.
\end{itemize}

\subsubsection{Dependency features}
Presence of certain dependency patterns in the antecedent. These features were extracted through the use of the Stanford Dependency Parser \citep{chen2014fast} on the utterances. For more details about the dependency relations tag set please refer to \citet{de2014universal}.
\begin{itemize}[label={}]
\item \texttt{ant\_neg} \\Signals the presence of a \textit{neg} dependency relation in the antecedent. The \textit{neg} dependency arises from an adverbial negation in the sentence (\textit{not}, \textit{don't}, \textit{never}, \dots). This feature helps to capture situations such as the following:
	\begin{example}
	\label{ex:ant_neg} \begin{dialogue}
			\speak{A} You're not getting any funny fits from that at all, June?
			\speak{B} Er no. \\ \bncref{H4P}{36--37}
		\end{dialogue}
	\end{example}
	Since the question in the antecedent is negative, the NSU in \ref{ex:ant_neg} is actually an Affirmative Answer, even though it contains a negative word.
This feature, in combination with the \texttt{aff\_neg} feature, addresses this situation.
\item \texttt{wh\_inter} \\Whether the antecedent contains a \textit{wh-interrogative} fragment such as the one in the following example:
	\begin{example}
		\label{ex:wh_inter} \begin{dialogue}
		\speak{A} And you know what the voltage is
		\speak{B} Yeah, two forty. \\ \bncref{GYR}{174--175}
		\end{dialogue}
	\end{example}
	The feature looks for a \textit{dobj} dependency with a wh-word then for an \textit{nsubj} dependency with the dependent element of the previous dependency, for instance in \ref{ex:wh_inter} we have \texttt{dobj(is-7, what-4)} and \texttt{nsubj(is-7, voltage-6)}.
This features tries to mitigate the absence of a question as antecedent for Short Answers such as \ref{ex:wh_inter}.
\end{itemize}

\subsubsection{Turn-taking features}
Features indicating certain patterns in the turn-taking of the dialogue.
\begin{itemize}[label={}]
\item \texttt{same\_who} \\Denotes whether the NSU and the antecedent were uttered by the same speaker. Sometimes dialogues do not provide the speaker information so an additional value \texttt{unk} is added for this cases.
This feature is particularly important to capture Check Questions which are almost always uttered by the same speaker.
\end{itemize}

\subsubsection{Similarity features}
Additional numeric features and similarity measures between the NSU and its antecedent.
\begin{itemize}[label={}]
\item \texttt{repeat\_last} \\This measures the number of words in common between the NSU and the last portion of the antecedent. Often happens that Repeated Affirmative Answers and Repeated Acknowledgments contain the last words in the antecedent.
\item \texttt{abs\_len} \\The total number of words in the NSU.
\item \texttt{cont\_len} \\The number of content-words in the NSU.
\item \texttt{local\_all} \\A feature that denotes the local alignment at the character-level between the NSU and the antecedent, computed using the Smith–Waterman algorithm \citep{smith1981identification}.
\item \texttt{lcs} \\A feature to express the longest common subsequence at the word-level between the NSU and its antecedent, computed using a modified version of the Needleman–Wunsch algorithm \citep{needleman1970general}, tailored to account for words instead of characters.
\item \texttt{lcs\_pos} \\The longest common subsequence at the POS-level between the NSU and its antecedent, computed with the same algorithm of above but using the list of POS tags instead of the list of words.
\end{itemize}



\section{Semi-Supervised Learning}
\label{sec:ssl}
The scarcity of labeled data is probably the major problem to face in this classification task. Even though the quality of the data is good enough, it is still difficult for a classifier to learn patterns out of $20$ instances or less for some classes (see Table \ref{table:nsus_rest_dist}).
However, a large amount of unlabeled data is available in the BNC.
There are many classification tasks, such as ours, in which it is hard or costly to label a large amount of instances while instead it is relatively cheap to extract unlabeled ones. The empirical question is whether the use of unlabeled data is useful to improve the classification performances. Semi-Supervised Learning techniques deal with this issue. They exploit the combination of a small amount of labeled data and a large amount of unlabeled data to try improve the classification accuracy. Even though it is still a young research field, semi-supervised learning has already found many fields of application \citep{liang2005semi,bergsma2010large}.

\subsection{Unlabeled data extraction}
\label{sec:unlabeled_data_extraction}
With the use of some heuristics it is possible to extract NSU instances of good quality from the BNC. We use a set of rules to determine whether an utterance in a dialogue transcript of the BNC is a probable NSU. The following is a list of such rules:
\begin{itemize}
\item The number of words in the NSU must be less than a given threshold;
\item The number of characters in the NSU must be higher than a given threshold;
\item The NSU must not contain only pauses, unclear passages and punctuation;
\item The NSU must not contain a greeting (e.g. \textit{hi}, \textit{hello}, \textit{good night});
\item The NSU must not contain a verb in any form.
\end{itemize}
An accuracy test was run over the corpus of NSUs: of the $1\,123$ NSUs examined, $1033$ where detected correctly by this set of rules, for an accuracy of $0.92$. The main flaws of the rules were mostly overlong NSUs, such as \ref{ex:detect_error1}, and presence of verbs, such as \ref{ex:detect_error2}.
\begin{example}
	\label{ex:detect_error1} \begin{dialogue}
		\speak{A} Was it a coal fire?
		\speak{B} Coal fire and er scrubbed the cabin out like that, soda water and soft soap.\footnote{A Repeated Affirmative Answer, but the additional content after the conjunction makes the NSU much longer. It is still a valid NSU since it does not have a full clausal structure.} \\ \bncref{H5G}{151--152}
	\end{dialogue}
\end{example}
\begin{example}
	\label{ex:detect_error2} \begin{dialogue}
		\speak{A} [\dots] the resistance the same the current goes up.
		\speak{B} Current goes up.\footnote{The NSU is a Repeated Acknowledgment. Repeating the words in the antecedent, it introduces a verb. It is still considered an NSU according to the definition of \citet{Fernandez:thesis}.} \\ \bncref{GYR}{112--113}
	\end{dialogue}
\end{example}

The detection of NSUs using the rules above is not the only problem to face. Perhaps more challenging is the selection of an antecedent for the NSU. As pointed out in Section \ref{sec:NSU}, the antecedent of an NSU is not always the preceding utterance. 
Nevertheless, as proved in the corpus study of \citet{Fernandez:thesis}, the percentage of the utterances whose antecedent is not the preceding utterance is rather low. Another result of the aforementioned work is that the case in which the antecedent is an utterance at distance greater than one is far more probable in a multi-party dialogue context.
In light of the above considerations, we restrict the instances we extract to only those from two-party dialogues and we always consider the preceding utterance as the antecedent of an NSU.
While there has been some previous work towards using machine learning techniques for the detection of the antecedent of NSUs in multi-party dialogue \citep{Schlangen:2005}, we consider sufficient the amount of unlabeled data we can extract following the previous rule.

In order to maximize the quality of the unlabeled data that we extract we also enforce some rules over the antecedent utterance:
\begin{itemize}
\item The number of words in the antecedent must be greater than the number of words in the NSU;
\item The antecedent must have a complete clausal form i.e. at least a verb phrase and a noun phrase.
\end{itemize}

Using the whole set of heuristics we extracted in total $3\,198$ new unlabeled NSU instances from the BNC (checked not to be already in the corpus).

\subsection{Semi-supervised learning techniques}
As previously mentioned, semi-supervised learning techniques are used when labeled data is scarce and unlabeled data is abundant. Every techniques tries to integrate the information yield by the unlabeled instances inside a learning model based on the available labeled data. In this section we give a brief and high-level description of the semi-supervised learning techniques that we have employed, namely: \textit{Self Training}, \textit{Transductive SVM} and \textit{Active Learning}.

\subsubsection{Self Training}
The simplest way to exploit unlabeled data is to automatically predict some unlabeled instances through a classifier built from the available labeled data then add them to the training data for the next step. This is an iterative process, at each step one or more newly labeled instances are added to the training set then the classifier is retrained and more unlabeled instances are predicted.

Various strategies can be used at each step:
\begin{itemize}
\item Add one or a few (random) instances at the time;
\item Add a few most confident instances;
\item Add all the first time, correct the wrong predictions the next times.
\end{itemize}
The last strategy as well as other variants can be cast as an Expectation-Maximization problem, especially when using a probabilistic learning model.

\subsubsection{Transductive SVM}
As already described in Section \ref{sec:SMO}, Support Vector Machines are one of the most studied and reliable family of classification algorithms. Transductive SVM (TSVM) is a variant of the standard SVM algorithm which exploits unlabeled data to help adjust the SVM model. The basic assumption under TSVM is that unlabeled instances from different classes are separated with large margin. Therefore, similarly to the standard SVM, TSVM tries to find the hyperplane that maximizes the unlabeled data margin i.e. considering unlabeled points as labeled ones. To decide whether an unlabeled point should be considered of one class or the other, clustering techniques are used e.g. $k$-nearest neighbors (the class of the majority of the neighbors or some other variant).

We will not go into mathematical details so we recommend the interested reader to \citet{vapnik1998statistical}, \citet{Collobert:2006}.

\subsubsection{Active Learning}
\label{sec:active_learning}
Annotating data is often a very expensive procedure, mostly because one needs to annotate a lot of instances in order to be able to reliably classify unseen ones. An idea to ease this problem is to let the learning algorithm choose which instance could be the most informative (i.e. the most difficult to predict) then annotate it manually. 
This technique has the advantage of reducing the cost of manual annotation of the instances by making informed guesses over the instances to label and discarding the redundant ones.

This kind of techniques is typically employed to cope with the scarcity of labeled data. In our case, the lack of sufficient training data is especially problematic due to the strong class imbalance between the NSU classes.

The Active Learning (AL) scheme, which is a special case of semi-supervised learning, trains the model over the available labeled data then queries the user for the label of one (or few more) instances then retrain the model and so on until convergence criteria are met, e.g. the wanted number of new instances is reached.

There can be different query strategies, some of them are:
\begin{itemize}
\item Uncertainty Sampling: queries the least confident instance (according to the probability of the prediction). A variant of that uses entropy to determine the most informative instance.
\item Query-by-committee: uses many different classifiers to predict unlabeled data then formulates the most informative query as the instance about which they most disagree.
\item Expected Model Change: selects the instance that would impart the	greatest change to the model, according to a decision-theoretic approach.
\item Expected Error Reduction: Another decision-theoretic approach that aims to minimize the \textit{risk}, that is the expected future error. The instances are selected on the basis of how much the model generalization error is likely to be reduced. A variant of this approach considers only the output variance of the model.
\end{itemize}

The particular active learning algorithm we employed in our experiments is a pool-based method\footnote{That involves drawing labeled instances from a ``pool'' that remains the same over the iterations, as opposed of stream-based ones in which sampling is done over a stream of data.} with uncertainty sampling \citep{lewis1994heterogeneous}. The sampling relies on \textit{entropy} as measure of uncertainty.
Given a particular (unlabeled) instance with a vector of feature values $\mathbf{f}$, we use the existing classifier to predict the class $C$ of the instance, and derive the probability distribution $P(C\!=\!c_i|\mathbf{f})$ for each possible output class $c_i$. We can then determine the corresponding entropy of the class $C$:
$$ H(C) = - \sum_i P(C\!=\!c_i|\mathbf{f})\log P(C\!=\!c_i|\mathbf{f}) $$
As seen in section \ref{sec:entropy}, entropy indicates the ``unpredictability'' of a random variable and also how much information it carries. The higher the entropy of the class of an instance the more information we gain by knowing it. The algorithm we employ (Algorithm \ref{alg:entropy_sampling}) selects the instances with highest entropy as the most informative ones. As argued in \citet{settles2010active}, entropy sampling is especially useful when there are more than two classes, as in our setting.
In practice, we applied the JCLAL active learning library\footnote{cf. \begin{scriptsize}\url{https://sourceforge.net/projects/jclal}\end{scriptsize}.} to extract and annotate 100 new instances of NSUs, which were subsequently added to the existing training data.

\IncMargin{2em}
\begin{algorithm}[t]
 \label{alg:entropy_sampling}
 \caption{$entropySampling(\Gamma, \mathbf{U}, k)$}

 \SetKwComment{Comment}{// }{}
 \Indm 
 \KwIn{The classifier $\Gamma$; the unlabeled data $\mathbf{U}$; the sample size $k$.}
 \KwOut{The $k$ instances with highest entropy.}
 \Indp

 $\mathbf{H} \leftarrow$ vector of the same size of $\mathbf{U}$\;
 
 \For{$i \in 1..|\mathbf{U}|$}{
 	$u \leftarrow \mathbf{U}[i]$\;
 	$\mathbf{P_{u}} \leftarrow classProbDist(\Gamma, u)$\;
 	$H_{u} \leftarrow - \sum_{p \in \mathbf{P_{u}}} p \, \log p$\;
 	$\mathbf{H}[i] \leftarrow H_{u}$\;
 }
 
 $\mathbf{U} \leftarrow sort(\mathbf{U}, \mathbf{H})$\Comment*{Sort $\mathbf{U}$ according to $\mathbf{H}$ (descending)}
 
 \Return $firstK(\mathbf{U}, k)$\;
\end{algorithm}
\DecMargin{2em}

\section{Evaluation}
\label{sec:eval}
In this section we discuss the evaluation of our experiments and their empirical results. We first discuss the evaluation metrics for the classification task we employed, then we present the evaluation results on each setting. 

\subsection{Metrics}

Given the dataset with a total of $N$ instances, the metrics are based on the amount of true positives ($TP$), true negatives ($TN$), false positive ($FP$) and false negatives ($FN$). 

\bgroup
\titlespacing\subsubsection{0pt}{0pt}{0pt}
\subsubsection{Accuracy}
\egroup
The ratio of the correctly classified instances over the total
$$ Acc = \frac{\sum_{c \in C} TP_c + TN_c}{N} $$
where $C$ is the set of the classes and $TP_c$ and $TN_c$ are respectively the true positives and the true negatives of the class $c \in C$.

\bgroup
\titlespacing\subsubsection{0pt}{0pt}{0pt}
\subsubsection{Precision}
\egroup
The ratio between the true positives and the total instances classified as positives. In a context with multiple classes (more than two) such as ours, the precision must be calculated per class, where the positive instances are the ones classified with the current class whereas the negative instances are the ones classified otherwise.
The per class precision is calculated as follows:
$$ Prec_c = \frac{TP_c}{TP_c + FP_c} $$

To have a summary value for all the classes we can compute the weighted average precision:
$$ Prec_{avg} = \frac{\sum_{c \in C} N_c \cdot Prec_c}{N} $$

\bgroup
\titlespacing\subsubsection{0pt}{0pt}{0pt}
\subsubsection{Recall}
\egroup
The recall is the ratio between the true positives and the total instances that are actually positives.
As for the precision, we can calculate the per class recall:
$$ Rec_c = \frac{TP_c}{TP_c + FN_c} $$

And the weighted average recall:
$$ Rec_{avg} = \frac{\sum_{c \in C} N_c \cdot Rec_c}{N} $$

\bgroup
\titlespacing\subsubsection{0pt}{0pt}{0pt}
\subsubsection{$F_1$-score}
\egroup
The $F_1$-score is the harmonic mean of precision and recall.
As for the other two measures, we compute the per class $F_1$-score:
\begin{align*}
F_{1,c} &= 2 \cdot \frac{Prec_c \cdot Rec_c}{Prec_c + Rec_c} \\
        &= \frac{2 \cdot TP_c}{ 2 \cdot TP_c + FP_c + FN_c}
\end{align*}

Then the weighted average $F_1$-score:
$$ F_{1,avg} = \frac{\sum_{c \in C} N_c \, F_{1,c}}{N} $$

\bgroup
\titlespacing\subsubsection{0pt}{0pt}{0pt}
\subsubsection{Student's $t$-test}
\label{sec:t-test}
\egroup
Empirical results alone can not assess whether a classifier performs \textit{better} than another. To assess that the performances of one classifier being higher than a second one is not due to the randomness associated with the data manipulation but to a \textit{statistically significant} difference between the classifiers one needs to prove with high confidence that the \textit{null hypothesis} is false. The \textit{null hypothesis} is a statement that is assumed to be true until evidence indicates otherwise. When comparing two learning systems, the \textit{null hypothesis} states that there is no difference between the performances of the two learning systems. To prove that a classifier performs better than another we need to disprove the \textit{null hypothesis} with a high degree of confidence. For this purpose we employ a Student's $t$-test, a widespread method to compare two sets of data. The $t$-test can be used to find the probability $p$ of the performance values of the two classifiers being drawn from the same mean.

To run the $t$-test, we compare the differences $\delta_i$ among the performance values of the two classifiers over the $n$ independent samples.
We first compute the mean of the differences:
$$ \bar{\delta} = \frac{1}{n} \sum_{i=1}^{n} \delta_i $$
Then we compute the $t$-statistic:
$$ t = \frac{\bar{\delta}}{\sqrt{\frac{1}{n(n-1)} \sum_{i=1}^{n} (\delta_i - \bar{\delta})^2}} $$

From the $t$-statistic we can derive the $p$-value from a \textit{Student's $t$-distribution} with $n-1$ degree of freedom. A small $p$-value means that it is unlikely that the samples show such a $t$-statistic by chance therefore we can assess that the difference in performance between the two classifiers is statistically significant.

In our case we use a paired $t$-test on the accuracy values of the 10-fold cross-validation over the dataset (thus $n=10$).
By convention, an acceptable $p$-value is $p \leq 0.05$. For our experiments we rely on the \texttt{t.test} function from the \textit{R} project, a framework for statistical computing \citep{R:manual}. 

\subsection{Empirical results}

\subsubsection{Baseline}
As in \citet{Fernandez:2007}, we evaluate our system in a 10-fold cross-validation fashion.
Weka's J48 algorithm was used as a comparing classifier. Thanks to the analysis of the resulting trees, we managed to imitate quite closely the behavior of their system as well as reaching a very close performance overall.
Even though we use the same feature set and the same algorithm the performance parameters turn out to be slightly lower than the ones claimed in \citet{Fernandez:2007}. That might be for a variety of reasons, for instance the way feature were extracted or how the parameters were tuned.
Nevertheless the overall performance is matched as well as many of the patterns in the scores.
Table \ref{table:baseline_comparison} shows the comparison between the performance parameters of the reference classification \citep{Fernandez:2007} and the values of the same parameters achieved by our implementation.

\begin{table}[t]
	\centering
	\begin{tabular}{|l | c c : c|c c : c|}
		\hline
		 & \multicolumn{3}{c|}{Our replica} & \multicolumn{3}{c|}{Reference classification} \\
		\textbf{NSU Class} & \begin{small}\textbf{Precision}\end{small} & \multicolumn{1}{c}{\begin{small}\textbf{Recall}\end{small}} & \begin{small}\textbf{$F_1$-score}\end{small} & \begin{small}\textbf{Precision}\end{small} & \multicolumn{1}{c}{\begin{small}\textbf{Recall}\end{small}} & \begin{small}\textbf{$F_1$-score}\end{small} \\ \hline
		Ack       & $0.97$ & $0.97$ & $0.97$ & $0.97$ & $0.95$ & $0.96$ \\
		AffAns    & $0.89$ & $0.84$ & $0.86$ & $0.83$ & $0.86$ & $0.84$ \\
		BareModPh & $0.63$ & $0.65$ & $0.62$ & $1.00$ & $0.70$ & $0.82$ \\
		CE        & $0.87$ & $0.89$ & $0.87$ & $0.92$ & $0.92$ & $0.94$ \\
		CheckQu   & $0.85$ & $0.90$ & $0.87$ & $0.83$ & $1.00$ & $0.91$ \\
		ConjFrag  & $0.80$ & $0.80$ & $0.80$ & $0.71$ & $1.00$ & $0.83$ \\
		FactMod   & $1.00$ & $1.00$ & $1.00$ & $1.00$ & $0.91$ & $0.95$ \\
		Filler    & $0.77$ & $0.70$ & $0.71$ & $0.50$ & $0.37$ & $0.43$ \\
		HelpReject & $0.13$ & $0.14$ & $0.14$ & $0.46$ & $0.33$ & $0.39$ \\
		PropMod   & $0.92$ & $0.97$ & $0.93$ & $1.00$ & $0.60$ & $0.75$ \\
		Reject    & $0.76$ & $0.95$ & $0.83$ & $0.76$ & $1.00$ & $0.86$ \\
		RepAck    & $0.74$ & $0.75$ & $0.70$ & $0.84$ & $0.86$ & $0.85$ \\
		RepAffAns & $0.67$ & $0.71$ & $0.68$ & $0.65$ & $0.68$ & $0.67$ \\
		ShortAns  & $0.86$ & $0.80$ & $0.81$ & $0.81$ & $0.83$ & $0.82$ \\
		Sluice    & $0.67$ & $0.77$ & $0.71$ & $0.95$ & $1.00$ & $0.98$ \\
		\hline
		\textbf{weighted avg.}	& $\textbf{0.89}$ & $\textbf{0.89}$ & $\textbf{0.88}$ & $\textbf{0.90}$ & $\textbf{0.90}$ & $\textbf{0.89}$ \\
		\hline
	\end{tabular}
	\caption{Performances comparison between \citet{Fernandez:2007} and our replica.}
	\label{table:baseline_comparison}
\end{table}

\subsubsection{Self-training and TSVM}
Both those two techniques did not perform particularly well, sometimes even worsening the classification accuracy. Self-training was implemented and tested in many variants but none were successful. One possible explanation is that the labeled data added at each step to the training data is always biased by the labeled data available in the initial training set. This may lead to adding redundant data that is not actually useful to improve the classification performances.
On the other hand, TSVM has been unsuccessful mostly due to computational performances of the implementation and other technical difficulties. It was impracticable to run it on a large amount of unlabeled data so we managed to test it only on few unlabeled instances and therefore no improvement was shown.

\begin{figure}[p]
	\centering
	\includegraphics[width=0.98\textwidth, height=0.45\textheight]{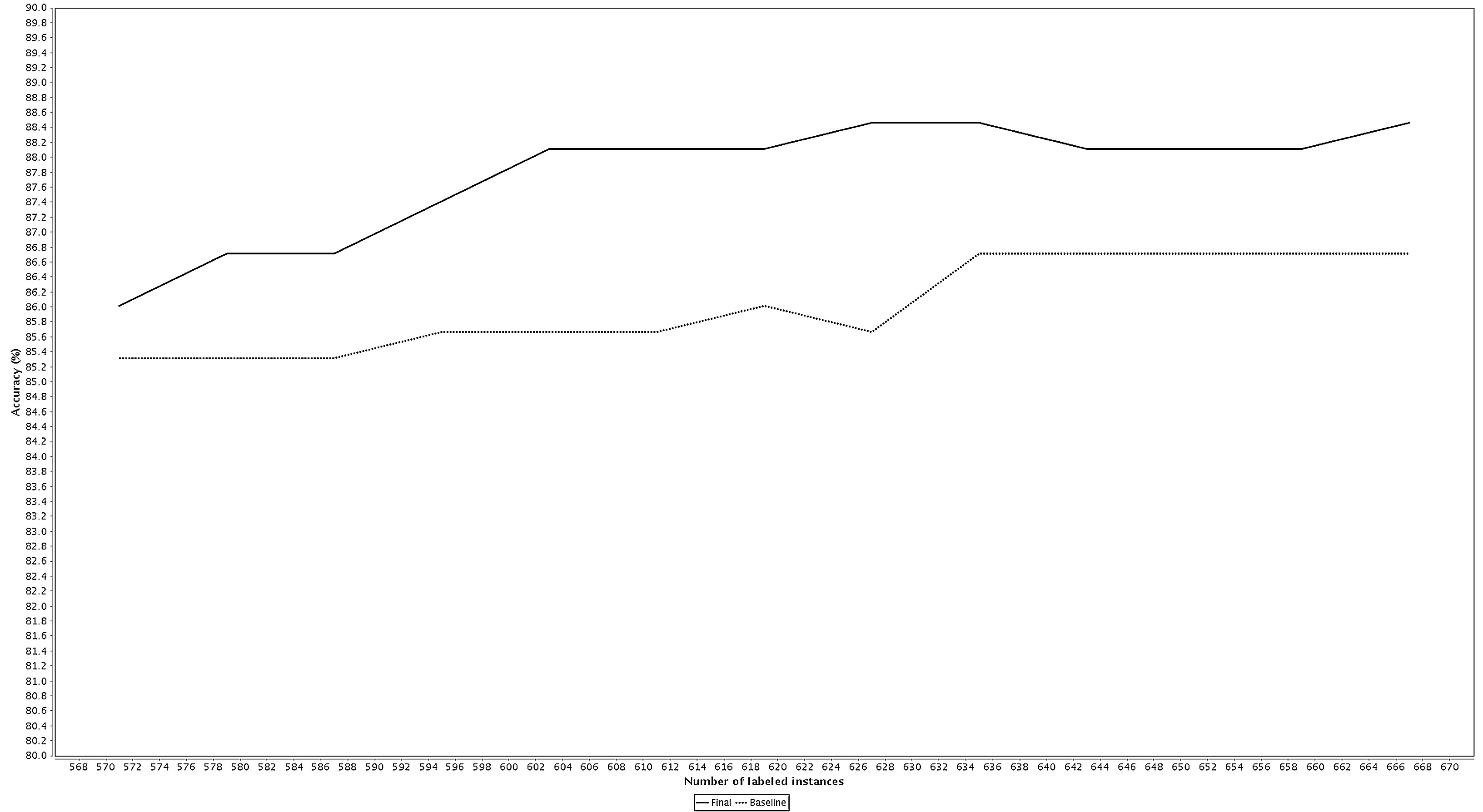}
	\caption{Learning curve for the accuracy (output of the JCLAL library).}
	\label{fig:al_accuracy}
\end{figure}

\begin{figure}[p]
	\centering
	\includegraphics[width=0.98\textwidth, height=0.45\textheight]{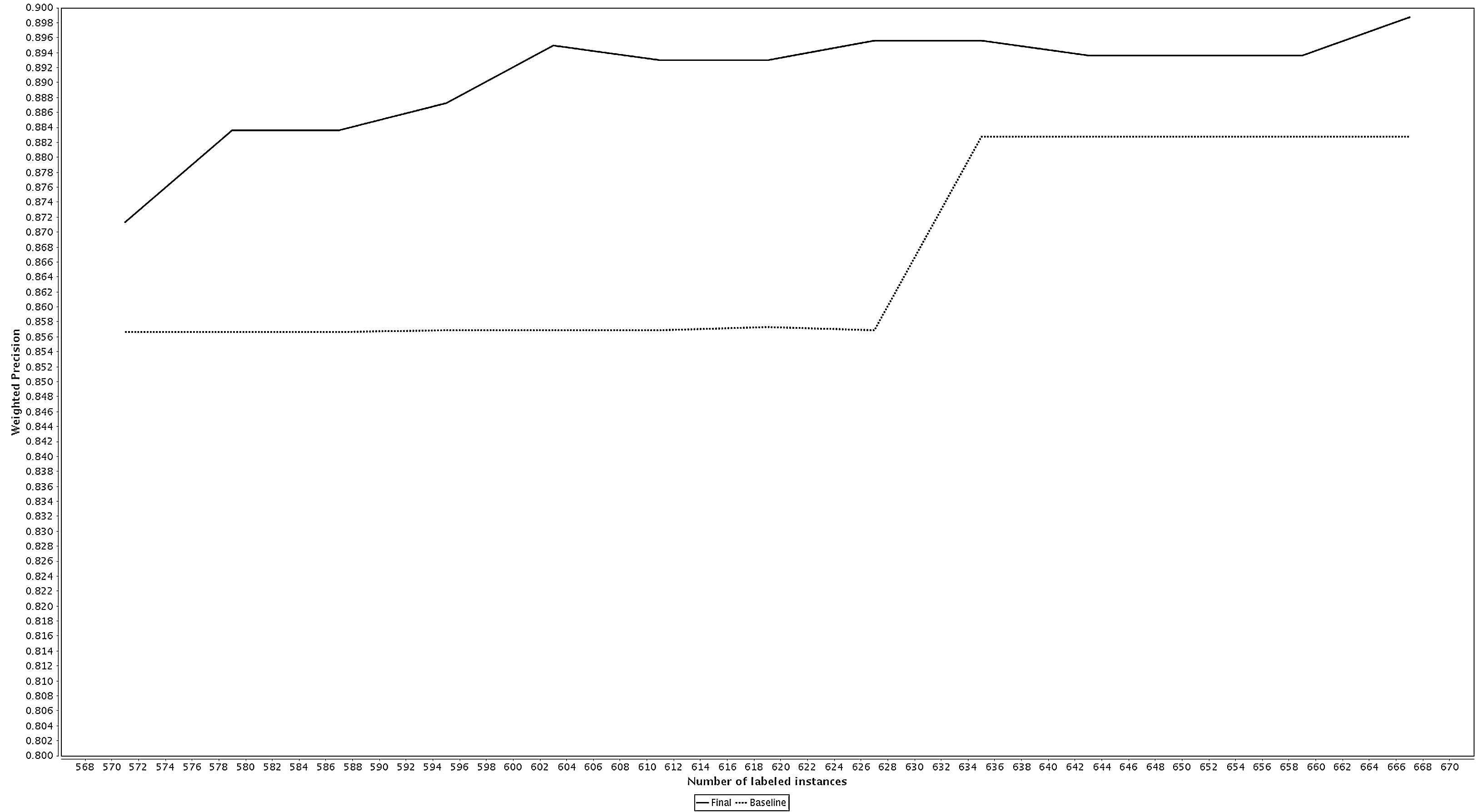}
	\caption{Learning curve for the precision (output of the JCLAL library).}
	\label{fig:al_precision}
\end{figure}

\begin{figure}[p]
	\centering
	\includegraphics[width=0.98\textwidth, height=0.45\textheight]{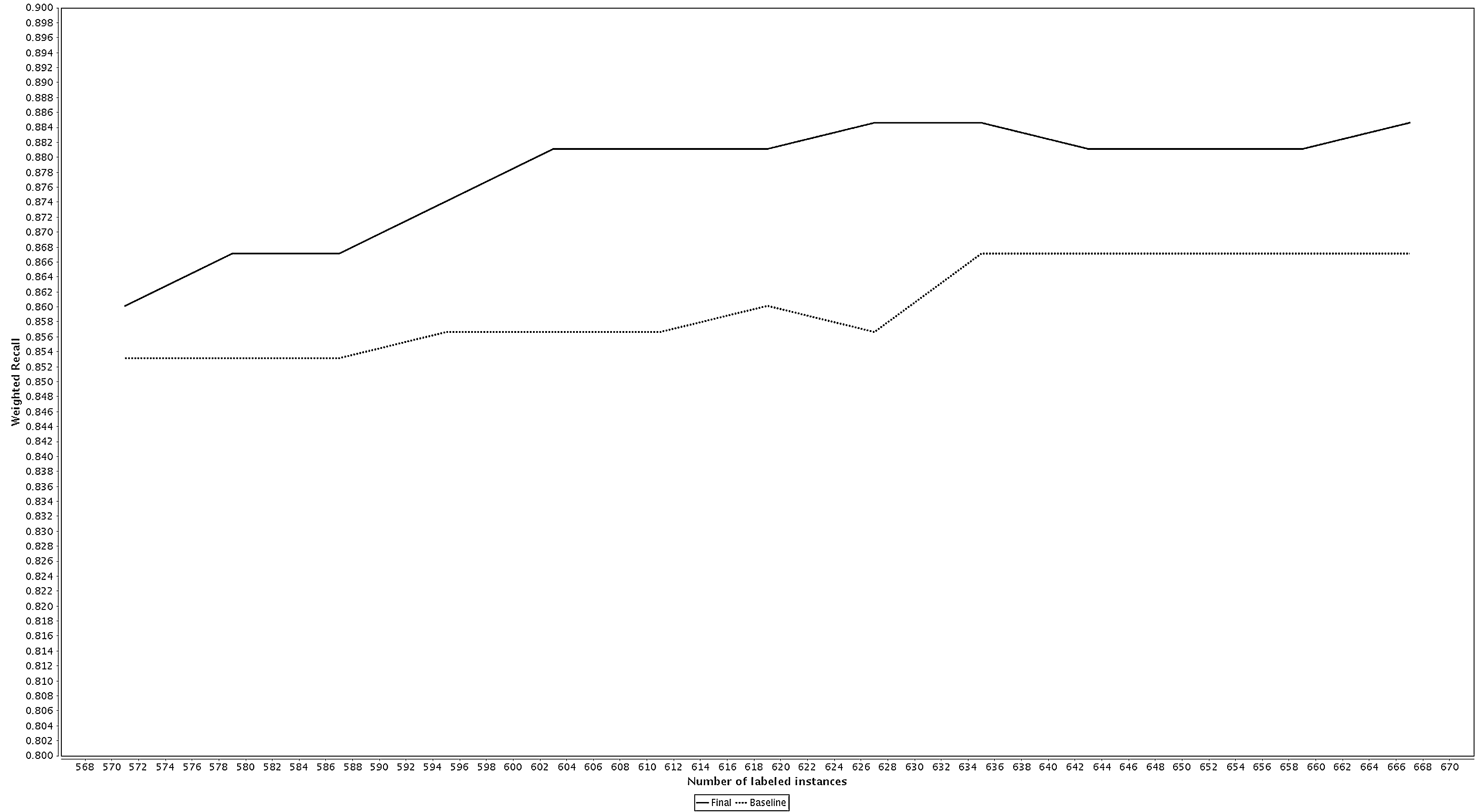}
	\caption{Learning curve for the recall (output of the JCLAL library).}
	\label{fig:al_recall}
\end{figure}

\begin{figure}[p]
	\centering
	\includegraphics[width=0.98\textwidth, height=0.45\textheight]{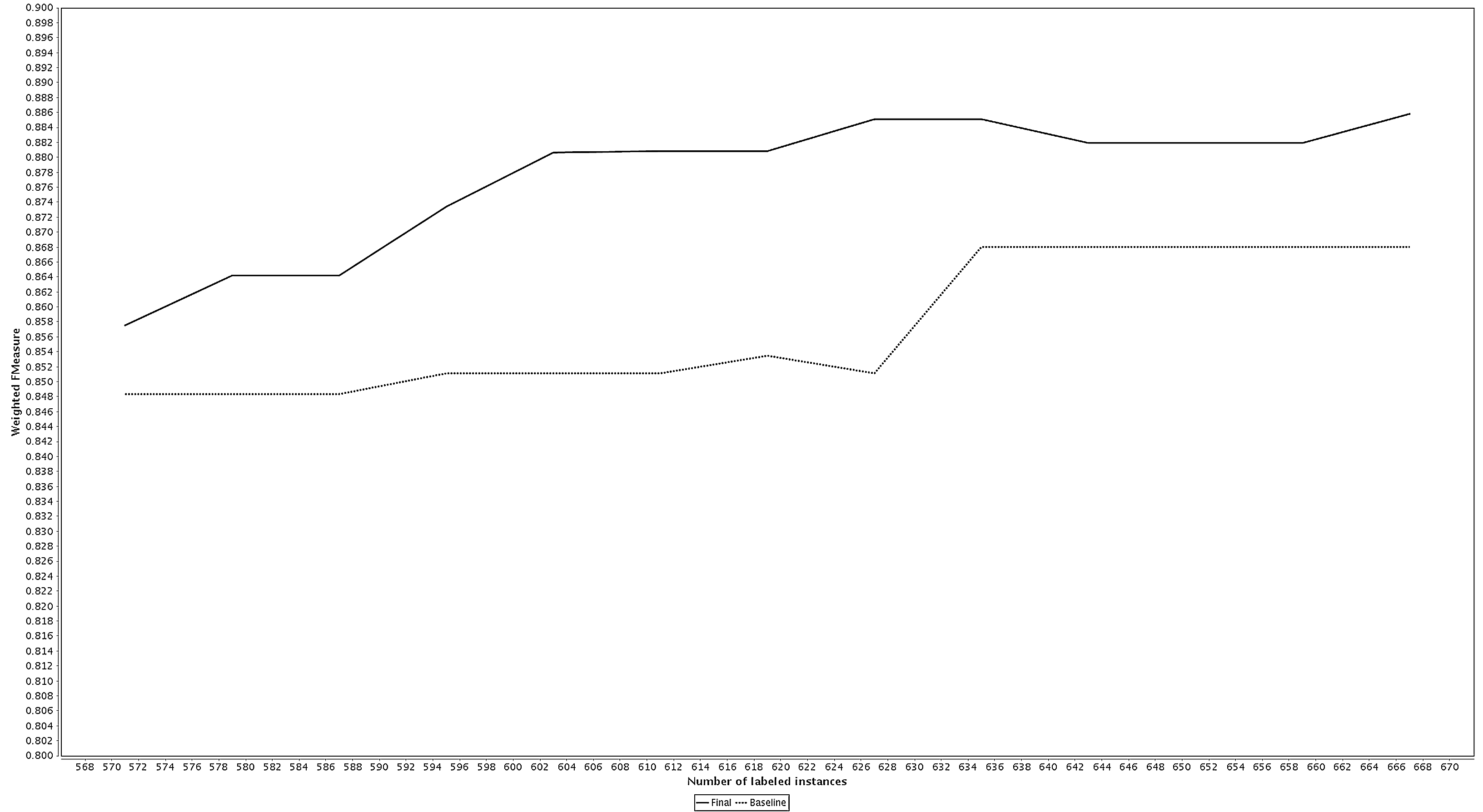}
	\caption{Learning curve for the $F_1$-score (output of the JCLAL library).}
	\label{fig:al_f-measure}
\end{figure}

\bgroup
\titlespacing\subsubsection{0pt}{0pt}{0pt}
\subsubsection{Active Learning}
\egroup
Our active learning experiment was carried out using the JCLAL library. For the active learning process we divided the dataset into three parts: training set (50\%), development set (25\%) and test set (25\%).
At each iteration, the JCLAL library builds a classifier on the training set and evaluates it over the development set. The same classifier is then used to select an instance from the unlabeled data, as described in Section \ref{sec:active_learning}. The user is then asked to annotate the selected instance.
The process iterates in this manner until the stopping criteria is met, that is when the goal of 100 newly annotated instances is reached. Table \ref{table:al_dist} shows the distribution of the instances annotated with Active Learning. From Table \ref{table:al_dist} we can see that the AL algorithm using the entropy measure prefers the instances that belongs to the classes that are most difficult to classify and, in particular, the ones that are ambiguous, such as Clarification Ellipsis and Sluices.
This process has been performed once with the extended feature set and the SMO classifier. Secondly, it has been simulated (i.e. using the data obtained in the previous run) using the baseline feature set instead. The Figures \ref{fig:al_accuracy}, \ref{fig:al_precision}, \ref{fig:al_recall}, \ref{fig:al_f-measure} show the learning curves\footnote{The graph showing how the performances change as the new labeled data extracted with Active Learning are inserted in the training set.}, respectively for the accuracy, precision, recall and $F_1$-score, of both the extended feature set and the baseline feature set.\footnote{Notice that the images are scaled on the $y$-axis to make the change visible.} All the performance measures are clearly improving as new instances become available, for both the extended feature set and the baseline one.


\begin{table}[t]
	\centering
	\begin{tabular}{|l r|}
		\hline
		\textbf{NSU Class} & \textbf{Instances} \\ \hline
		Helpful Rejection & $21$ \\
		Repeated Acknowledgment    & $17$ \\
		Clarification Ellipsis        & $17$ \\		
		Acknowledgment       & $11$ \\
		Propositional Modifier   & $9$ \\
		Filler    & $9$ \\
		Sluice    & $3$ \\
		Repeated Affirmative Answer & $3$ \\
		Factual Modifier   & $3$ \\
		Conjunct Fragment  & $3$ \\
		Short Answer  & $2$ \\
		Check Question   & $2$ \\		
		\hline
		 \textbf{tot.} & $\textbf{100}$ \\
		\hline
	\end{tabular}
	\caption{Distribution of the classes of the instances annotated with Active Learning.}
	\label{table:al_dist}
\end{table}

In the end, the test set has been used to evaluate the overall performances of the various settings. Table \ref{table:al_dev} and Table \ref{table:al_test} show the results of the experiments respectively over the development set and the test set. The results on the test set show that the inclusion of the active learning data is only beneficial when combined with the extended feature set.

We also performed an evaluation of the various settings using 10-fold cross-validation over the full dataset. The evaluation results based on the active learning procedure (AL) refer to the performance of the system after the inclusion of all newly annotated instances. The novel data was added to the training set of each fold. 

We compare the results of the various settings using the J48 algorithm (Table \ref{table:al_j48}) and SMO algorithm (Table \ref{table:al_svm}). The use of active learning was successful and, in the end, the use of the SMO classifier with the extended feature set and the inclusion of the AL instances constitutes our final approach.

The results show a significant improvement of the classification performance between the baseline and the final approach. Using a paired $t$-test with a 95\% confidence interval between the baseline and the final results (as detailed in Section \ref{sec:t-test}), the improvement in classification accuracy is statistically significant with a $p$-value of $6.9\times 10^{-3}$.

The SVM algorithm does not perform particularly well with the baseline feature set but scales better than the J48 classifier after the inclusion of the additional features. Overall, the results demonstrate that the classification can be improved using a modest amount of additional training data combined with an extended feature set. However, we can observe from Table \ref{table:al_details} that some NSU classes remain difficult to classify even with the insertion of additional training data. For instance, Helpful Rejections are still the most difficult classes to classify, even with the addition of 21 new instances. One of the problems with Helpful Rejections is that they are connected to their antecedents mainly at the semantic level. Consider the following example of Helpful Rejection that is hard to classify:
\begin{example}
\begin{dialogue}
\label{ex:wrong_help_rej}
\speak{A} There was one which you said Ernest Morris was born in 1950.
\speak{B} Fifteen. \bncref{J9A}{372--373}
\end{dialogue}
\end{example}

\begin{table*}[t]
	\centering
	\begin{tabular}{|l @{\hskip 54pt}| c | c | c | c |}
		\hline
		\textbf{Training set (feature set)}                        & \begin{small}\textbf{Accuracy}\end{small} & \begin{small}\textbf{Precision}\end{small} & \begin{small}\textbf{Recall}\end{small} & \begin{small}\textbf{$F_1$-score}\end{small} \\ \hline
		Train-set (baseline)      &           0.853 &       	 0.857 &	     0.853 &	        0.848 \\ \hline
		Train-set (extended)      &           0.860 &	         0.871 &	     0.860 &	        0.858 \\ \hline
		Train-set + AL (baseline) &           0.867	&            0.883 &	     0.867 &	        0.868 \\ \hline
		Train-set + AL (extended) &           0.884 &	         0.899 &	     0.885 &	        0.886 \\ \hline
	\end{tabular}
	\caption{Performances of the SMO classifier in the various settings on the development set.}
	\label{table:al_dev}
\end{table*}

\begin{table*}[t]
	\centering
	\begin{tabular}{|l | c | c | c | c |}
		\hline
		\textbf{Training set (feature set)}                        & \begin{small}\textbf{Accuracy}\end{small} & \begin{small}\textbf{Precision}\end{small} &  \begin{small}\textbf{Recall}\end{small} & \begin{small}\textbf{$F_1$-score}\end{small} \\ \hline
		Train-set + Dev-set (baseline)      &             0.906	&              0.911 &	          0.906 &	           0.903 \\ \hline
		Train-set + Dev-set (extended)      &             0.928 &	           0.937 &	          0.929 &	           0.930 \\ \hline
		Train-set + Dev-set + AL (baseline) &             0.898 &	           0.911 &         	  0.898 &	           0.898 \\ \hline
		Train-set + Dev-set + AL (extended) &    \textbf{0.932} &     \textbf{0.945} &   \textbf{0.932} &     \textbf{0.935} \\ \hline
	\end{tabular}
	\caption{Performances of the SMO classifier in the various settings on the test set.}
	\label{table:al_test}
\end{table*}

It is clear that, for the Helpful Rejection in \ref{ex:wrong_help_rej}, morpho-syntactic and lexical features, such as the ones we employ, are of little use in classifying this utterance. Most of the connection is at the semantic level therefore we would need to use features that exploit semantic patterns. At the same time, the use of this type of features would add several layers of complexity at the feature extraction process.
Other examples of difficult classes are the Repeated Affirmative Answers and Repeated Acknowledgments. They are highly ambiguous because they can be misclassified between each other, with their respective non-repeated classes and sometimes with other NSU classes. An example of ambiguous Repeated Acknowledgment can be the following:
\begin{example}
\label{ex:wrong_rep_ack}
\begin{dialogue}
\speak{A} Selected period. 
\speak{B} Selected period, right, Andrew?\footnote{In the dialogue, the speaker B is asking the same question to many people in turns.} \\ \bncref{JK8}{114--115}
\end{dialogue}
\end{example}
The instance in \ref{ex:wrong_rep_ack} contains also a question therefore it is often misclassified with other question denoting NSU classes.
It is clear that handling these type of NSU requires to perform a deeper semantic analysis of the connection with their antecedents then design appropriate semantic features. The extraction of additional labeled data is also especially important for both the feature engineering and the learning process of the classifiers. This two approaches may be the starting points of any future work on this task.

\begin{table*}[p]
	\centering
	\begin{tabular}{|l @{\hskip 54pt}| c | c | c | c |}
		\hline
		\textbf{Training set (feature set)}                        & \begin{small}\textbf{Accuracy}\end{small} & \begin{small}\textbf{Precision}\end{small} & \begin{small}\textbf{Recall}\end{small} & \begin{small}\textbf{$F_1$-score}\end{small} \\ \hline
		Train-set (baseline)      &           0.885 &            0.888 &         0.885 &            0.879 \\ \hline
		Train-set (extended)      &           0.889 &            0.904 &         0.889 &            0.889 \\ \hline
		Train-set + AL (baseline) &           0.890 &            0.896 &         0.890 &            0.885 \\ \hline
		Train-set + AL (extended) &           0.896 &            0.914 &         0.896 &            0.897 \\ \hline
	\end{tabular}
	\caption{Performances of the J48 classifier in the various settings using 10-fold cross-validation.}
	\label{table:al_j48}
\end{table*}

\begin{table*}[p]
	\centering
	\begin{tabular}{|l | c | c | c | c |}
		\hline
		\textbf{Training set (feature set)}                        & \begin{small}\textbf{Accuracy}\end{small} & \begin{small}\textbf{Precision}\end{small} &  \begin{small}\textbf{Recall}\end{small} & \begin{small}\textbf{$F_1$-score}\end{small} \\ \hline
		Train-set (baseline feature set)      &             0.881 &              0.884 &            0.881 &              0.875 \\ \hline
		Train-set (extended feature set)      &             0.899 &          	 0.904 &            0.899 &	             0.896 \\ \hline
		Train-set + AL (baseline feature set) &             0.883 &              0.893 &            0.883 &              0.880 \\ \hline
		Train-set + AL (extended feature set) &    \textbf{0.907} &     \textbf{0.913} &   \textbf{0.907} &     \textbf{0.905} \\ \hline
	\end{tabular}
	\caption{Performances of the SMO classifier in the various settings using 10-fold cross-validation.}
	\label{table:al_svm}
\end{table*}

\begin{table*}[p]
	\centering
	\begin{tabular}{|l | c  c : c | c  c : c |}
		\hline
		 & \multicolumn{3}{c|}{\textbf{Baseline}}     & \multicolumn{3}{c|}{\textbf{Final approach}} \\
		\textbf{NSU Class}                 & \begin{small}\textbf{Precision}\end{small} & \multicolumn{1}{c}{\begin{small}\textbf{Recall}\end{small}} & \begin{small}\textbf{$F_1$-score}\end{small} & \begin{small}\textbf{Precision}\end{small} & \multicolumn{1}{c}{\begin{small}\textbf{Recall}\end{small}} & \begin{small}\textbf{$F_1$-score}\end{small}   \\ \hline
		Ack   & 0.97 & 0.97 & 0.97 &     0.97 & 0.98 & 0.97 \\ \hline
		AffAns & 0.89 & 0.84 & 0.86 &     0.81 & 0.90 & 0.85 \\ \hline
		BareModPh & 0.63 & 0.65 & 0.62 &     0.77 & 0.75 & 0.75 \\ \hline
		CE & 0.87 & 0.89 & 0.87 &     0.88 & 0.92 & 0.89 \\ \hline
		CheckQu & 0.85 & 0.90 & 0.87 &     1.00 & 1.00 & 1.00 \\ \hline
		ConjFrag & 0.80 & 0.80 & 0.80 &     1.00 & 1.00 & 1.00 \\ \hline
		FactMod & 1.00 & 1.00 & 1.00 &     1.00 & 1.00 & 1.00 \\ \hline
		Filler                 & 0.77 & 0.70 & 0.71 &     0.82 & 0.83 & 0.78 \\ \hline
		HelpReject & 0.13 & 0.14 & 0.14 &     0.31 & 0.43 & 0.33 \\ \hline
		PropMod & 0.92 & 0.97 & 0.93 &     0.92 & 1.00 & 0.95 \\ \hline
		Reject & 0.76 & 0.95 & 0.83 &     0.90 & 0.90 & 0.89 \\ \hline
	 	RepAck          & 0.74 & 0.75 & 0.70 &     0.77 & 0.77 & 0.77 \\ \hline
	 	RepAffAns     & 0.67 & 0.71 & 0.68 &     0.72 & 0.55 & 0.58 \\ \hline
	 	ShortAns & 0.86 & 0.80 & 0.81 &     0.92 & 0.86 & 0.89 \\ \hline
	 	Sluice                 & 0.67 & 0.77 & 0.71 &     0.80 & 0.84 & 0.81 \\ \hline
	\end{tabular}
	\caption{Per class performances comparison between the baseline (J48, baseline feature set) and the final approach (SMO, extended feature set, AL instances).}
	\label{table:al_details}
\end{table*}


\section{Summary}

This chapter presented the task of classifying non-sentential utterances and our approach to address this problem. This task is formulated as a machine learning problem and we follow and extend the work of \citet{Fernandez:2007}. We use their corpus as a gold-standard and a replica of their approach as a baseline. The data, the machine learning algorithm used and the feature set of the baseline were discussed respectively in Section \ref{sec:data}, \ref{sec:ml_alg}, \ref{sec:baseline}. The two main problems we faced in our work have been the scarcity of the labeled data and the imbalance in the distribution of the classes. To address these problems we extended the baseline approach in two ways: using a larger feature set (detailed in Section \ref{sec:extended}) and employing semi-supervised learning techniques to exploit the abundance of unlabeled data. We described in Section \ref{sec:ssl} the semi-supervised learning techniques that we employed, namely: Self Training, Transductive SVM and Active Learning. Section \ref{sec:eval} shows the empirical results we got from our experiments. While the extended feature set alone did not make an improvement on the performances of the classifiers, its use in combination with Active Learning made a modest but significant difference.

\chapter{Resolution of Non-Sentential Utterances}
\label{ch:resolution}
As introduced in Chapter \ref{ch:background}, the resolution of an NSU is the task of reconstructing its meaning from the dialogue context. \citet{Fernandez:thesis} proposes a set of rules to resolve NSUs based on TTR, the logical framework from \citet{cooper2004type} further developed then in \citet{Ginzburg:interactivestance}. One limitation of logical frameworks such as TTR is their inability to directly represent (and reason over) uncertain knowledge. Moreover, many dialogue domains contain variables that are only partially observable. We have to take into account a certain degree of stochastic behavior when modeling dialogue since we still have an imperfect understanding of its dynamics. The stochastic component is especially important in dealing with NSUs since they do not have a precise meaning by themselves and, as argued in \citet{Ginzburg:interactivestance}, they are in principle highly ambiguous.

For this reason we propose a new approach to the resolution of NSUs that takes probabilistic account of the variables involved and the procedures used. We employ the probabilistic rules formalism of \citet{Lison2015} (detailed in Section \ref{sec:prob_model}) to encode the NSU resolution procedures as probabilistic rules. Probabilistic rules are similar, to a certain extent, to the update rules developed by \citet{Ginzburg:interactivestance} (described in Section \ref{sec:formal}). For this reason probabilistic rules are particularly suited for our purpose since, in this way, we could reuse many theoretical aspects from \citet{Ginzburg:interactivestance} and \citet{Fernandez:thesis}. We reinterpreted the variables in the dialogue state as random variables and straightforwardly ``converted'' the resolution rules into probabilistic rules.  

In the next sections we explain how we represented the variables in the dialogue state and how we translated the rules from \citet{Fernandez:thesis} into probabilistic rules. First we describe the theoretical aspects from \citet{Ginzburg:interactivestance} we employ in our approach. We present then the design of our dialogue context and the rules to resolve the NSUs.

We surely take a much simpler approach than \citet{Ginzburg:interactivestance} in the modeling of the dialogue state, abstracting intentionally from many details that would add complexity to the modeling. Indeed there are a number of issues that arise in the resolution of NSUs that need to be treated with proper lexical and semantic resources that we did not include. However, in the end our goal for this work is not to formulate a complete theory of NSU resolution but rather to provide a \textit{proof-of-concept} implementation for the resolution of the NSUs in the dialogue context with the probabilistic rules formalism.

This framework has also been implemented and tested with the OpenDial toolkit. We developed a dialogue system able to update the dialogue state probabilistically with update rules similar to the ones from \citet{Ginzburg:interactivestance}. The system is also able to resolve toy examples in an interactive way. A detailed example of the behavior of the system is given in Section \ref{sec:testing}. More high-level details about the rules for the state update (complementary to the rules for the NSU resolution) can be found in the Appendix \ref{app:1}\footnote{For more technical details about the implementation and examples of interaction visit: \\\begin{scriptsize}\url{https://github.com/paolodragone}\end{scriptsize}}.

\section{The resolution task}
\label{sec:res_task}
\begin{figure}[h]
	\centering
	\includegraphics[height=0.19\textheight]{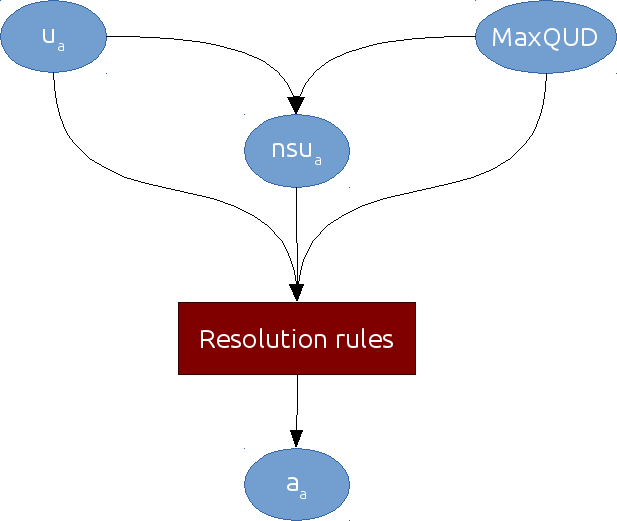}
	\caption{The basic schema for the NSU resolution task.}
	\label{fig:resolution_task}
\end{figure}

The resolution of an NSU is the task of extracting its meaning from the dialogue context. More precisely, let $u_a$ and $\textit{nsu}_a$ represent respectively the word word sequence making up the NSU and its type according to the taxonomy presented in Section \ref{sec:taxonomy}. We also assume MaxQUD to be a high-level semantic representation of the antecedent, as mentioned in Section \ref{sec:dialogue_context}. Through a resolution procedure, we want to extract $a_a$ i.e. the high-level semantic representation of the NSU. The right resolution procedure is selected on the basis of the type of the NSU. In our case the value of $\textit{nsu}_a$ is retrieved using the classifier developed in Chapter \ref{ch:classification} which takes as input the raw NSU and the antecedent. Figure \ref{fig:resolution_task} shows a schema of the task just defined.
Indeed this is the simplest way to define the task. The resolution procedure may also be dependent of other variables in the dialogue state such as the Facts. In principle, the resolution task is defined independently from the actual semantic representation of the utterances. It is also defined independently from the rules used to update the variables in the dialogue state such as QUD and Facts. In practice, define a set of rules that are generic enough to handle every possible case and behave independently from the state update rules is a difficult task and still an open research problem.
\section{Theoretical foundation}
\label{sec:theoretical_foundation}
As previously stated, we rely on \citet{Fernandez:thesis} and \citet{Ginzburg:interactivestance} for the theoretical notions needed to represent the dialogue state and to develop the NSU resolution rules. In Section \ref{sec:formal} we detailed the basic concepts of TTR, the utterance representation and the update rules for the dialogue state. In this section we describe the notions needed for the resolution of the NSUs. In particular we describe how we can exploit the \textit{parallelism} between the NSU and its antecedent that we mentioned in Section \ref{sec:NSU}. We discuss here the concepts that \citet{Ginzburg:interactivestance} defines to address the resolution of NSUs then we will describe how we adapt those concepts to our needs in the next section.

\subsection{Partial Parallelism}
Instances of NSU classes such as Acknowledgment and Affirmative Answers are related to their antecedent as a whole, that is to understand their meaning one as to consider not a specific aspect of the antecedent but the entire sentence. On the other hand, there are NSU classes, such as Short Answers and Sluices, that show a more fine grained \textit{parallelism} between their instances and their antecedents i.e. they may refer in particular to certain aspects of the antecedent. In the theory of \citet{Ginzburg:interactivestance}, this concept is named Partial Parallelism\footnote{\citet{Fernandez:thesis} previously addressed this concept as Sentential Antecedent (SA).}. Partial Parallelism is one way to categorize NSU classes according to the relation with their antecedents. NSU classes are categorized as +/-ParPar in order to find the right way to treat them. An NSU class categorized as +ParPar involves the access to one or more sub-utterances from its antecedent. On the contrary, -ParPar NSU classes do not need to know the internal structure of their antecedents to be resolved. Table \ref{table:parpar-nsus} shows how NSU classes are categorized in this way.

\begin{table}[h]
	\centering
	\begin{tabular}{|l l|}
		\hline
		\multicolumn{1}{|c}{\textbf{-ParPar}} &  \multicolumn{1}{c|}{\textbf{+ParPar}} \\ \hline
		Plain Acknowledgment         &	Short Answer \\
		Plain Affirmative Answer     & Repeated Acknowledgment \\
		Plain Rejection		         & Clarification Ellipsis \\
		Factual Modifier             & Repeated Affirmative Answer \\		
		Check Question               & Sluice \\		
		Propositional Modifier		 & Helpful Rejection \\
		& Filler \\
		& Bare Modifier Phrase \\
		& Conjunct \\ \hline
	\end{tabular}
	\caption{An overview of the NSU classes divided according to Partial Parallelism.}
	\label{table:parpar-nsus}
\end{table}

\subsection{Propositional lexemes}
\label{sec:prop_lex}
-ParPar NSU classes are realized (mainly) by \textit{propositional lexemes}, i.e. words that can stand alone and form a proposition with full contextual meaning. Among those classes there are the Plain Affirmative Answers, Plain Rejections and Propositional Modifiers which respectively are realized by the words \textit{yes}, \textit{no} and adverbials such as \textit{probably} and \textit{possibly}.

Those classes of NSU arise from polar questions such as \ref{ex:prop_lex1}.
\begin{example}
\label{ex:prop_lex1}
\begin{dialogue}
\speak{A} Will you go to the party on Saturday?
\speak{B} \textbf{Yes.} / \textbf{No.} / \textbf{Probably.}
\end{dialogue}
\end{example} 

The semantic content of these stand-alone lexemes can be modeled as a function $R$ of the content of the antecedent polar question.

For Plain Affirmative Answers, $R$ is the \textit{Identity} ($Id$) relation, i.e. the function that returns the argument itself. This means that the positive answer ``yes'' to a polar question is equivalent to the assertion of a proposition with the same content as the polar question.

For Plain Rejections, $R$ is the relation $Neg$. $Neg$ indicates the negation of a proposition $p$ although it is sensitive to the polarity of $p$, meaning that, when $p$ is positive, $Neg(p)$ is the negation of $p$ (denoted with $\bar{p}$) whereas, when $p$ is negative, $Neg(p)$ is $p$ itself. This rule is needed to account for the asymmetry in the meaning of negative answers to negative questions. A negative answer to a negative question does not equate a positive one, as exemplified in \ref{ex:prop_lex2} (rephrased from \citet{Ginzburg:interactivestance}).
\begin{example}
\label{ex:prop_lex2}
\begin{dialogue}
\speak{A} Did Paul not leave?
\speak{B} No. (= Paul did not leave.)
\end{dialogue}
\end{example} 

For Propositional Modifiers, $R$ is a relation $PropRel$ which applies different modalities on the basis of the lexical meaning of the word used as modifier, e.g. ``probably'' would have a different modality than ``clearly''.

\subsection{Focus Establishing Constituents}
To account for the partial parallelism between NSUs and their antecedents stemming out from the instances of the classes of the +ParPar group we need to keep track of the \textit{focal} sub-utterances of the antecedents i.e. of the elements of QUD. For this reason we employ the notion of \textit{focus establishing constituents} (FEC) from the theory of \citet{Ginzburg:interactivestance}\footnote{The concept was previously formalized by \citet{Fernandez:thesis} as \textit{topical constituents}.}. The FECs are relevant constituents in the elements of QUD that may be used to resolve NSUs.
Consider the following example:
\begin{example}
\label{ex:fec_ex}
\begin{dialogue}
\speak{A} \textbf{A friend} is coming to the party.
\speak{B} \textbf{Who?}
\end{dialogue}
\end{example}
The noun phrase ``A friend'' in the first sentence of \ref{ex:fec_ex} is the one which the following Sluice is referring to. Roughly the Sluice can be resolved in such a manner: ``\textit{Who is your friend that is coming to the party?}''.
It is clear that the aforementioned sub-utterance has to be contextually available to allow the resolution of the subsequent Sluice.
In this we follow \citet{Ginzburg:interactivestance}, who defines a set of rules to follow to make FECs contextually available. In particular we are interested in the following ones:
\begin{itemize}
\item The FEC associated with a \textit{wh}-interrogative is the \textit{wh}-phrase\footnote{As in \citet{Ginzburg:interactivestance} we consider only unary \textit{wh}-interrogatives. Refer to \citet{Fernandez:thesis} for an account of utterances with multiple \textit{wh}-interrogatives} itself:
\begin{example}
\begin{dialogue}
\speak{A} \textbf{Who} is organizing the party?
\speak{B} Paul.
\end{dialogue}
\end{example}
\item The FEC associated with a polar interrogative or declarative utterance can be any (quantified) noun phrases:
\begin{example}
\begin{dialogue}
\speak{A} \textbf{A friend} is organizing a party and \textbf{many people} are coming.
\speak{B} Who?
\end{dialogue}
\end{example}
\item The FEC associated with a clarification request is the sub-utterance that has to be clarified i.e. any sub-utterance in the antecedent:
\begin{example}
\begin{dialogue}
\speak{A} Is Paul organizing a party?
\speak{B} Paul? / Organizing? / A party?
\end{dialogue}
\end{example}

\end{itemize}



\subsection{Understanding and acceptance}
\label{sec:acceptance}
The classes of Plain Acknowledgments and Check Questions are used to handle understanding and acceptance in the conversation.
Plain Acknowledgments are used to send a direct feedback of understanding or acceptance of the previous utterances.
Understanding involves grasping successfully the content of an utterance while acceptance is a sign of shared belief which therefore updates the Facts with the accepted utterance and removes the corresponding issue from the QUD.
As argued in \citet{Fernandez:thesis}, understanding does not always imply acceptance, and Plain Acknowledgments are ambiguous in this distinction. Despite this difference, we assume that Plain Acknowledgments are used to show acceptance, therefore the use of a Plain Acknowledgments also downdates the QUD. On the other hand, understanding is assumed to be shown by any utterance that is not a Clarification Ellipsis.

Check Questions are used in conversation to request an explicit feedback about the understanding/acceptance of the previous utterance. 


\subsection{Sluicing}
\label{sec:sluicing}
Sluices can take a wide range of meanings depending on the particular situation. To formalize the meaning of Sluices, \citet{Fernandez:2007} distinguish four types of Sluices that convey different meanings: \textit{Direct Sluices}, \textit{Reprise Sluices}, \textit{Clarification Sluices}, \textit{\textit{Wh}-anaphor}.

The aforementioned paper describes a machine learning experiment to automatically classify Sluices according to these types. \citet{Ginzburg:interactivestance} describes several different treatments for every group of Sluices.

In our work we do not distinguish between those type of Sluices but we confine ourselves for simplicity to direct Sluices only. Direct Sluices, such as the one in \ref{ex:direct_sluice}, are used to query the other speaker for additional information about some aspect of the antecedent.
\begin{example}
\label{ex:direct_sluice}
\begin{dialogue}
\speak{A} Can I have some toast please?
\speak{B} Which sort? \\ \bncref{KCH}{104--105}
\end{dialogue}
\end{example}

\section{Dialogue context design}
\label{sec:context_design}
As mentioned before, the dialogue context is represented as a Bayesian network containing a set of random variables representing the current information state. The values of those random variables can represent virtually anything, from the raw utterances to their semantic representation. The variables in the dialogue context are inspired by \citet{Ginzburg:interactivestance}. In order to make the transition from the rules of \citet{Fernandez:thesis} to probabilistic rules as direct as possible, we mimic the basic dynamic of the DGB detailed in Section \ref{sec:formal}. For our semantics we do not employ TTR because it would add unnecessary complexity to our formalization. In this section we first describe the semantics we adopt and then we discuss the random variables that compose the dialogue context.

\subsection{Semantics}
The semantic content of the utterance is represented by logical predicates, individuals and variables. Predicates are labeled as words or camel-case phrases and can present zero or more arguments. Individuals are labeled with uppercase abbreviations such as IND for generic individuals or E for events. Variables are labeled with an uppercase X. Both individuals and variables are uniquely identified by a numeric subscript.

Predicates represent the high-level semantic meaning of the constituents of the utterances. Intuitively, predicates without variables as argument can represent propositions such as \ref{ex:pred_prop}. As discussed in Section \ref{sec:utt_repr}, polar questions and \textit{wh}-questions can be seen as functions from/to record types. Polar questions take as argument the empty record type. Following this schema, in our formalism polar questions are denoted by predicates with no variables, whereas \textit{wh}-questions are denoted by predicates containing one or more variables, as exemplified by \ref{ex:pred_quest}.
\begin{example}
\label{ex:pred_prop}
$ \text{Paul is a friend of yours.} \qquad \text{friend}(\text{addr}, \text{Paul}) $
\end{example}
\begin{example}
\label{ex:pred_quest}
$ \text{Is Paul a friend of yours?} \qquad \text{friend}(\text{addr}, \text{Paul}) $ \\
$ \text{Who is your friend?} \qquad \text{friend}(\text{addr}, \text{X}_1) $
\end{example}
Retrieving the semantic representation from the raw utterances is a Natural Language Understanding (NLU) task, a completely different task with respect to the resolution of NSUs. We do not attempt to generate predicates from raw utterances instead we make use of simple handcrafted predicates in our examples, abstracting the necessity of NLU to retrieve the meaning of all the utterances that are not NSUs.
We try to keep the problem of NSU resolution generic separating it as much as possible from the NLU task.


\subsection{Dialogue acts}
As seen in Section \ref{sec:utt_repr}, to represent the ``purpose'' of an utterance, we need to use an \textit{illocutionary relation}, also known as dialogue act. The set of dialogue acts we employ in our dialogue context is a small subset of the ones defined by \citet{Ginzburg:interactivestance}:
\begin{itemize}
\item \textit{Assert}, denoting the act of asserting a proposition;
\item \textit{Ask}, denoting the act of posing a question;
\item \textit{Ground}, denoting the act of understanding what being previously said;
\item \textit{Accept}, denoting the act of accepting what being previously said.
\end{itemize}

Assertions are applied to propositions and they are implicitly considered truthful unless they violate some predicates in the Facts. Asking a query is the act of posing questions and they are piled up in the QUD until they are resolved by an answer. The act of answering to a query corresponds, in the case of a \textit{wh}-interrogative, to finding the valid arguments to the variables of the question. In case of a polar question, the answer is derived simply by its truth status, denoted by the presence of the same predicate in the Facts. In our formalization we use ``Ground'' to represent the act understanding. Acceptance is the act of resolving an issue, which involves updating Facts and downdating QUD.



\subsection{Variables of the dialogue context}
\label{sec:variables}
For our formalization, as in TTR, we assume the availability of various data structures such as variables, lists, sets and complex types. The probabilistic rules formalism provides those structures out of the box as possible values for the random variables. Random variables are denoted with the notation ``$\mathtt{var_1}$''. Array element accesses are indicated with the square brackets notation, such as ``$\texttt{array[}0\texttt{]}$''. Sets are denoted with the classical mathematical notation ``$\{e_1,\dots,e_n\}$''. Complex type accesses are denoted with the dot notation, such as ``$\texttt{complexVar}.\texttt{var1}$''. The classical operations on sets are available such as the union and the intersection. Array concatenation is denoted with the $+$ symbol. Now we describe the variables used in our formalization of the dialogue context.

\bgroup
\titlespacing\subsubsection{0pt}{6pt}{0pt}
\subsubsection{$\mathtt{u_a}$, $\mathtt{u_b}$, $\mathtt{a_a}$, $\mathtt{a_b}$}
\egroup
As a convention, raw utterances and dialogue acts are indicated respectively with the letters $u$ and $a$ and a subscript denotes the speaker. We record in separate variables only the last utterance and dialogue act of each speaker.

\bgroup
\titlespacing\subsubsection{0pt}{6pt}{0pt}
\subsubsection{$\mathtt{nsu_a}$}
\egroup
A random variable that contains the distribution over the NSU classes returned by the classifier for the latest recorded utterance. It uses \texttt{max-qud} to refer to the antecedent therefore the probabilistic inference framework takes care of finding the most probable antecedent for the current NSU.
Besides the values of the NSU classes, a distinct value \texttt{NoNsu} is used to account for input utterances that are not NSUs. To determine whether an utterance is an NSU or not we used the same detection rules explained in Section \ref{sec:unlabeled_data_extraction}.

\bgroup
\titlespacing\subsubsection{0pt}{6pt}{0pt}
\subsubsection{\texttt{new-fec}}
\egroup
The set of FECs introduced by the NLU of the last recorded utterance. It is also a buffer variable used in the NSU resolution to encode the focal constituents of the newly resolved NSU. It is used also to hold FECs of the utterance that is being inserted in the \texttt{qud}.

\bgroup
\titlespacing\subsubsection{0pt}{6pt}{0pt}
\subsubsection{\texttt{facts}}
\egroup
A set of predicates representing the common knowledge of the users. The predicates in \texttt{facts} contain only individuals as arguments (i.e. no variables) and they are implicitly considered truthful. 

\bgroup
\titlespacing\subsubsection{0pt}{6pt}{0pt}
\subsubsection{\texttt{qud}}
\egroup
As defined in Section \ref{sec:formal}, the QUD is a partially ordered set containing the issues currently under discussion. Its  ordering determines the ``priority'' of the issues to be resolved. Here instead \texttt{qud} is represented as a vector and the \texttt{max-qud} variable denotes the index of the MaxQUD element (see below).
Each element in \texttt{qud} has a number of sub-fields:
\begin{itemize}
\item \texttt{utt}: The raw utterance associated to the current question under discussion;
\item \texttt{q}: The semantic representation of the utterance;
\item \texttt{fec}: An array of topical sub-utterances used in the resolution of the NSUs.
\end{itemize}
The \texttt{qud} is incremented by adding elements in the tail (growing numbers) and decremented in a random-access fashion, usually by removing the MaxQUD element (which could not be the last element) after its resolution. We denote with $\mathtt{qud_{size}}$ the size of \texttt{qud}.

\bgroup
\titlespacing\subsubsection{0pt}{6pt}{0pt}
\subsubsection{\texttt{max-qud}}
\egroup
Despite being represented as the maximal element of QUD in \citet{Ginzburg:interactivestance}, here \texttt{max-qud} actually denotes the index of such element which therefore is retrieved in this way: \texttt{qud[max-qud]}.
In \citet{Ginzburg:interactivestance}, MaxQUD is given from the partial ordering imposed on QUD. This ordering is often similar but not limited to the behavior of a stack.
At our disposal we have the full power of probabilistic modeling which enables us to encode \texttt{max-qud} as a random variable with a prior that gives more probability to the highest element in \texttt{qud}. The function used to give this prior to \texttt{max-qud} is $$ P(\texttt{max-qud}\!=\!i) = e^{i - \mathtt{qud_{size}}} $$
where $i < \mathtt{qud_{size}}$ is an index in \texttt{qud}.
In this way, the prior most probable MaxQUD is the last element inserted in the QUD but the probability can be modified by other contextual elements by probabilistic inference on the dialogue state.

\section{NSU resolution rules}
\label{sec:rules}
Here we present the probabilistic rules that handle the resolution of NSUs. For each rule we also present an example of usage.
Since they are a (almost) direct translation of the deterministic rules from \citet{Fernandez:thesis}, most of them have deterministic effects (i.e. a single effect with probability 1). Nonetheless the updates are handled probabilistically by the probabilistic rules framework through probabilistic inference over the Bayesian network representing the dialogue state. We show an example of probabilistic update in Section \ref{sec:rule_ack}, which is valid for every other resolution rule.

\subsection{Acknowledgments}
\label{sec:rule_ack}
The only requirement for Acknowledgment resolution is to have at least one issue under discussion to be accepted. As explained in Section \ref{sec:acceptance}, we assume that an explicit Acknowledgment is a sign of acceptance of the latest issue under discussion.
For Repeated Acknowledgments, \citet{Fernandez:thesis} requires to have co-referentiality between the repeated constituent in the NSU and the relative constituent in the FEC of MaxQUD. We decided to drop this requirement assuming that the co-reference is always present when the classifier assigns the class \textit{RepAck} to the current NSU. This assumption does not affect the system given that the effect on the state variables is the same for both \textit{Ack}s and \textit{RepAck}s. The rule for Acknowledgments is the following\footnote{The symbol $\leftarrow$ indicates the assignment of the right-hand side value to the left-hand side variable.}:
\begin{flalign*}
\ \ \ \ 
& \textit{ack}: \\
& \ \ \ \ 
\textbf{if} \ ((\mathtt{\texttt{nsu}_a}\!=\!\textit{Ack} \lor \mathtt{\texttt{nsu}_a}\!=\!\textit{RepAck}) \land \texttt{max-qud}\!>\!0)\ \textbf{then} & \\
& \ \ \ \ \;\;\;\;\; 
\begin{cases}
	P(\mathtt{a_a} \leftarrow \text{Accept}()) = 1
\end{cases}
\end{flalign*}

Consider the following example.
\begin{example}
\label{ex:ack}
\begin{dialogue}
\speak{B} I am going to the party.
\speak{A} OK.
\end{dialogue}
\end{example}
The dialogue context of \ref{ex:ack} is:
$$
\begin{array}{l c l}
	\texttt{max-qud} &=& 1 \\
	\texttt{qud[max-qud].q} &=& \text{goingToParty}(\text{IND}_1) \\
	\texttt{nsu}_{\texttt{a}} &=& \textit{Ack} \\
\end{array}
$$
After the application of the rule: $$\mathtt{a_a} = \text{Accept}()$$

Notice that this may be an oversimplification since often the values of the variables in the dialogue state are not determined with full probability but rather the variables encode a probability distribution over a set of values. For instance, it is often the case that the classifier will retrieve the type of the NSU in a probability distribution with one value with large probability and other few values with smaller probability scores. In this case we could have a situation resembling the following:
$$
	\texttt{nsu}_{\texttt{a}} = \begin{cases}
		\textit{Ack} & \text{with probability } 0.75 \\
		\textit{AffAns} & \text{with probability } 0.2 \\
		\textit{CheckQu} & \text{with probability } 0.05 \\
	\end{cases}
$$
The case above would result in the following distribution of $\mathtt{a_a}$\footnote{The actual distribution would not necessarily assign None as alternative value because other rules may be triggered by the other values of $\texttt{nsu}_{\texttt{a}}$.}:
$$
	\texttt{a}_{\texttt{a}} = \begin{cases}
		\text{Accept}() & \text{with probability } 0.75 \\
		\text{None} & \text{with probability } 0.25 \\
	\end{cases}
$$
Since the dialogue state is a Bayesian network, the update rules will return a distribution of values that is dependent on both the distribution assigned by the rule (in this case only one value with full probability) and the distributions of the variables the rule depend on. These considerations can be of course extended to all the other classes so in the following sections we will only point out the most relevant use cases.

\subsection{Affirmative Answers}
The context for an Affirmative Answer contains a polar question $q(\textbf{y})$ as MaxQUD.
As for the Acknowledgments, we drop the requirement of co-referentiability between the repeated constituent of the \textit{RepAffAns} and the same constituent in the FECs of the MaxQUD element.

An Affirmative Answer to a polar question corresponds to asserting the same semantic content (predicate) of the question. The following is the rule to handle Affirmative Answers\footnote{As a convention, quantified variables and quantified individuals in the rule definitions are indicated respectively as $x$ and $y$. Vectors of variables or individuals are indicated respectively as $\textbf{x}$ and $\textbf{y}$.}\footnote{As in this case, a probabilistic effect might contain several assignments. Hereafter, for readability, we write the sequence of assignments in a vertical notation.}:
\begin{flalign*}
\ \ \ \ 
& \textit{affAns}: \\
& \ \ \ \ \forall\,q,\textbf{y} \\
& \ \ \ \ 
\textbf{if} \ ((\mathtt{\texttt{nsu}_a}\!=\!\textit{AffAns} \lor \mathtt{\texttt{nsu}_a}\!=\!\textit{RepAffAns}) \land \texttt{qud[max-qud].q}\!=\! q(\textbf{y}))\ \textbf{then} & \\
& \ \ \ \ \;\;\;\;\; 
\begin{cases}
	P\begin{pmatrix} 
		\mathtt{\texttt{a}_a} \leftarrow \text{Assert}(q(\textbf{y})), \\
		\texttt{new-fec} \leftarrow \texttt{qud[max-qud].fec}
	\end{pmatrix} = 1
\end{cases}
\end{flalign*}
An example of application of the \textit{affAns} rule can be:
\begin{example}
\label{ex:affans}
\begin{dialogue}
\speak{B} Are you going to the party?
\speak{A} Yes.
\end{dialogue}
\end{example}
The context of \ref{ex:affans} is the following:
$$
\begin{array}{l c l}
	\texttt{max-qud} &=& 1 \\
	\texttt{qud[max-qud].q} &=& \text{goingToParty}(\text{IND}_2) \\
	\texttt{qud[max-qud].fec} &=& \{\} \\
	\texttt{nsu}_{\texttt{a}} &=& \textit{AffAns} \\
\end{array}
$$
After the application of the rule:
$$
\begin{array}{lcl}
	\mathtt{a_a} &=& \text{Assert}(\text{goingToParty}(\text{IND}_2)) \\
	\texttt{new-fec} &=& \{\} \\
\end{array}
$$

\subsection{Rejections}
\label{sec:reject}
As for the Affirmative Answers, the context of Rejections is a polar question $q(\textbf{y})$, but, as explained in Section \ref{sec:prop_lex}, we need to distinguish the cases in which $q$ is positive or negative. We will define the following function \textit{Neg} indicating the negation of a proposition $p$ (or equivalently a question): 
$$
	\textit{Neg}(p) = \begin{cases}
		\bar{p} & \text{if } p \text{ is positive} \\
		p & \text{if } p \text{ is negative}
	\end{cases}
$$
where $\bar{p}$ is the negative of $p$. As an extension of the above notation, we indicate a proposition that is explicitly negative as $\bar{p}$.

Rejections are handled by the following rule:
\begin{flalign*}
\ \ \ \ 
& \textit{reject}: \\
& \ \ \ \ \forall\,q,\textbf{y} \\
& \ \ \ \ 
\textbf{if} \ (\mathtt{\texttt{nsu}_a}\!=\!\textit{Reject} \land \texttt{qud[max-qud].q}\!=\! q(\textbf{y}))\ \textbf{then} & \\
& \ \ \ \ \;\;\;\;\; 
\begin{cases}
	P\begin{pmatrix} 
		\mathtt{\texttt{a}_a} \leftarrow \text{Assert}(\textit{Neg}(q)(\textbf{y})), \\
		\texttt{new-fec} \leftarrow \texttt{qud[max-qud].fec}
	\end{pmatrix} = 1
\end{cases} \\
& \ \ \ \ 
\textbf{else if} \ (\mathtt{\texttt{nsu}_a}\!=\!\textit{Reject} \land \texttt{qud[max-qud].q}\!=\! \bar{q}(\textbf{y}))\ \textbf{then} & \\
& \ \ \ \ \;\;\;\;\; 
\begin{cases}
	P\begin{pmatrix} 
		\mathtt{\texttt{a}_a} \leftarrow \text{Assert}(\bar{q}(\textbf{y})), \\
		\texttt{new-fec} \leftarrow \texttt{qud[max-qud].fec}
	\end{pmatrix} = 1
\end{cases}
\end{flalign*}
The following is an example for the above rule:
\begin{example}
\label{ex:rej}
\begin{dialogue}
\speak{B} Are you going to the party?
\speak{A} No.
\end{dialogue}
\end{example}
The context of \ref{ex:rej} is the following:
$$
\begin{array}{l c l}
	\texttt{max-qud} &=& 1 \\
	\texttt{qud[max-qud].q} &=& \text{goingToParty}(\text{IND}_2) \\
	\texttt{qud[max-qud].fec} &=& \{\} \\
	\texttt{nsu}_{\texttt{a}} &=& \textit{Reject} \\
\end{array}
$$
After the application of the rule:
$$
\begin{array}{lcl}
	\mathtt{a_a} &=& \text{Assert}(\textit{Neg}(\text{goingToParty})(\text{IND}_2)) \\
	\texttt{new-fec} &=& \{\} \\
\end{array}
$$

\subsection{Propositional Modifiers}
\label{sec:prop_mod}
As Affirmative Answers and Rejections, Propositional Modifiers are triggered by polar questions. As seen in Section \ref{sec:prop_lex}, their resolution corresponds to asserting the predicate of the polar question modified by a certain modality given by the lexical meaning of the NSU itself. 

We define the function $\textit{PropRel}_M(p)$ that modifies the meaning of a proposition $p$ (or equivalently a question) with the modality $M$. The modality is given by the lexical meaning of the word used in the NSU, here indicated for simplicity as the word itself (contained in the variable $\mathtt{u_a}$).

The rule for Propositional Modifiers states:
\begin{flalign*}
\ \ \ \ 
& \textit{propMod}: \\
& \ \ \ \ \forall\,q,\textbf{y} \\
& \ \ \ \ 
\textbf{if} \ (\mathtt{\texttt{nsu}_a}\!=\!\textit{PropMod} \land \texttt{qud[max-qud].q}\!=\! q(\textbf{y}))\ \textbf{then} & \\
& \ \ \ \ \;\;\;\;\; 
\begin{cases}
	P\begin{pmatrix} 
		\mathtt{\texttt{a}_a} \leftarrow \text{Assert}(\textit{PropRel}_{\mathtt{u_a}}(q)(\textbf{y})), \\
		\texttt{new-fec} \leftarrow \texttt{qud[max-qud].fec}
	\end{pmatrix} = 1
\end{cases}
\end{flalign*}

Here is an example of application of the above rule:
\begin{example}
	\label{ex:propmod}
	\begin{dialogue}
		\speak{B} Are you going to the party?
		\speak{A} Probably.
	\end{dialogue}
\end{example}

The dialogue state of \ref{ex:propmod} before the application of the rule is the following:
$$
\begin{array}{l c l}
	\texttt{max-qud} &=& 1 \\
	\texttt{qud[max-qud].q} &=& \text{goingToParty}(\text{IND}_2) \\
	\texttt{qud[max-qud].fec} &=& \{\} \\
	\texttt{nsu}_{\texttt{a}} &=& \textit{PropMod} \\
\end{array}
$$
After the application of the rule we would have:
$$
\begin{array}{lcl}
	\mathtt{a_a} &=& \text{Assert}(PropRel_{probably}(\text{goingToParty})(\text{IND}_2)) \\
	\texttt{new-fec} &=& \{\} \\
\end{array}
$$

Conversely to Affirmative Answers and Rejections, the Propositional Modifiers need to take into account the lexical meaning of the modifier used to update the dialogue state accordingly. This requires a set of lexicalized update rules to properly react to each possible modality of the modified proposition. However, these rules will only take place at the level of action selection and context update therefore it is still possible to resolve this kind of NSUs in a general way, as previously explained in Section \ref{sec:prop_lex}.

An example of lexicalized rule for updating the context in the presence of a modified proposition can be the following.
\begin{flalign*}
\ \ \ \ 
& \textit{factsIncrement}_{\textit{PropRel}}: \\
& \ \ \ \ \forall\,p,\textbf{y} \\
& \ \ \ \ 
\textbf{if} \ (\mathtt{a_b}\!=\!\text{Accept}(\textit{PropRel}_{\text{probably}}(p)(\textbf{y})))\ \textbf{then} & \\
& \ \ \ \ \;\;\;\;\; 
\begin{cases}
	P(\texttt{facts} \leftarrow \texttt{facts} \cup \{p(\textbf{y})\} \cup \texttt{new-fec}) = 0.75
\end{cases} \\
& \ \ \ \ 
\textbf{else if} \ (\mathtt{a_b}\!=\!\text{Accept}(\textit{PropRel}_{\text{unlikely}}(p)(\textbf{y})))\ \textbf{then} & \\
& \ \ \ \ \;\;\;\;\; 
\begin{cases}
	P(\texttt{facts} \leftarrow \texttt{facts} \cup \{p(\textbf{y})\} \cup \texttt{new-fec}) = 0.25
	\end{cases}
\end{flalign*}

This rule handles the update of the \texttt{facts} variable when the addressee decides to accept a proposition modified by a ``probably'' relation or by an ``unlikely'' relation. The latter is realized by updating \texttt{facts} with a high probability while the former updates \texttt{facts} with a low probability.

While the above rule has handcrafted probabilities, they can in principle be learned from actual data. Of course this would be only possible given a corpus containing NSU instances annotated with the dialogue acts and state updates at each step. When an instance of Propositional Modifier is encountered, the probabilities of the effects are updated according to the relative state update move.

\subsection{Check Questions}
As defined in Section \ref{sec:acceptance}, Check Questions are used to ask for understanding/acceptance of the latest issue being raised. In practice this means asking the latest asserted proposition as a polar question. The following is the rule to handle this type of NSUs:
\begin{flalign*}
\ \ \ \ 
& \textit{checkQu}: \\
& \ \ \ \ \forall\,p,\textbf{y} \\
& \ \ \ \ 
\textbf{if} \ (\mathtt{\texttt{nsu}_a}\!=\!\textit{CheckQu} \land \texttt{qud[max-qud].q}\!=\! p(\mathbf{y}))\ \textbf{then} & \\
& \ \ \ \ \;\;\;\;\; 
\begin{cases}
	P\begin{pmatrix} 
		\mathtt{\texttt{a}_a} \leftarrow \text{Ask}(p(\textbf{y})), \\
		\texttt{new-fec} \leftarrow \texttt{qud[max-qud].fec}
	\end{pmatrix} = 1
\end{cases}
\end{flalign*}
An example of application of the previous rule is the following:
\begin{example}
\label{ex:checkqu}
\begin{dialogue}
\speak{A} I am going to the party.
\speak{A} OK?
\end{dialogue}
\end{example}
The dialogue context of \ref{ex:checkqu} is:
$$
\begin{array}{l c l}
	\texttt{max-qud} &=& 1 \\
	\texttt{qud[max-qud].q} &=& \text{goingToParty}(\text{IND}_1) \\
	\texttt{qud[max-qud].fec} &=& \{\} \\
	\texttt{nsu}_{\texttt{a}} &=& \textit{CheckQu} \\
\end{array}
$$
After the application of the rule:
$$
\begin{array}{lcl}
	\mathtt{a_a} &=& \text{Ask}(\text{goingToParty}(\text{IND}_1)) \\
	\texttt{new-fec} &=& \{\} \\
\end{array}
$$

\subsection{Short Answers}
The antecedent of Short Answers is assumed to be a \textit{wh}-question. As stated previously, in this work we limit ourselves to unary \textit{wh}-interrogatives i.e. questions with only one unknown variable $x$. Short answers are resolved by applying them to the MaxQUD \textit{wh}-interrogative then asserting the resulting proposition. The application of the Short Answer is done by substituting every occurrences of the variable $x$ with $u_a$ (or equivalently a high-level representation of it). The following is the rule for Short Answers:
\begin{flalign*}
\ \ \ \ 
& \textit{shortAns}: \\
& \ \ \ \ \forall\,q,x,\textbf{y},p_i,\textbf{y}_i \\
& \ \ \ \ 
\textbf{if} \ (\mathtt{\texttt{nsu}_a}\!=\!\textit{ShortAns}\ \land\ \texttt{qud[max-qud].q}\!=\!q(x,\textbf{y})\ \land \\
& \ \ \ \ 
\ \ \ \ \{p_1(x,\textbf{y}_1), \dots , p_n(x, \textbf{y}_n)\}\!\subseteq\!\texttt{qud[max-qud].fec}) \ \textbf{then} & \\
& \ \ \ \ \;\;\;\;\; 
\begin{cases}
	\begin{pmatrix} 
	\mathtt{a_a} \leftarrow \text{Assert}(q(\mathtt{u_a}, \textbf{y})), \\
	\texttt{new-fec} \leftarrow \{p_1(\mathtt{u_a},\textbf{y}_1), \dots , p_n(\mathtt{u_a}, \textbf{y}_n)\}
	\end{pmatrix} = 1
\end{cases}
\end{flalign*}

An example of use of this rule is:
\begin{example}
	\label{ex:shortans}
	\begin{dialogue}
		\speak{B} Who is your friend organizing the party?
		\speak{A} Paul.
	\end{dialogue}
\end{example}

To make the example more appropriate we added a constituent that will be resolved by the rule together with the dialogue act. The context of \ref{ex:shortans} is:
$$
\begin{array}{l c l}
	\texttt{max-qud} &=& 1 \\
	\texttt{qud[max-qud].q} &=& \text{organizingTheParty}(\text{X}_1) \\
	\texttt{qud[max-qud].fec} &=& \{ \text{friend}(\text{IND}_2,\text{X}_1) \} \\
	\texttt{nsu}_{\texttt{a}} &=& \textit{ShortAns} \\
\end{array}
$$

After the application of the rule:
$$
\begin{array}{lcl}
	\mathtt{a_a} &=& \text{Assert}(\text{organizingTheParty}(\text{Paul})) \\
	\texttt{new-fec} &=& \{ \text{friend}(\text{IND}_2,\text{Paul}) \} \\
\end{array}
$$

\subsection{Sluices}
As argued in Section \ref{sec:sluicing}, we limit ourselves to the treatment of direct Sluices. Even for this type of Sluices only, the resolution rules are many since they have to account for the lexical meaning of each \textit{wh}-word. Furthermore, the meaning of \textit{wh}-words can be modified in many ways, e.g. ``how many'', ``how long'', ``who else'', ``what about''. This would require an extensive treatment for this kind of NSUs that we do not attempt to elaborate.

Nevertheless we will show some rules to treat simple direct Sluices like the ones in \ref{ex:sluice1} and \ref{ex:sluice2}.
\begin{example}
\label{ex:sluice1}
\begin{dialogue}
\speak{B} A friend is coming to the party.
\speak{A} Who?
\end{dialogue}
\end{example}
\begin{example}
\label{ex:sluice2}
\begin{dialogue}
\speak{B} Paul is throwing a party.
\speak{A} When?
\end{dialogue}
\end{example}

The context of the Sluices is a MaxQUD with at least one variable (raised by either a \textit{wh}-question or a proposition with some undefined reference). 
The Sluice is used to request some kind of information regarding one of the FECs of the antecedent. The requested information as well as the context generated by the resolution depend on the lexical meaning of the \textit{wh}-word. For instance, the following rule treats the Sluice ``\textit{who?}''.
\begin{flalign*}
\ \ \ \ 
& \textit{sluice}_{\textit{who}}: \\
& \ \ \ \ \forall\,q,x,\textbf{y},p_i,\textbf{y}_i \\
& \ \ \ \ 
\textbf{if} \ (\mathtt{\texttt{nsu}_a}\!=\!\textit{Sluice} \ \land \ \text{``who''}\!\in\!\mathtt{u_a}\ \land \ \texttt{qud[max-qud].q}\!=\! q(x, \textbf{y})\ \land \\
& \ \ \ \ 
\ \ \ \  \{p_1(x,\textbf{y}_1), \dots , p_n(x, \textbf{y}_n)\}\!\subseteq\!\texttt{qud[max-qud].fec})\  \textbf{then} & \\
& \ \ \ \ \;\;\;\;\; 
\begin{cases}
	P\begin{pmatrix} 
	\mathtt{a_a} \leftarrow \text{Ask}(\text{named}(x, \hat{x})), \\
	\texttt{new-fec} \leftarrow \{p_1(x,\textbf{y}_1), \dots , p_n(x, \textbf{y}_n)\} \cup \{person(x)\}
	\end{pmatrix} = 1
\end{cases}
\end{flalign*}
where $\hat{x}$ is a newly created variable.
Such a Sluice asks about the identity (here simplified by the name) of a \textit{person} which is referred to in the antecedent.

We can use as example the transcript \ref{ex:sluice1}. The context before the application of the rule is:
$$
\begin{array}{l c l}
	\texttt{max-qud} &=& 1 \\
	\texttt{qud[max-qud].q} &=& \text{comingToParty}(\text{X}_1) \\
	\texttt{qud[max-qud].fec} &=& \{ \text{friend}(\text{IND}_2,\text{X}_1) \} \\
	\texttt{nsu}_{\texttt{a}} &=& \textit{Sluice} \\
\end{array}
$$
After the application of the rule we have:
$$
\begin{array}{lcl}
	\mathtt{a_a} &=& \text{Ask}(\text{named}(\text{X}_1)) \\
	\texttt{new-fec} &=& \{ \text{friend}(\text{IND}_2,\text{X}_1), \text{person}(\text{X}_1) \} \\
\end{array}
$$

An interesting result of this rule in combination with the probabilistic inference employed in OpenDial is how the ambiguity in the FECs is handled. As argued in \citet{Ginzburg:interactivestance}, the antecedent of a Sluice can contain more than one potential FEC as exemplified by the following transcript.
\begin{example}
\label{ex:ambig_sluice}
\begin{dialogue}
\speak{B} \textbf{A friend of mine} is organizing a party for \textbf{his girlfriend}.
\speak{A} Who? (= Who is your friend? / Who is his girlfriend?)
\end{dialogue}
\end{example}

The resolution of this kind of ambiguities is automatically handled in a probabilistic fashion. In \ref{ex:ambig_sluice} the representation of the antecedent (MaxQUD) would be the following:
\begin{align*}
	\texttt{qud[max-qud].q}\ \ \ \ &=\ \text{organizingPartyFor}(\text{X}_1,\text{X}_2) \\
	\texttt{qud[max-qud].fec}\ &=\ \{\text{friend}(\text{IND}_1,\text{X}_1),\text{girlfriend}(\text{X}_2,\text{X}_1)\} \\
\end{align*}
In this case, the above rule would be applied to both variables $\text{X}_1$ and $\text{X}_2$, resulting in two possible assignments of $\mathtt{a_a}$ and \texttt{new-fec} with $0.5$ probability:
\begin{align*}
	\mathtt{a_a}\ &=\ \begin{cases}
	\text{Ask}(\text{named}(\text{X}_1,\text{X}_3)) & \text{with probability } 0.5 \\
	\text{Ask}(\text{named}(\text{X}_2,\text{X}_3)) & \text{with probability } 0.5 \\
	\end{cases} \\
	\texttt{new-fec}\ &=\ \begin{cases}
	\{\text{friend}(\text{IND}_1,\text{X}_1),\text{person}(\text{X}_1)\} & \text{with probability } 0.5 \\
	\{\text{girlfriend}(\text{X}_2,\text{X}_1),\text{person}(\text{X}_2)\} & \text{with probability } 0.5 \\
	\end{cases}
\end{align*}
Without any prior, the probabilistic inference over the dialogue state would assign equal probability to each possible assignment of $\mathtt{a_a}$. A more sophisticated approach may use some notion of \textit{saliency} as a prior to adjust the probabilities of each focal constituent. For example one could adjust the probability according to whether the constituent is a subject or an object in the antecedent. In the previous example the friend would have had more probability mass (e.g. $0.8$) and the girlfriend less probability mass (e.g. $0.2$). It is also possible that the prior saliency could depend on other variables such as the Facts or other contextual factors. Moreover, the parameters of the saliency function could also be learned from data.


\subsection{Clarification Ellipsis}
Clarification Ellipsis are a kind of \textit{clarification requests}. To resolve clarification requests and their elliptical variants, \citet{Ginzburg:interactivestance} includes a general theory of grounding and clarification requests. This theory would add a non-trivial amount of complexity to our formalization so we shall assume in our formalization that the latest utterance always grounded unless a clarification request comes afterwards. Therefore we resolve Clarification Ellipsis on the MaxQUD element without adding any other structure to the dialogue state.
We also consider the Clarification Ellipsis to have only a clausal confirmation reading, leaving aside their other possible readings which would require a more elaborate approach (more details in \citet{Ginzburg:interactivestance}). The clausal confirmation reading can be exemplified by \ref{ex:conf}.
\begin{example}
\label{ex:conf}
\begin{dialogue}
\speak{A} Is Paul coming to the party?
\speak{B} Paul? (= Are you asking if \textbf{Paul} is coming to the party?)
\end{dialogue}
\end{example}

The clausal confirmation reading can be interpreted as asking a polar question about the constituent brought about by the Clarification Ellipsis.

\begin{flalign*}
\ \ \ \ 
& \textit{CE}_{\textit{conf}}: \\
& \ \ \ \ \forall\,p,x,\textbf{y} \\
& \ \ \ \ 
\textbf{if} \ (\mathtt{\texttt{nsu}_a}\!=\!\textit{CE}\ \land \ \mathtt{u_a}\!=\!\text{``}x\text{?''} \land \ \\ & \ \ \ \ \ \ \ (\texttt{qud[max-qud].q}\!=\! p(x, \textbf{y})\ \lor \ p(x, \textbf{y})\!\in\!\texttt{qud[max-qud].fec}))\  \textbf{then} & \\
& \ \ \ \ \;\;\;\;\; 
\begin{cases}
	P\begin{pmatrix} 
	\mathtt{a_a} \leftarrow \text{Ask}(p(x, \textbf{y}))
	\end{pmatrix} = 1
\end{cases}
\end{flalign*}

As an example we can show the application of the rule on \ref{ex:conf}. The context is:
$$
\begin{array}{l c l}
	\texttt{max-qud} &=& 1 \\
	\texttt{qud[max-qud].q} &=& \text{comingToParty}(\text{IND}_1) \\
	\texttt{qud[max-qud].fec} &=& \{ \text{named}(\text{IND}_1,\text{Paul}) \} \\
	\texttt{nsu}_{\texttt{a}} &=& \textit{CE} \\
\end{array}
$$
The result of the application of the rule is:
$$
\begin{array}{lcl}
	\mathtt{a_a} &=& \text{Ask}(\text{named}(\text{IND}_1, \text{Paul})) \\
	\texttt{new-fec} &=& \{\} \\
\end{array}
$$

\section{Implementation and use case example}
\label{sec:testing}
In this section we exemplify some usages of the rules on a real-world conversation. It shows some example of behavior of the rules over a selected transcript from the \textsc{Communicator} dataset. However, we want to point out that this section is not intended to give an empirical evaluation of the rules which is very far from being trivial since such an evaluation would require (at least) the availability of a fully annotated dataset of transcripts with the dialogue acts and the context updates at each step.

The \textsc{Communicator} \citep{walker2001} dataset is a set of transcripts of interactions between a dialogue system and human testers. The \textsc{Communicator} dataset contains transcripts of conversations for booking flight tickets. The interactions are mainly ``machine-driven'' meaning that the system drives the conversation, it asks questions and the user only answer to those questions. In this scenario it is possible to find many answers NSUs such as Short Answers, Affirmative Answers and Rejections. To test the resolution rules over this transcript we integrated the rules within a dialogue system developed with the OpenDial toolkit.

Next we will briefly talk about the architecture of our dialogue system and then we elaborate the step-by-step description of the example of interaction with the system using the chosen transcript from the \textsc{Communicator} dataset.
\newpage

\subsection{Dialogue system architecture}
\begin{figure}[h]
	\centering
	\includegraphics[height=0.35\textheight]{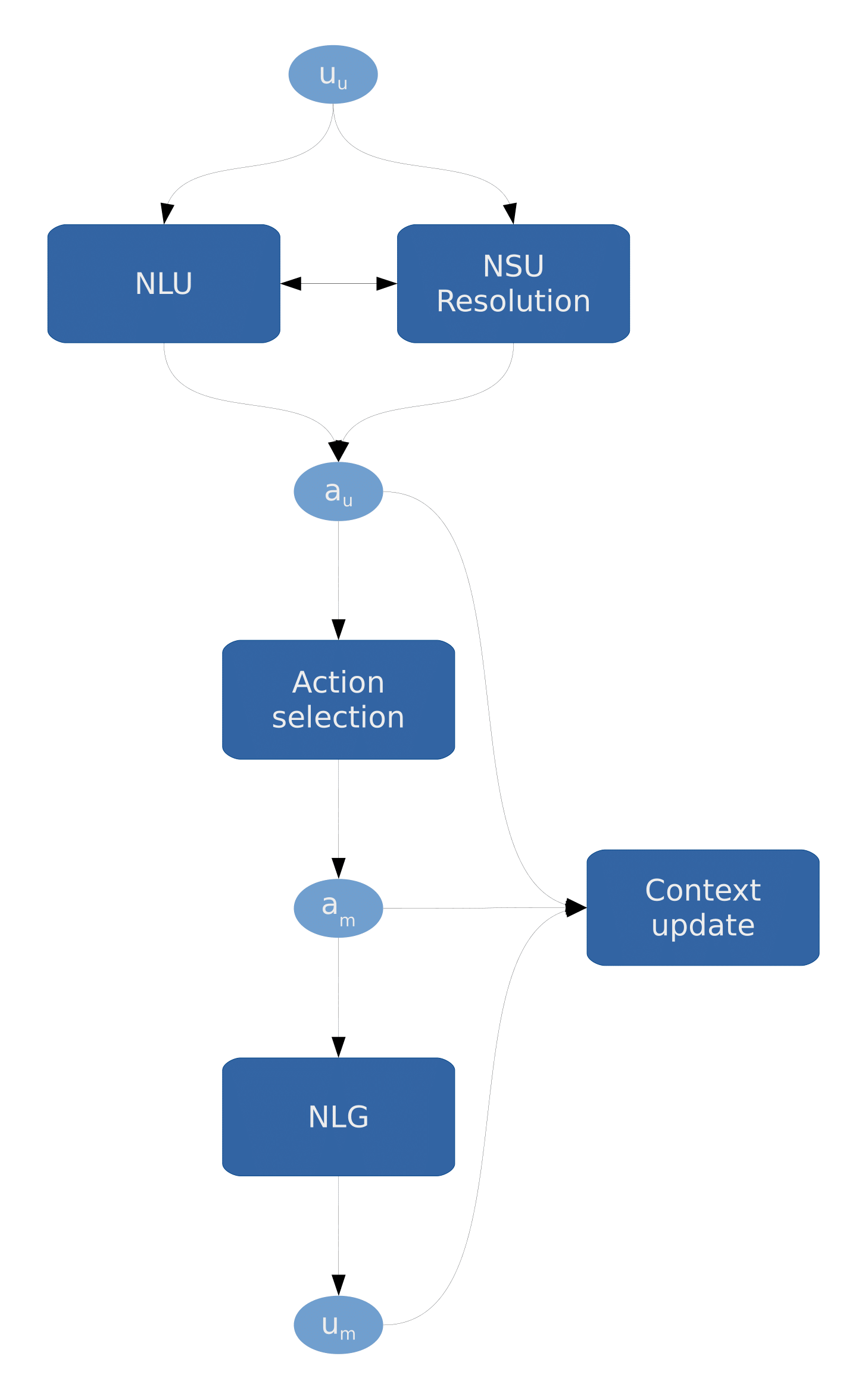}
	\caption{Schema of the basic work-flow of our dialogue system.}
	\label{fig:system_workflow}
\end{figure}

As defined in \citet{LisonThesis2014}, \textquote{a \textit{spoken dialogue system} is a computational agent that can converse with humans through everyday spoken language}. These systems have a complex structure formed of many different parts, however they are usually formed by the following major components:
\begin{itemize}
\item \textit{Natural language understanding} (NLU), maps the textual utterances into a high-level semantic representation;
\item \textit{Dialogue management}, updates the dialogue state and plans the actions to perform;
\item \textit{Natural language generation} (NLG), generates the linguistic realization of the planned actions or dialogue acts;
\end{itemize}
The resolution of NSUs is closely related to the NLU task and to the dialogue management in presence of such utterances. For its proper operation, our implementation indeed includes also shallow NLU and NLG modules as well as a very simple action selection procedure.
Figure \ref{fig:system_workflow} shows the work-flow of the system. The system takes as input a user utterance $u_u$. The NLU module generates the semantic representation of the utterance $a_u$, the NSU resolution takes place at this stage and involves the recovery of the right semantic form for the incomplete utterance using the information available in the dialogue context. The classifier, right before the resolution, generates the content of the variable $\textit{nsu}_u$. The action selection module decides what to do with the user utterance, producing a semantic representation of the action to perform $a_m$ (or dialogue act of the sentence to be uttered). The NLG module transforms $a_m$ into its linguistic form $u_m$. Throughout the process the context is updated by rules triggered by $a_u$, $a_m$ and $u_m$.

\subsection{Example}
In this section we present an extended example of the usage of some of the rules on the example transcript \ref{ex:communicator_1} from the \textsc{Communicator} dataset. The transcript \ref{ex:communicator_1} encodes a conversation between a dialogue system and a user with the purpose of booking a ticket for a flight.
\begin{example}
\label{ex:communicator_1}
\textsc{Communicator 1693:04}
\begin{dialogue}
\speak{     1. M} [\dots] What are your travel plans?
\speak{     2. U} I'd like to travel from Columbus Ohio to Phoenix Arizona on Thursday October 5th.
\speak{     3. M} What time do you want to leave Columbus?
\speak{     4. U} Before 6 P.M.
\speak{     5. M} Do you have a preferred airline?
\speak{     6. U} No.
\speak{     7. M} I have you going from Columbus Ohio to Phoenix Arizona on October 5th. Is that correct?
\speak{     8. U} Yes.
\speak{     9. M} Will you return to Columbus from Phoenix?
\speak{10. U} No.
\speak{11. M} Is Phoenix your final destination?
\speak{12. U} Yes.
\speak{13. M} Hold on while I check availability.
\end{dialogue}
\end{example}

As previously remarked, this type of conversations is driven by the system which asks the questions to fill the pieces of information it needs to properly retrieve the available flight options. The user answers to the questions posed by system, often using NSUs such as Short Answers, Affirmative Answers and Rejections. It is clear that this example covers only few of the available NSU classes but still it suffice at explaining the overall behavior of our implementation. More elaborate examples would have been complicated to understand and to read therefore we address the interested reader to the actual implementation\footnote{cf. \begin{scriptsize}\url{https://github.com/paolodragone}\end{scriptsize}} to have a more detailed look at the behavior of the system.

We want to stress that the predicates we use are simple and handcrafted. A much more sophisticated NLU module would be required to automatize the process of extracting those predicates from the sentences which is a completely different problem from the one we addressed in this work.

The following is a step-by-step explanation of the systems behavior through the updates of the dialogue state. The process is intrinsically probabilistic, for every move there are many updates with a certain probability but, for readability, we avoid to show each possible probabilistic update and we show the most probable one instead.
\begin{enumerate}
\item The system greets and asks for the the travel plans of the user, updating the QUD with the new question:
\begin{flalign*}
	\ \ \ \ &\mathtt{u_m} = \text{[\dots] What are your travel plans?} &
	\\
	&\mathtt{a_m} = \text{Ask}(\text{travelPlans}(\textit{x}_1, \textit{x}_2, \textit{x}_3)) &
	\\
	&\texttt{qud[1].q} = \text{travelPlans}(\textit{x}_1, \textit{x}_2, \textit{x}_3) &
	\\
	&\texttt{max-qud} = 1 &
\end{flalign*}
\item The user asserts its travel plans resolving the current issue under discussion:
\begin{flalign*}
	\ \ \ \ &\mathtt{u_u} = \text{I'd like to travel from Columbus Ohio to Phoenix Arizona on }&\\&\text{\ \ \ \ \ \,\, Thursday October 5th.} & \\
	& \mathtt{a_u} = \text{Assert}(\text{travelPlans}(\text{C}_1, \text{C}_2, \text{D}_1)) &\\
	& \texttt{new-fec} = \{\text{city}(\text{C}_1, \text{Columbus}), \text{city}(\text{C}_2, \text{Phoenix}), \text{date}(\text{D}_1, \text{05-10})\} &
\end{flalign*}
The system then accepts the assertion and resolves the issue:
\begin{flalign*}
	&\ \ \ \ \texttt{facts} = \{\text{travelPlans}(\text{C}_1, \text{C}_2, \text{D}_1),\\ 
	&\ \ \ \ \ \ \ \ \ \ \ \ \ \ \ \ \,\,\, \text{city}(\text{C}_1, \text{Columbus}), \text{city}(\text{C}_2, \text{Phoenix}), \text{date}(\text{D}_1, \text{05-10})\}& \\
	&\ \ \ \ \texttt{max-qud} = 0 &
\end{flalign*}
\item The system asks for time of departure, a new issue arises:
\begin{flalign*}
	\ \ \ \ &\mathtt{u_u} = \text{What time do you want to leave Columbus?} & \\
	& \mathtt{a_u} = \text{Ask}(\text{departTime}(\textit{x}_1)) &\\
	&\texttt{qud[1].q} = \text{departTime}(\textit{x}_1) & \\	
	&\texttt{max-qud} = 1 &
\end{flalign*}
\item The user resolves the issue with a Short Answer. The meaning of this NSU is inferred from the MaxQUD: the departure time must be before the given hour. The Short Answer resolution rule gives the following result:
\begin{flalign*}
	\ \ \ \ &\mathtt{u_u} = \text{Before 6 P.M.} & \\
	& \texttt{nsu}_\texttt{u} = \textit{ShortAns} & \\
	& \mathtt{a_u} = \text{Assert}(\text{departTime}(\text{T}_1)) &\\
	&\texttt{new-fec} = \{\text{before}(\text{T}_1, \text{T}_2), \text{time}(\text{T}_2, \text{18:00})\} &
\end{flalign*}
The system again acknowledges the answer of the user inserting his assertion in the set of Facts and downdates the QUD with the resolved issue:

\begin{flalign*}
	&\ \ \ \ \texttt{facts} = \{\text{travelPlans}(\text{C}_1, \text{C}_2, \text{D}_1),\\ 
	&\ \ \ \ \ \ \ \ \ \ \ \ \ \ \ \ \,\,\, \text{city}(\text{C}_1, \text{Columbus}), \text{city}(\text{C}_2, \text{Phoenix}), \text{date}(\text{D}_1, \text{05-10}), \\
	&\ \ \ \ \ \ \ \ \ \ \ \ \ \ \ \ \,\,\, \text{departTime}(\text{T}_1),\text{before}(\text{T}_1, \text{T}_2), \text{time}(\text{T}_2, \text{18:00})\}& \\
	&\ \ \ \ \texttt{max-qud} = 0 & 
\end{flalign*}
\item The system asks again a question about the preferred airline of the user. This time it is a polar question (notice the absence of a variable in the question predicate).
\begin{flalign*}
	\ \ \ \ &\mathtt{u_m} = \text{Do you have a preferred airline?} &
	\\
	&\mathtt{a_m} = \text{Ask}(\text{havePreferredAirline}(\text{user})) &
	\\
	&\texttt{qud[1].q} = \text{havePreferredAirline}(\text{user}) &
	\\
	&\texttt{max-qud} = 1 &
\end{flalign*}
\item The user gives a negative answer to the previous question using an NSU. Again the meaning is inferred from the MaxQUD: the user does not have a preferred airline. From Section \ref{sec:reject} we recall that the resolution rule in this case applies the \textit{Neg} function to the predicate to indicate its negative form.
\begin{flalign*}
	\ \ \ \ &\mathtt{u_u} = \text{No.} & \\
	& \texttt{nsu}_\texttt{u} = \textit{Reject} & \\
	& \mathtt{a_u} = \text{Assert}(\textit{Neg}(\text{havePreferredAirline})(\text{user})) &
\end{flalign*}
Again the answer of the user is inserted in the Facts:
\begin{flalign*}
	&\ \ \ \ \texttt{facts} = \{\text{travelPlans}(\text{C}_1, \text{C}_2, \text{D}_1),\\ 
	&\ \ \ \ \ \ \ \ \ \ \ \ \ \ \ \ \,\,\, \text{city}(\text{C}_1, \text{Columbus}), \text{city}(\text{C}_2, \text{Phoenix}), \text{date}(\text{D}_1, \text{05-10}), \\
	&\ \ \ \ \ \ \ \ \ \ \ \ \ \ \ \ \,\,\, \text{departTime}(\text{T}_1),\text{before}(\text{T}_1, \text{T}_2), \text{time}(\text{T}_2, \text{18:00}), \\
	&\ \ \ \ \ \ \ \ \ \ \ \ \ \ \ \ \,\,\, \textit{Neg}(\text{havePreferredAirline})(\text{user})\}& \\
	&\ \ \ \ \texttt{max-qud} = 0 &
\end{flalign*}
\item The system summarizes the pieces of information gained so far and queries the user for their correctness. A check question like this can be represented as asking the first proposition as a polar question in the following way:
\begin{flalign*}
	\ \ \ \ &\mathtt{u_m} = \text{I have you going \dots\ Is that correct?} &
	\\
	&\mathtt{a_m} = \text{Ask}(\text{travelPlans}(\text{C}_1, \text{C}_2, \text{D}_1)) &
	\\
	&\texttt{qud[1].q} = \text{travelPlans}(\text{C}_1, \text{C}_2, \text{D}_1) &
	\\
	&\texttt{qud[1].fec} = \{\text{city}(\text{C}_1, \text{Columbus}), \text{city}(\text{C}_2, \text{Phoenix}), \text{date}(\text{D}_1, \text{05-10})\} & \\
	&\texttt{max-qud} = 1 &
\end{flalign*}
\item The user confirms the information stored by the system with an Affirmative Answer. We recall that an Affirmative Answer is equivalent to stating the polar question in MaxQUD as it is:
\begin{flalign*}
	\ \ \ \ &\mathtt{u_u} = \text{Yes.} & \\
	& \texttt{nsu}_\texttt{u} = \textit{AffAns} & \\
	& \mathtt{a_u} = \text{Assert}(\text{travelPlans}(\text{C}_1, \text{C}_2, \text{D}_1)) & \\
	& \texttt{new-fec} = \{\text{city}(\text{C}_1, \text{Columbus}), \text{city}(\text{C}_2, \text{Phoenix}), \text{date}(\text{D}_1, \text{05-10})\} &
\end{flalign*}
The system acknowledges the answer by downdating the QUD but leaving Facts as it is since no additional information was included.
\begin{flalign*}
	&\ \ \ \ \texttt{max-qud} = 0 & \\
\end{flalign*}
\item The system asks a new question about a possible return flight.
\begin{flalign*}
	\ \ \ \ &\mathtt{u_m} = \text{Will you return to Columbus from Phoenix?} &
	\\
	&\mathtt{a_m} = \text{Ask}(\text{return}(\text{C}_2, \text{C}_1)) &
	\\
	&\texttt{qud[1].q} = \text{return}(\text{C}_2, \text{C}_1) &
	\\
	& \texttt{qud[1].fec} = \{\text{city}(\text{C}_1, \text{Columbus}), \text{city}(\text{C}_2, \text{Phoenix})\} & \\
	&\texttt{max-qud} = 1 &
\end{flalign*}
\item The user says that he will not return from its destination.
\begin{flalign*}
	\ \ \ \ &\mathtt{u_u} = \text{No.} &
	\\
	&\mathtt{a_u} = \text{Assert}(\textit{Neg}(\text{return})(\text{C}_2, \text{C}_1)) &
	\\
	&\texttt{facts} = \{\text{travelPlans}(\text{C}_1, \text{C}_2, \text{D}_1),\\ 
	&\ \ \ \ \ \ \ \ \ \ \ \ \,\,\, \text{city}(\text{C}_1, \text{Columbus}), \text{city}(\text{C}_2, \text{Phoenix}), \text{date}(\text{D}_1, \text{05-10}), \\
	&\ \ \ \ \ \ \ \ \ \ \ \ \,\,\, \text{departTime}(\text{T}_1),\text{before}(\text{T}_1, \text{T}_2), \text{time}(\text{T}_2, \text{18:00}), \\
	&\ \ \ \ \ \ \ \ \ \ \ \ \,\,\, \textit{Neg}(\text{havePreferredAirline})(\text{user}), \\
	&\ \ \ \ \ \ \ \ \ \ \ \ \,\,\, \textit{Neg}(\text{return})(\text{C}_2, \text{C}_1) \}& \\
	&\texttt{max-qud} = 0 &
\end{flalign*}
\item The system asks one last question about a its final destination.
\begin{flalign*}
	\ \ \ \ &\mathtt{u_m} = \text{Is Phoenix your final destination?} &
	\\
	&\mathtt{a_m} = \text{Ask}(\text{finalDest}(\text{C}_2)) &
	\\
	&\texttt{qud[1].q} = \text{finalDest}(\text{C}_2) &
	\\
	&\texttt{qud[1].fec} = \{\text{city}(\text{C}_2, \text{Phoenix})\} & \\
	&\texttt{max-qud} = 1 &
\end{flalign*}
\item The user confirms that Phoenix is his last destination.
\begin{flalign*}
	\ \ \ \ &\mathtt{u_u} = \text{Yes.} &
	\\
	&\mathtt{a_u} = \text{Assert}(\text{lastDest}(\text{C}_2)) &
	\\
	&\texttt{facts} = \{\text{travelPlans}(\text{C}_1, \text{C}_2, \text{D}_1),\\ 
	&\ \ \ \ \ \ \ \ \ \ \ \ \,\,\, \text{city}(\text{C}_1, \text{Columbus}), \text{city}(\text{C}_2, \text{Phoenix}), \text{date}(\text{D}_1, \text{05-10}), \\
	&\ \ \ \ \ \ \ \ \ \ \ \ \,\,\, \text{departTime}(\text{T}_1),\text{before}(\text{T}_1, \text{T}_2), \text{time}(\text{T}_2, \text{18:00}), \\
	&\ \ \ \ \ \ \ \ \ \ \ \ \,\,\, \textit{Neg}(\text{havePreferredAirline})(\text{user}), \\
	&\ \ \ \ \ \ \ \ \ \ \ \ \,\,\, \textit{Neg}(\text{willReturn})(\text{user}), & \\
	&\ \ \ \ \ \ \ \ \ \ \ \ \,\,\, \text{lastDest}(\text{C}_2) \}& \\
	&\texttt{max-qud} = 0 &
\end{flalign*}

\item After gathering up all the information needed, contained in the Facts, the system starts checking the availability of the flights for the user. 
\end{enumerate}

A possible continuation would be that the system finds some alternative flights to show to the user and ask him which one he or she prefers. The understanding in the dialogues in not always perfect though. It often happens that the system asks for repetition or even that the user resets or interrupts the conversation. This makes the dialogue transcripts from the \textsc{Communicator} dataset very unpredictable and a good starting poin to evaluate the benefits and limitations of probabilistic approaches to NSU resolution.

This example gives enough detail to understand the basic behavior of the system in this particular dialogue domain. It is of course limited in many ways and extensions could be sought in different directions. Testing the system on a different domain covering other types of NSUs should be perhaps the first way to check the validity of the rules. However, a complete evaluation would require the manual annotation of several dialogue transcripts, which is not an easy task.

Besides, the behavior of the rules rely on a fixed semantics that is highly domain-dependent. To make the system scalable to different domains one needs to integrate other grammatical and lexical resources to make the semantics as generic as possible. It is clear though that a complete domain independent framework would be difficult to achieve.

Our system is of course still a prototype of a complete dialogue system for NSU interpretation. There are many improvement needed to achieve a working system. For instance the inclusion of rules to handle the NSU classes not covered by our work. A more general theory of grounding is also needed to properly account for the clarification requests. Extend every other assumption we made to simplify the development is another important goal for possible future works. Furthermore, to enhance the capabilities of the system, the integration of other natural language understanding modules is needed as well. Anaphora Resolution \citep{mitkov2014anaphora} and Named Entity Recognition \citep{nadeau2007survey} are just a few of the problems concerning the correct interpretation of NSUs.

\section{Summary}
In this chapter we presented how to resolve the semantic meaning of NSUs. The resolution is a task that aims at extracting the meaning of an NSU given the dialogue context. To do so, we relied on the previous work of \citet{Fernandez:thesis} which presented a series of rules to resolve the meaning of the NSUs from a TTR-encoded dialogue context. As previously argued throughout this thesis, the use of a purely logic-based formalism, such as TTR, has some disadvantages in dealing with partially observable inputs and stochastic events when compared to a probabilistic approach. We showed how to reformulate the rules from \citet{Fernandez:thesis} using the probabilistic rules formalism \citep{LisonThesis2014} in order to include a probabilistic account of the dialogue state. We made use of a portion of the dialogue context theory from \citet{Ginzburg:interactivestance} to encode the basic elements of the dialogue state needed for the resolution of the NSUs (see Section \ref{sec:formal}). We presented in Section \ref{sec:rules} the probabilistic rules for the resolution of the NSUs. In Section \ref{sec:testing} we also described a step-by-step example of the usage of the rules. The framework presented in this chapter has also been implemented and tested with OpenDial \citep{semdial2015_opendial}.

\chapter{Conclusion}

This chapter concludes the present thesis by summarizing the contributions of our work. The chapter also points out some of the possible developments that can be pursued in future works.

\section{Contributions}

In this thesis we described our research work regarding non-sentential utterances. NSUs are utterances that do not have a complete sentential form but convey a full meaning. However, they require to be ``interpreted'' i.e. their meaning must be extracted from the context of the dialogue. Our experiments concerned two separate aspects of the interpretation of the NSUs, namely:
\begin{itemize}
\item The classification of NSUs given their context;
\item The resolution of the semantic content of the NSUs from the dialogue context;
\end{itemize}

In Chapter \ref{ch:background} we presented the background knowledge needed to the development of our work. We discussed in Section \ref{sec:NSU} the concept of non-sentential utterance, referring to \citet{Fernandez:thesis} as our theoretical basis. From the aforementioned work we employ the same taxonomy and corpus of NSUs. We then explain our methodology for the interpretation of NSUs, also based on the theory from \citet{Fernandez:thesis}. To interpret an NSU, we first classify it according to the aforementioned taxonomy using machine learning then we ``resolve'' its meaning from the dialogue context through a resolution procedure dependent on its type. 

\citet{Fernandez:thesis} develops a set of resolution procedures based on Type Theory with Records \citep{cooper2004type,Ginzburg:interactivestance}. In Section \ref{sec:formal} we briefly describe the aspects of TTR and the theory of dialogue context from \citet{Ginzburg:interactivestance} that we needed in our work. 

In this thesis we argue that a purely logical framework such as TTR may have disadvantages in dealing with the uncertain nature of the NSUs. In our view a proper alternative is a probabilistic approach to the resolution of NSUs. To this end we employ the probabilistic rules formalism and the theory from \citet{LisonThesis2014} as probabilistic representation of the dialogue state and the NSU resolution procedures. We detail the basic aspects of the theory from \citet{LisonThesis2014} in Section \ref{sec:prob_model}. In \citet{LisonThesis2014} the dialogue state is represented as a Bayesian network and its dynamics are described by probabilistic rules. Probabilistic rules are \textit{if \dots\ then \dots\ else \dots} constructs that map logical conditions to probabilistic effects.

The focus of Chapter \ref{ch:classification} is on the classification of NSUs, which is the task of inferring the type of a given NSU from its context. The context of a NSU is formed by its ``antecedent'', the preceding utterance that holds its hidden meaning. Our work on the classification of NSUs is based on \citet{Fernandez:2007}. We replicated their approach and set it as our baseline, as explained in Section \ref{sec:baseline}. In Section \ref{sec:extended} we describe the new features we use to extend the baseline feature set. The extended feature set alone was not enough to achieve an improvement of the classification performances. The major problems in this respect were the scarcity of labeled data and the class imbalance. To address those problems we employed semi-supervised learning techniques that we detail in Section \ref{sec:ssl}. Our experiments show that the combination of the extended feature set and new training instances labeled with Active Learning led to a significant improvement of the classification accuracy. Nevertheless we argue that further analysis and testing need a larger amount of labeled data to be carried out properly. In \citet{Dragone:2015} we present our findings in the classification of NSUs using Active Learning. 

In Chapter \ref{ch:resolution} we detail the resolution of NSUs. The NSU resolution is the task of extracting the meaning of a given NSU from the dialogue context. We explain the process that we employ for the NSU resolution in Section \ref{sec:res_task}. In Section \ref{sec:theoretical_foundation} we describe the theoretical concepts we need from \citet{Ginzburg:interactivestance}. The description of our theory starts in Section \ref{sec:context_design} where we explained the design of the dialogue context. To model our dialogue context we take inspiration from \citet{Ginzburg:interactivestance}, however we reinterpret its constructs as random variables. The random variables in the dialogue state interact with each other through the probabilistic rules. The resolution rules that we developed are explained in Section \ref{sec:rules}. Finally, in Section \ref{sec:testing}, we show a detailed example of the use of the resolution rules over a transcript from the \textsc{Communicator} dataset \citep{walker2001}.

Our approach to the resolution of NSUs is intended to be a proof-of-concept for this task. We showed how we could reuse many concepts from the theory of \citet{Fernandez:thesis} and \citet{Ginzburg:interactivestance} and ``translate'' the resolution rules based on TTR into probabilistic rules. The works on classification and interpretation carried out in this thesis were also presented in \citet{dragone2015non}.

\section{Future developments}
In this section we list a series of ideas for possible future works that come out directly from our findings and from the assumptions we made. 

\subsubsection{Improve the NSU classification performances}
In our work on the classification of NSUs we experimented many different approaches in seeking an improvement of the classification accuracy. However, there are many other paths that we did not explore or aspects in our approaches that can be improved. Here we discuss a few of the possible extension of our work. 

There are classes of NSU that are intrinsically difficult to predict such as Helpful Rejections and other pairs of classes that are difficult to discriminate such as Repeated Acknowledgments and Repeated Affirmative Answers. A common issue in trying to predict those classes is that the parallelism with their antecedent is almost entirely at the semantic level. This requires deeper understanding of the phenomena and the use of features that exploit semantic relations in the NSU instances. We did not use any semantic feature since it would have added a non-trivial amount of complexity to our feature extraction algorithms. The deeper understanding of ``difficult'' classes and the use of such features may be a good starting point to any feature work on this topic.

Using additional features does not avoid the problem of class imbalance in the dataset. Many techniques could be experimented to try to mitigate this issue. An example may be an over-sampling technique such as \textit{SMOTE}\footnote{Synthetic Minority Over-sampling Technique.} \citep{chawla2002smote}. The aforementioned work shows that the combination of SMOTE and majority class under-sampling leads to better classification performances on certain domains.

Perhaps the most difficult issue to overcome is the scarcity of labeled data. Our work shows that additional training data is indeed useful to improve the classification performances but we still lack enough data to run proper evaluations. We did not use the instances labeled with Active Learning as test data. Additional data for the gold standard should be composed of high-quality, manually annotated instances extracted within a corpus study that closely follows the original one from \citet{Fernandez:2002}.

\subsubsection{Incorporate additional elliptical phenomena}
The corpus study from \citet{Fernandez:2002} is focused on data extracted from the British National Corpus therefore confined to only certain kind of dialogue domains. As argued by \citet{IBMWatson:2015}, there are many elliptical phenomena that do not fit well in the taxonomy from \citet{Fernandez:2002}. An interesting follow up work on non-sentential utterances might try to find new elliptical phenomena in different dialogue domains and try to extend the taxonomy. Another point that may be considered is that some classes include large variety of forms and functions such as Short Answers and Helpful Rejections. A possible development could be to increase the granularity of such classes to try to capture more subtle differences.

\subsubsection{Extend our NSU resolution approach}
Our study on the resolution of NSUs pioneers a rule-based approach that involves a probabilistic dynamics of the dialogue state.
There are still many issues to address:
\begin{itemize}
\item Increase the coverage of the rules to all the classes that were not covered by our work: Factual Modifiers, Helpful Rejections and so on.
\item Develop a proper mechanism of rule adaptation in the presence of lexical modifiers e.g. for Sluices such as ``For how long?''.
\item Include grammatical and lexical resources to extract more complex meanings. A simple Short Answer that would not be covered by our rules is:
\begin{example}
\begin{dialogue}
\speak{A} Who is coming tomorrow?
\speak{B} Nobody.
\end{dialogue}
\end{example}
\item Properly evaluate the rules on testing data from different dialogue domains.
\end{itemize}
The last point, perhaps the most important one, would require the development of a corpus of dialogue transcripts annotated with each semantic move and state update. In turn this would require to develop a generic representation of the semantic content of the utterances which is a non-trivial task by itself. A possibility could be to use TTR as semantic representation and reformulate the rules accordingly.
Using TTR as basic semantic formalism it could be an interesting challenge to develop probabilistic rules to address a larger set of linguistic phenomena besides NSUs.

\subsubsection{Combine different NSU resolution approaches}
For our work on the NSU resolution we develop a rule-based approach based on a probabilistic representation of the dialogue state. It bares similarities with statistical approaches for the resolution such as \citet{IBMWatson:2015}. Their work is concentrated on follow-up NSU questions such as:
\begin{example}
\label{ex:ibm}
\begin{dialogue}
\speak{A} How much for this model?
\speak{B} \dots
\speak{A} For this other one?
\end{dialogue}
\end{example}
Their approach is based on the combination of keywords from the follow-up question and the original one. From the combination of keywords they build possible meaningful ``completions'' of the NSU e.g. a completion for the NSU at the third line in \ref{ex:ibm} would be ``How much for this other model?''. After generating each possible completions they rank them according to some score and pick the best one. 

Differently from ours, their approach does not use a high-level semantic representation of the utterances. It would be interesting to try to combine their statistical approach to our probabilistic rule-based one.

\subsubsection{Compare our approach with other existing systems}
Another useful comparison, and perhaps integration, should be made with the systems originally developed on the theory of \citet{Fernandez:thesis}, namely SHARDS \citep{Fernandez:shards} and one of its extensions CLARIE \citep{Purver:2006}. The former is a system for ellipsis resolution that can handle Short Answer, Sluices and Affirmative Answers. The latter is a dialogue system developed to deal with \textit{clarification requests} and, among them, Clarification Ellipsis, implementing the theory of \citet{ginzburg2004clarification} on top of the GoDiS dialogue system \citep{larsson2000godis}. Both are based on the HPSG\footnote{Head-driven Phrase Structure Grammar.} framework from \citet{ginzburg2000interrogative}, which is substantially different from our current design. Our framework lacks a grammar and other lexical resources that are indeed needed to build a functional system. It would be interesting to further develop our approach taking advantage from aspects of those systems and perhaps even integrate them into our architecture based on probabilistic rules.

\begin{appendices}
\chapter{Context update rules}
\label{app:1}

In this appendix we provide an overview of the probabilistic rules used for updating the context that have been implemented in the dialogue system for the testing of the resolution rules.
As described in Section \ref{sec:testing}, the dialogue system that we implemented is focused on conversation of the human-machine kind therefore we will be using the notations $a_m$ and $u_m$ to refer respectively to the dialogue act performed by the system and the corresponding raw utterance. Following the aforementioned interaction model, the dialogue context is only representing the pieces of information known by the system. The context update rules are needed in order to make the dialogue context evolve along with the user acts and relative system reactions. In particular we need the rules for updating the QUD and the Facts variables. These rules are inspired by, but not limited to, \citet{Ginzburg:interactivestance}. Section \ref{sec:formal} gives the background knowledge for the rules from \citet{Ginzburg:interactivestance}.

The rules shown in this appendix are not meant to give an extensive look on the system architecture but rather an high-level insight on the behavior of the system. These rules may differ from the actual implementation due to technicalities but still they fit for the purpose of the explanation. We will not talk about other modules of the system that we used in the implementation (i.e. NLU, NLG and action selection) because they have been implemented merely for toy examples of interaction on simple domains and they are not directly concerning the interpretation of NSUs. Follows a description of each context update rule.

\subsubsection{QUD increment}
NSUs are mostly reactionary utterances to previously raised issues. We aim at using the NSU resolution rules to interpret the content of the user NSUs. For this reason we concentrate on ``machine-driven'' conversations such as the one used in Section \ref{sec:testing}. We assume that in this type of dialogues issues are raised only by the system while the user limits to answer. In this scenario is common to find NSUs uttered by the users, as it often happens in the \textsc{Communicator} dataset.

Given this setting, we update the QUD only when the system raises a new issue and we downdate it only when the system accepts a user assertion which resolves the maximal element. We also take into account that asking a question and asserting a proposition may have different probabilities to update the QUD. In the rules below we encode the asking with full probability and the asserting with a probability of $0.75$, although, as remarked many times throughout this thesis, those probabilities may actually be estimated on real data.
\begin{flalign*}
\ \ \ \ 
& \textit{qud-increment}: \\
& \ \ \ \ \forall\ q,\textbf{x} \\
& \ \ \ \ 
\textbf{if} \ (\mathtt{\texttt{a}_m}\!=\!\text{Ask}(q(\textbf{x})))\  \textbf{then} & \\
& \ \ \ \ \;\;\;\;\; 
\begin{cases}
	P\begin{pmatrix} 
	\mathtt{qud[qud.size + 1].q} \leftarrow q(\textbf{x}), \\
	\mathtt{qud[qud.size + 1].utt} \leftarrow \mathtt{u_m}, \\
	\mathtt{qud.size} \leftarrow \mathtt{qud.size} + 1
	\end{pmatrix} = 1
\end{cases} \\
& \ \ \ \
\textbf{else if} \ (\mathtt{\texttt{a}_m}\!=\!\text{Assert}(q(\textbf{x})))\  \textbf{then} & \\
& \ \ \ \ \;\;\;\;\; 
\begin{cases}
	P\begin{pmatrix} 
	\mathtt{qud[qud.size + 1].q} \leftarrow q(\textbf{x}), \\
	\mathtt{qud[qud.size + 1].utt} \leftarrow \mathtt{u_m}, \\
	\mathtt{qud.size} \leftarrow \mathtt{qud.size} + 1
	\end{pmatrix} = 0.75
\end{cases}
\end{flalign*}
The rule below handles the update of the FEC of the newly added QUD element. The rule adds to the FEC of the new element of QUD only the \texttt{new-fec} predicates sharing at least a variable with the proposition predicate.
\begin{flalign*}
\ \ \ \ 
& \textit{fec-update}: \\
& \ \ \ \ \forall\ p,\textbf{x},p',\textbf{x}',x \\
& \ \ \ \
\textbf{if} \ ((\mathtt{\texttt{a}_m}\!=\!\text{Ask}(p(\textbf{x})) \lor \mathtt{\texttt{a}_m}\!=\!\text{Assert}(p(\textbf{x}))) \land p'(\textbf{x}')\!\in\!\texttt{new-fec} \land x\!\in\!\textbf{x} \land x\!\in\!\textbf{x}')\  \textbf{then} & \\
& \ \ \ \ \;\;\;\;\; 
\begin{cases}
	P\begin{pmatrix} 
	\mathtt{qud[qud.size + 1].fec} \leftarrow \mathtt{qud[qud.size + 1].fec} \cup \{p'(\textbf{x}')\}
	\end{pmatrix} = 1
\end{cases}
\end{flalign*}
\subsubsection{QUD downdate}
The QUD is downdated when the system accepts a proposition asserted by the user. The system responds with an Accept act which will remove the MaxQUD from the QUD.
\begin{flalign*}
\ \ \ \ 
& \textit{qud-downdate}: \\
& \ \ \ \ 
\textbf{if} \ (\mathtt{\texttt{a}_m}\!=\!\text{Accept}(p))\  \textbf{then} & \\
& \ \ \ \ \;\;\;\;\; 
\begin{cases}
	P\begin{pmatrix} 
	\texttt{qud[max-qud].q} \leftarrow \text{None}, \\
	\texttt{qud[max-qud].utt} \leftarrow \text{None}, \\
	\texttt{qud[max-qud].fec} \leftarrow \text{None}, \\
	\texttt{qud.size} \leftarrow \texttt{qud.size} - 1 \\
	\end{pmatrix} = 1
\end{cases}
\end{flalign*}
\subsubsection{Max-qud update}
As soon as the QUD array is updated, the MaxQUD is updated too. As explained in Section \ref{sec:variables}, the \texttt{max-qud} variable is defined as the index of the MaxQUD inside the QUD array and its stack-like behavior is determined by an exponentially decreasing probability with maximum on the last inserted element.
\begin{flalign*}
\ \ \ \ 
& \textit{max-qud-update}: \\
& \ \ \ \ \forall i \\
& \ \ \ \ 
\textbf{if} \ (i>0\ \land\ i\leq \texttt{qud.size}\ \land\ \texttt{qud[}i\texttt{].q}\neq\text{None})\  \textbf{then} & \\
& \ \ \ \ \;\;\;\;\; 
\begin{cases}
	P(\texttt{max-qud} \leftarrow i) = e^{i-\texttt{qud.size}}
\end{cases}
\end{flalign*}
\subsubsection{Facts increment}
As mentioned above, the dialogue context encodes the knowledge of the system and so are the Facts. The Facts variable contains only predicates accepted by the system. In the rule below we incorporate the ones presented in Section \ref{sec:prop_mod} for Propositional Modifiers and the ones for handling Rejections and Affirmative Answers. Again we point out that while the probabilities are handcrafted for simplicity here, they can be actually learned from data.
\begin{flalign*}
\ \ \ \ 
& \textit{facts-increment}: \\
& \ \ \ \ \forall\ p,\textbf{y} \\
& \ \ \ \ 
\textbf{if} \ (\mathtt{a_m}\!=\!\text{Accept}(\textit{PropRel}_{\text{probably}}(p)(\textbf{y})))\ \textbf{then} & \\
& \ \ \ \ \;\;\;\;\; 
\begin{cases}
	P(\texttt{facts} \leftarrow \texttt{facts} \cup \{p(\textbf{y})\} \cup \texttt{new-fec}) = 0.75
\end{cases} \\
& \ \ \ \ 
\textbf{else if} \ (\mathtt{a_m}\!=\!\text{Accept}(\textit{PropRel}_{\text{unlikely}}(p)(\textbf{y})))\ \textbf{then} & \\
& \ \ \ \ \;\;\;\;\; 
\begin{cases}
	P(\texttt{facts} \leftarrow \texttt{facts} \cup \{p(\textbf{y})\} \cup \texttt{new-fec}) = 0.25
\end{cases} \\
& \ \ \ \ \dots \\
& \ \ \ \ 
\textbf{else if} \ (\mathtt{a_m}\!=\!\text{Accept}(\textit{Neg}(p)(\textbf{y})))\ \textbf{then} & \\
& \ \ \ \ \;\;\;\;\; 
\begin{cases}
	P(\texttt{facts} \leftarrow \texttt{facts} \cup \{\textit{Neg}(p)(\textbf{y})\} \cup \texttt{new-fec}) = 1
\end{cases} \\
& \ \ \ \ 
\textbf{else if} \ (\mathtt{a_m}\!=\!\text{Accept}(p(\textbf{y})))\ \textbf{then} & \\
& \ \ \ \ \;\;\;\;\; 
\begin{cases}
	P(\texttt{facts} \leftarrow \texttt{facts} \cup \{p(\textbf{y})\} \cup \texttt{new-fec}) = 1
\end{cases}
\end{flalign*}
\end{appendices}

\backmatter
\begingroup
\setstretch{1.2}
\setlength\bibitemsep{10pt}
\printbibliography
\endgroup

\end{document}